\documentclass[runningheads]{llncs}
\usepackage{graphicx}
\usepackage{amsmath,amssymb}
\usepackage{color}
\usepackage{comment}
\usepackage{subcaption}
\usepackage{dsfont}
\usepackage{multirow}
\usepackage{hyperref}
\usepackage{breakcites}
\newcommand{\rev}[1]{{\textcolor{black}{#1}}}
\begin{document}
\pagestyle{headings}
\mainmatter
\def\ACCV20SubNumber{259}  
\newcommand{\weidi}[1]{{\textcolor{red}{[Weidi: #1]}}}
\title{Betrayed by Motion: Camouflaged Object Discovery via Motion Segmentation }
\titlerunning{Betrayed by Motion}
\author{Hala Lamdouar \and
Charig Yang \and
Weidi Xie \and
Andrew Zisserman}
\authorrunning{H. Lamdouar et al.}
\institute{Visual Geometry Group, University of Oxford \\
\email{\{lamdouar,charig,weidi,az\}@robots.ox.ac.uk}\\ [2pt]
\url{http://www.robots.ox.ac.uk/~vgg/data/MoCA}}

\maketitle

\begin{abstract}
The objective of this paper is to design a computational architecture that discovers camouflaged objects in videos, 
specifically by exploiting motion information to perform object segmentation.
We make the following three contributions:
(i) We propose a novel architecture that consists of two essential components for breaking camouflage,
namely, a differentiable registration module to align consecutive frames based on the background, 
which effectively emphasises the object boundary in the difference image, 
and a motion segmentation module with memory that discovers the moving objects,
while maintaining the object permanence even when motion is absent at some point.
(ii) We collect the first large-scale Moving Camouflaged Animals (MoCA) video dataset, 
which consists of over 140 clips across a diverse range of animals~($67$ categories).
(iii) We demonstrate the effectiveness of the proposed model on MoCA, 
and achieve competitive performance on the unsupervised segmentation protocol on DAVIS2016 by only relying on motion.
\begin{keywords}
Camouflage Breaking; Motion Segmentation;
\end{keywords}
\end{abstract}

\section{Introduction}

\begin{figure}[!htb]
\footnotesize
\centering
\begin{subfigure}[t]{.32\textwidth}
  \centering
  \includegraphics[width=\linewidth]{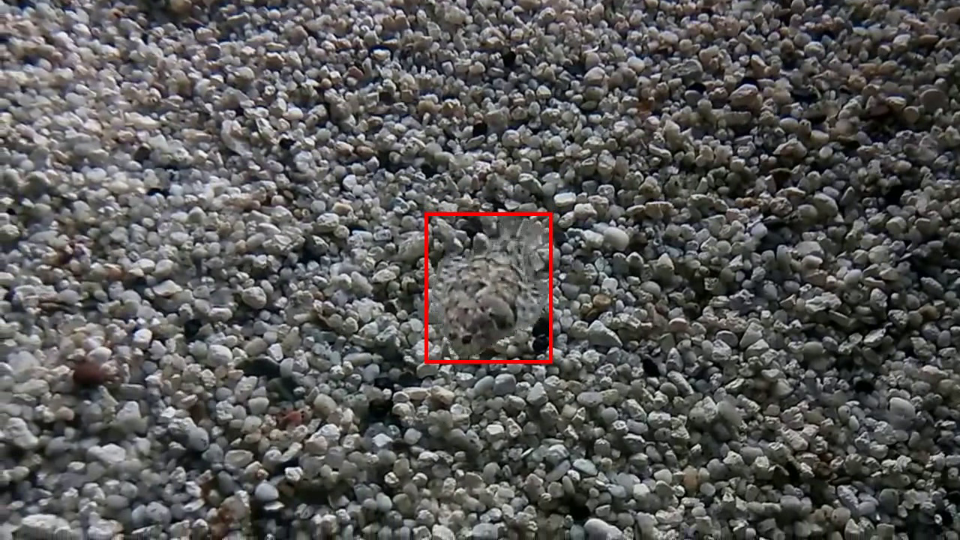}
  \caption{t-th frame.}
  \label{fig:fish1a}
\end{subfigure}%
\hfill
\begin{subfigure}[t]{.32\textwidth}
  \centering
  \includegraphics[width=\linewidth]{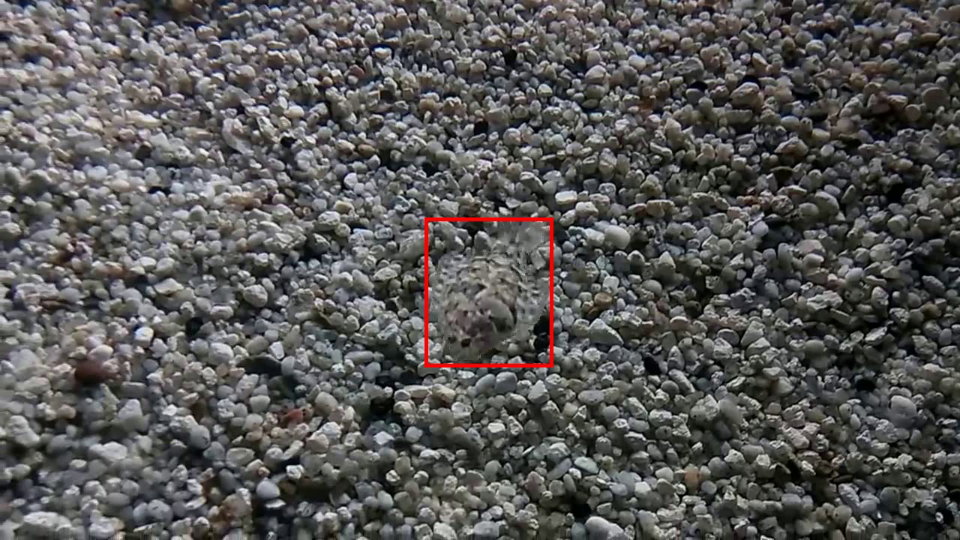}
  \caption{(t+1)-th frame.}
  \label{fig:fish1b}
\end{subfigure}
\hfill
\begin{subfigure}[t]{.32\textwidth}
  \centering
  \includegraphics[width=\linewidth]{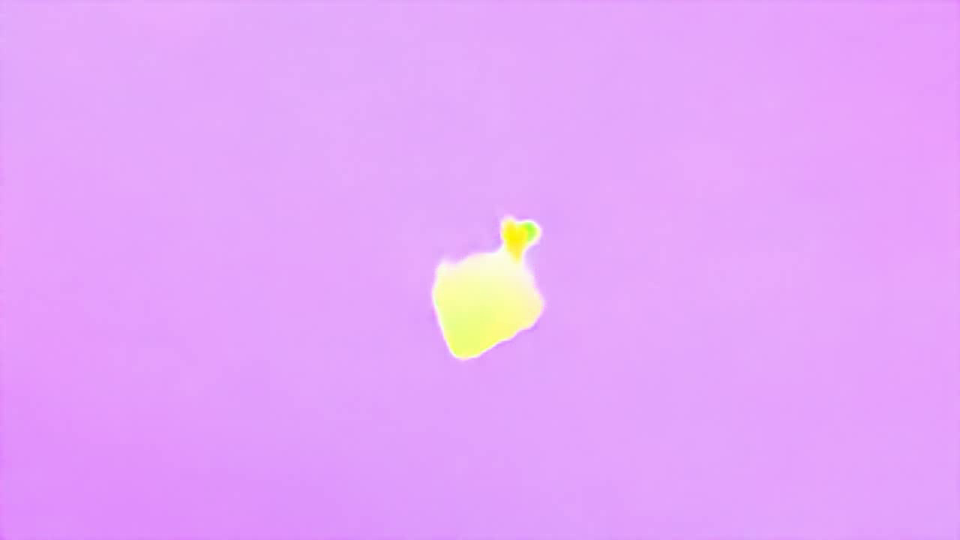}
  \caption{Optical flow.}
  \label{fig:fish1c}
\end{subfigure}
\caption{Two consecutive frames from the camouflage dataset, 
with a bounding box denoting the salient object.
When the object starts moving, even subtly, we are able to detect it more easily, as shown, in the computed optical flow.}
\label{fig:fish}
\end{figure}

We consider a fun yet challenging problem of breaking animal camouflage by exploiting their motion.
Thanks to years of evolution,
animals have developed the ability to hide themselves in the surrounding environment 
to prevent being noticed by their prey or predators.
Consider the example in Figure~\ref{fig:fish1a},
discovering the fish by its appearance can sometimes be extremely challenging, 
as the animal's texture is indistinguishable from its background environment.
However, when the fish starts moving, even very subtly, 
it becomes apparent from the motion, as shown in Figure~\ref{fig:fish1c}.
Having the ability to segment objects both in still images, where this is possible, 
and also from motion, matches well to the two-stream hypothesis in neuroscience. 
This hypothesis suggests that the human visual cortex consists of two different processing streams: 
the ventral stream that performs recognition, 
and the dorsal stream that processes motion~\cite{Goodale92},
providing strong cues for visual attention and detecting salient objects in the scene.

In recent years,
computer vision research has witnessed tremendous progress, 
mainly driven by the ability of learning effective representations for detecting, 
segmenting and classifying objects in {\em still images}.
However, the assumption that objects can be well-segmented by their appearance alone is clearly an oversimplification; that is to say, if we draw an analogy from the two-stream hypothesis, 
the computer vision systems trained on images can only mimic the function of the ventral stream.
The goal of this paper is to develop a computational architecture that is able to process motion representations for breaking camouflage, 
{\em e.g.}~by taking optical flow~(Figure~\ref{fig:fish1c}) as input, 
and predicting a segmentation mask for the animal of interest.

Unfortunately, 
simply relying on motion will not solve our problem completely, 
as, {\em first}, 
optical flow estimation itself remains extremely challenging and is under active research.
In practice, modern optical flow estimation techniques provide a fairly good indication of rough object motion, but not fine-grained details, {\em e.g.}~the exact shape of the objects and their contours.
To compensate for the missing details, 
we propose to use a differentiable registration module for aligning consecutive frames, 
and use the difference of the registered images as auxiliary information to determine the exact contour;
{\em Second},
if the motion stops at  certain points in the video sequence,
then a memory module is required to maintain the object permanence, as is done in~\cite{Tokmakov19},
{\em i.e.}~to capture the idea that objects continue to exist even when they cannot be seen explicitly in the motion representation.

Another main obstacle encountered when addressing the challenging task of camouflage breaking is the limited availability of benchmarks to measure progress. 
In the literature,
there is a Camouflaged Animals video dataset, released by Bideau {\em et al.}~\cite{Bideau16a}, 
but this has only 9 clips on 6 different kinds of animals~(about 840 frames).
To overcome this limitation, 
we collect a {\bf Mo}ving {\bf C}amouflaged {\bf A}nimal dataset, termed {\bf MoCA}, 
which consists of $141$ video sequences~(37K frames), 
depicting 67 kinds of camouflaged animals moving in natural scenes.
Both temporal and spatial annotations are provided in the form of tight bounding boxes on every $5${th} frame and the rest are linearly interpolated. 

To summarize, in this paper, we make the following contributions:
{\em First}, 
we propose a novel architecture with two essential components for breaking camouflage, 
namely,
a differentiable registration module to align the background~(regions other than the camouflaged animal) of consecutive frames, 
which effectively highlights the object boundary in the difference image, 
and a motion segmentation module with memory that discovers moving objects,
while maintaining the object permanence even when motion is absent at some point.
{\em Second},
we collect the first large-scale video camouflage benchmark~(MoCA),
which we release to the community for measuring progress on camouflage breaking and object tracking.
{\em Third}, we demonstrate the effectiveness of the proposed model on MoCA, 
outperforming the previous video segmentation approaches using motion.
In addition, we also benchmark on DAVIS2016,
achieving competitive performance on the unsupervised segmentation protocol despite using only motion.
Note that, 
DAVIS2016 is fundamentally different from MoCA, 
in the sense that the objects are visually distinctive from the background, 
and hence motion may not be the most informative cue for segmentation.
\section{Related Work}
Our work cuts across several areas of research with a rich literature, 
we can only afford to review some of them here.\\

{\noindent}
{\bf Video Object Segmentation}~\cite{Bideau16a,Ponttuset17,xu18,Brox10,Ochs11,Papazoglou13,Jain17,Tokmakov19,Dave19,iccv19_stm,cvpr19_feelvos,Vondrick18,Wang19,Lai19,Lai20,tpami18_osvos-s,bmvc17_OnAVOS,cvpr17_OSVOS,Fragkiadaki12,Keuper15,Yang19,Lu_2019_CVPR} 
refers to the task of localizing objects in videos with pixel-wise masks.
In general, two protocols have recently attracted an increasing interest from the vision community~\cite{Ponttuset17,xu18},
namely unsupervised video object segmentation~({\bf unsupervised VOS}), 
and semi-supervised video object segmentation~({\bf semi-supervised VOS}).
The former aims to automatically separate the object of interest~(usually the most salient one) 
from its background in a video sequence;
and the latter aims to re-localize one or multiple targets that are specified in the first frame of a video with pixel-wise  masks.
The popular methods to address the unsupervised VOS have extensively relied on a combination of appearance and motion cues,
{\em e.g.}~by clustering trajectories~\cite{Brox10,Ochs11}, or by using two stream networks~\cite{Papazoglou13,Jain17,Tokmakov19,Dave19}; or have purely used  appearance~\cite{Wang19,Yang19,Jun17,Fan19}.
For semi-supervised VOS,
prior works can roughly be divided into two categories,
one is based on mask propagation~\cite{iccv19_stm,cvpr19_feelvos,Vondrick18,Wang19,Lai19,Lai20},
and the other is related to few shot learning or online adaptation~\cite{tpami18_osvos-s,bmvc17_OnAVOS,cvpr17_OSVOS}.\\

{\noindent}{\bf Camouflage Breaking}~\cite{Bideau16a,Le19} 
is closely related to the unsupervised VOS, 
however, it poses an extra challenge,
as the object's appearance from {\em still image} can rarely provide any evidence for segmentation, {\em e.g.}~boundaries.
As such, the objects or animals will only be apparent when they start to move. 
In this paper, 
we are specifically interested in this type of problem, 
{\em i.e.}~breaking the camouflage in a class-agnostic manner, 
where the model takes no prior knowledge of the object's category, shape or location,
and is asked to discover the animal with pixel-wise segmentation masks whenever they move.\\

{\noindent}{\bf Image Registration/Alignment} is a long-standing vision problem with 
the goal of transferring one image to another with as many pixels in correspondence as possible.
It has been applied to numerous applications such as video stabilization, 
summarization, and the creation of panoramic mosaics. 
A comprehensive review can be found in~\cite{Hartley04c,szeliski2004image}.
In general, the pipeline usually involves both correspondence estimation 
and transformation estimation.
Traditionally, the alignment methods apply hand-crafted features, 
{\em e.g.}~SIFT~\cite{Lowe99}, for keypoint detection and matching in a pair of images, 
and then compute the transformation matrix by solving a linear system.
To increase the robustness of the geometric transformation estimation, 
RANdom SAmple Consensus (RANSAC)~\cite{Fischler81} is often adopted.
In the deep learning era, 
researchers have constructed differentiable architectures that enable end-to-end optimization for the entire pipeline.
For instance, \cite{Brachmann17} proposed a differentiable RANSAC by relaxing the sparse selection with a soft-argmax operation.
Another idea is to train a network with binary classifications on the inliers/outliers correspondences
using either ground truth supervision or a soft inlier count loss, 
as in~\cite{Brachmann18,Ranftl18,Rocco18,Brachmann19},
and solve the linear system with weighted least squares.
\section{A video registration and segmentation network}
\begin{figure}[!htb]
    \centering
    \includegraphics[width=1\textwidth]{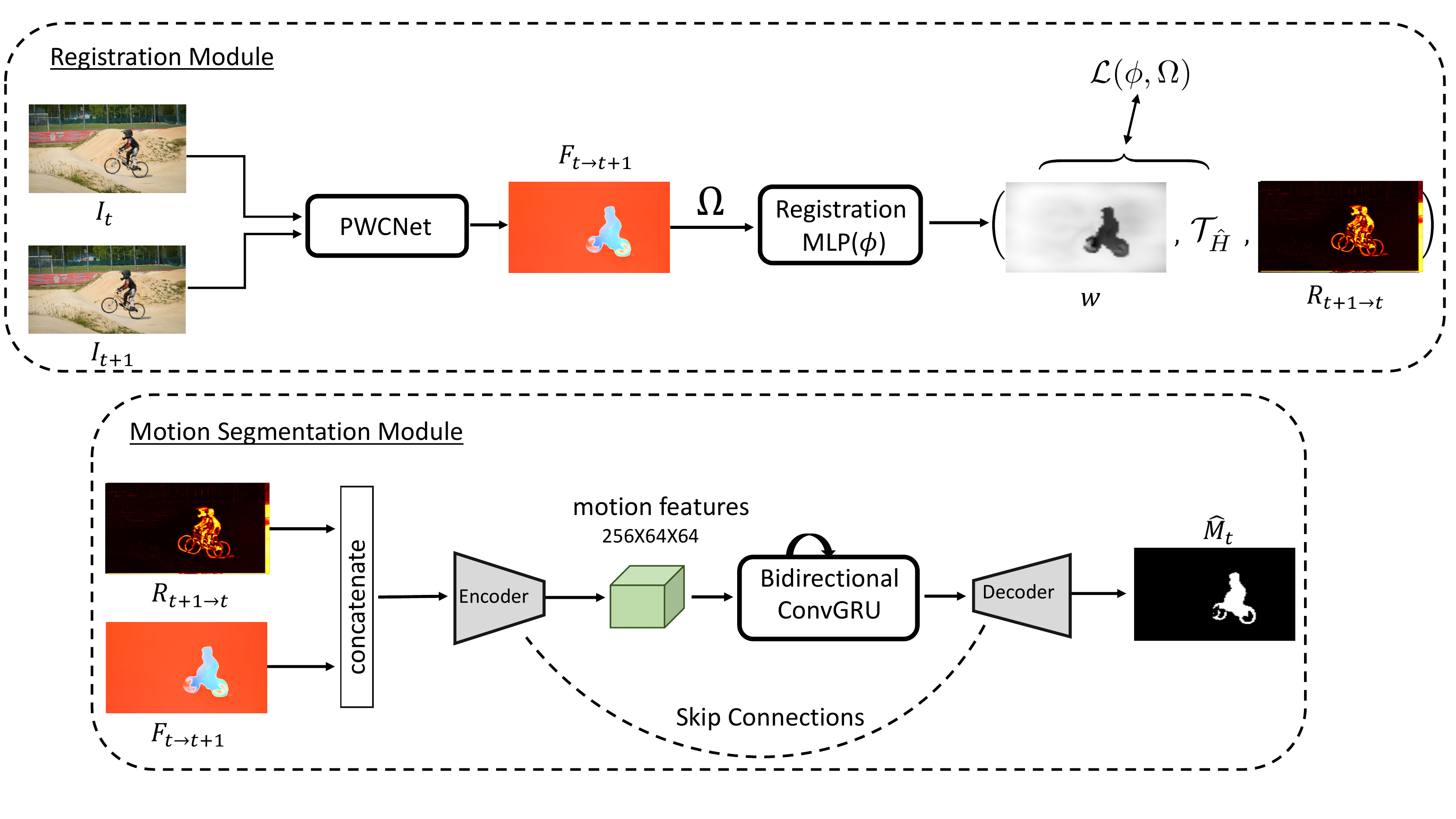}
    \caption{Architecture Overview. 
    The proposed architecture is composed of two different modules, 
    namely registration and motion segmentation.}
    \label{fig:architecture}
\end{figure}

At a high level, we propose a novel computational architecture for breaking animal camouflage, 
which {\em only} considers motion representation as input, {\em e.g.}~optical flow,
and produces the segmentation mask for the moving objects.
Specifically, as shown in Figure~\ref{fig:architecture}, the model consists of two modules: 
(i) a differentiable registration module for aligning consecutive frames, 
and computing the {\em difference} image to highlight the fine-grained detail of the moving objects, 
{\em e.g.}~contours and boundaries~(Section~\ref{sec:diff_regist});
and (ii) a motion segmentation network, 
which takes optical flow together with the difference image from the registration as input, 
and produces a detailed segmentation mask~(Section~\ref{sec:network}).

\subsection{Motion Representation}
\label{sec:optical_flow}
In this paper, we utilise optical flow as a representation of motion.
Formally, consider two frames in a video sequence, $I_t$ and $I_{t+1}$, 
each of dimension $\mathbb{R}^{H\times W\times 3}$,
the optical flow is defined as the displacement field $F_{t\rightarrow t+1} \in \mathbb{R}^{H\times W\times 2}$ 
that maps each pixel from $I_t$ to a corresponding one in $I_{t+1}$, such that:
\begin{equation}
I_{t}(\mathbf{x}) = I_{t+1}(\mathbf{x}+F_{t\rightarrow  t+1}(\mathbf{x})),
\label{eq:flow2}
\end{equation}
where $\mathbf{x}$ represents the spatial coordinates $(x,y)$ and $F$ represents the vector flow field in both horizontal and vertical directions. 
In practice, we use the pretrained PWCNet~\cite{Sun18} for flow estimation,
some qualitative examples can be found in Figure~\ref{fig:fish} and Figure~\ref{fig:RegNet}.

 \subsection{Differentiable Registration Module}
\label{sec:diff_regist}
One of the main challenges of segmentation with optical flow
is the loss of rich fine-grained details due to motion approximations.
In order to recover the sharp contours of the objects under motion, 
we seek a low level RGB signal which ideally suppresses the background and highlights the object's boundaries.
In this paper, 
we use the image difference between consecutive frames after camera motion compensation. 
To this end, a reasonable assumption to make is that 
the foreground object undergoes an independent motion with respect to the global transformation of the background. 
We propose a differentiable registration module for estimating this transformation, 
which we approximate with a homography~($\mathcal{T}_{\hat{H}}$) between consecutive frames, 
and then compute the {\em difference} image after alignment~($R_{t+1 \rightarrow t} = |\mathcal{T}_{\hat{H}}(I_{t+1}) - I_{t}|$), which will provide cues for the animal contours.

The key here is to train a registration module that accepts a set of correspondences obtained from the consecutive frames,
outputs an homography transformation matrix~($H$), and an inlier weight map~($w$), 
which, ideally, acts like a RANSAC process, 
and has $1$'s for every background pixel,
and $0$'s for every foreground pixel~(moving object's).
In this paper, 
we parametrize the registration module with Multiple Layer Perceptrons (MLPs),
{\em i.e.}~$[H, w] = \phi(\{p_s, p_t\}; \theta_r)$,
where $p_s \in \mathcal{R}^{mn \times 2}$ denotes the spatial coordinates of all pixels~(normalized within the range $[-1, 1]$) in the source image, 
and their corresponding position in the target image~($p_t \in \mathcal{R}^{mn \times 2}$), 
based on the computed optical flow, and $\theta_r$ are the trainable parameters.

\subsubsection{Homography Transformation.}
In order to be self-contained, we summarise here the homography computation. 
Mathematically, 
a homography transformation~($\mathcal{T}_{H}$) maps a subset of points $\mathcal{S}_{s}$ from the source image  to a subset of points $\mathcal{S}_{t}$ in the target image;
in our case, the source and target images refer to $I_{t+1}$ and $I_t$ respectively: 

\begin{equation}
\ \forall {p_{i}}^{s} \in \mathcal{S}_{s},\   \exists  {p_{i}}^{t} \in \mathcal{S}_{t} \ \ \  {p_{i}}^{t}  = \mathcal{T}_{H}({p_{i}}^{s}) = \alpha_{i} H {p_{i}}^{s},
\end{equation}
where $H$ is the matrix associated with the homography transformation $\mathcal{T}_{H}$ with $8$ degrees of freedom, and $\alpha_{i}$ a non-zero scalar. 
This formulation can be expressed using homogeneous coordinates of ${p_{i}}^{s} $ and ${p_{i}}^{t} $ as:
\begin{equation}
\label{eqn:homogeneous}
\begin{pmatrix}
{x_{i}}^{t}\\
{y_{i}}^{t}\\
1
\end{pmatrix}  = \alpha_{i} H  \begin{pmatrix}
{x_{i}}^{s}\\
{y_{i}}^{s}\\
1
\end{pmatrix}.
\end{equation}
Using the standard Direct Linear Transform (DLT) \cite{Hartley04a}, 
the previous equation can be written as:
\begin{equation}
A \ vec(H) = \mathbf{0}, 
\label{eqn:homogeneous1}
\end{equation}
where  $vec(H)=\begin{pmatrix} h_{11} & h_{12} & h_{13} & h_{21}&h_{22} & h_{23}& h_{31} & h_{32}&  h_{33} \end{pmatrix}^{T} $ is the vectorised homography matrix and $A$ the data matrix.
The homography $H$ can therefore be estimated by solving such over-complete linear equation system. For more details on the DLT computation, \rev{refer to~\ref{ap:homography_estim}}.

\subsubsection{Training objective.}
\label{ref:regloss}
In order to train the registration module~($\phi(\cdot; \theta_r)$),
we can optimize: 
\begin{align}
    \mathcal{L} = \frac{1}{\sum w} \sum_{\Omega} w \cdot ||\mathcal{T}_H(p_s) - p_t||_2  + R(w),
\end{align}
where $\Omega$ refers to all the pixels on the $mn$ grid.
Note that, the homography $\mathcal{T}_{H}$ transformation in this case can be solved with a simple 
weighted least square~(WLS) and differentiable SVD~\cite{Ranftl18} for parameter updating.
To avoid trivial solution, where the weight map can be full of zeros that perfectly minimize the loss,
we add a regularization term~($R(w)$), 
that effectively encourages as many inliers as possible:
\begin{align}
    &R(w) = -\gamma \sum_{p \in \Omega} l_p - \frac{1}{mn} \sum_\Omega \big\{ l_p \cdot \text{log}(w) + (1- l_p) \cdot \text{log}(1-w) \big\}, \\
    & \text{where \hspace{8pt}} l_p =\sigma \{(\epsilon - ||\mathcal{T}_H(p_s) - p_t||_2) / \tau\}.  \nonumber
\label{eq:reg}
\end{align}
In our training, $\gamma=0.05$, $\tau=0.01$, $\epsilon = 0.01$ and $m=n=64$. 
The first term in $R(w)$~($l_p$) refers to a differentiable inlier counting~\cite{Brachmann18,Rocco18}.
The rest of the terms aim to minimize the binary cross-entropy at each location of the inlier map, 
as in~\cite{Moo18}, 
forcing the predictions to be classified as inlier~(1's) or outlier~(0's).

\subsection{Motion Segmentation Module~}
\label{sec:network}
After introducing the motion representation and registration, 
we consider a sequence of frames from the video, 
$I \in \mathcal{R}^{T \times 3 \times H \times W}$, 
where the three channels refer to a concatenation of the flow~(2 channels) and difference image~(1 channel).
For simplicity, 
here we use a variant of UNet~\cite{Ronneberger15} with the bottleneck feature maps being recurrently processed.

Specifically, the {\bf Encoder} of the segmentation module will independently process the current inputs,
ending up with motion features of $\mathcal{R}^{T \times 256 \times 64 \times 64}$,
where $T, 256, 64, 64$ refer to the number of frames, number of channels, height, and width respectively.
After the Encoder, 
the  {\bf memory module}~(a bidirectional convGRU~\cite{Ballas16} is used in our case) 
operates on the motion features, 
updating them by aggregating the information from time steps in both directions.
The {\bf Decoder} takes the updated motion features, 
and produces an output binary segmentation mask, {\em i.e.}~foreground vs background. 
The Motion Segmentation Module is trained with pixelwise binary cross-entropy loss. 

This completes the descriptions of the two individual modules used in the proposed architectures. Note that the entire model is trained together as it is end-to-end differentiable.
\section{MoCA: a new Moving Camouflaged Animal dataset}
\label{sec:moca}
One of the main obstacles encountered when addressing the challenging task of camouflage breaking is the limited availability of datasets. A comparison of existing datasets in Table~\ref{tab:compare_stat}.
Bideau {\em et al.} published the first Camouflaged Animals video dataset with only 9 clips,
and Le {\em et al.} proposed the CAMO dataset with only single image camouflage,
and therefore not suitable for our video motion segmentation problem.

To overcome this limitation, 
we collect the {\bf Mo}ving {\bf C}amouflaged {\bf A}nimal dataset, termed {\bf MoCA}, 
which consists of $141$ video sequences depicting various camouflaged animals moving in natural scenes. 
We include the PWC-Net optical flow for each frame and provide both spatial annotations in the form of bounding boxes and motion labels every $5${-th} frame with the rest  linearly interpolated. 
\subsection{Detailed statistics}
\begin{table}[t]
\caption{Statistics for recent camouflage breaking datasets; 
``\# Clips'' denotes the number of clips in the dataset; 
``\# Images'' denotes the number of frames or images in the dataset; 
``\# Animals'' denotes the number of different animal categories in the dataset;
``Video'' indicates whether videos are available.}
\setlength{\tabcolsep}{8pt}
\footnotesize
\centering
\begin{tabular}{|c|c|c|c|c|c|c|}
\hline
Datasets 			 		& \# Clips & \# Images & \# Animals  & Video     \\ \hline
Camouflaged Animals~\cite{Bideau16} & $9$	   & $839$ 	& $6$   	    & $\checkmark$        \\ 
CAMO~\cite{Le19}& $0$     & $1250$   	& $80$    	    & $\times$         \\ 
\textbf{MoCA~(Ours)}	& $141$   & $37$K 	& $67$ 	    & $\checkmark$  \\  \hline
\end{tabular}
\label{tab:compare_stat}
\end{table}
MoCA contains 141 video sequences
collected from YouTube, at the highest available resolution, mainly $720 \times 1280$, and sampled at $24 fps$. A total of 
 37250 frames spanning 26 minutes. 
Each video represents a continuous sequence depicting one camouflaged animal, 
ranging from 1.0 to 79.0 seconds.
\begin{figure}[!htb]
\centering
\footnotesize
\begin{subfigure}{.40\textwidth}
  \centering
  \includegraphics[width=\textwidth]{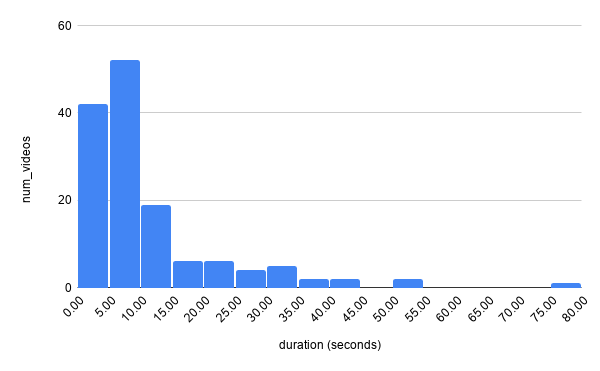}
  \caption{Distribution of video lengths.}
  \label{fig:stat1}
\end{subfigure}
\begin{subfigure}{.56\textwidth}
  \centering
  \includegraphics[width=\textwidth]{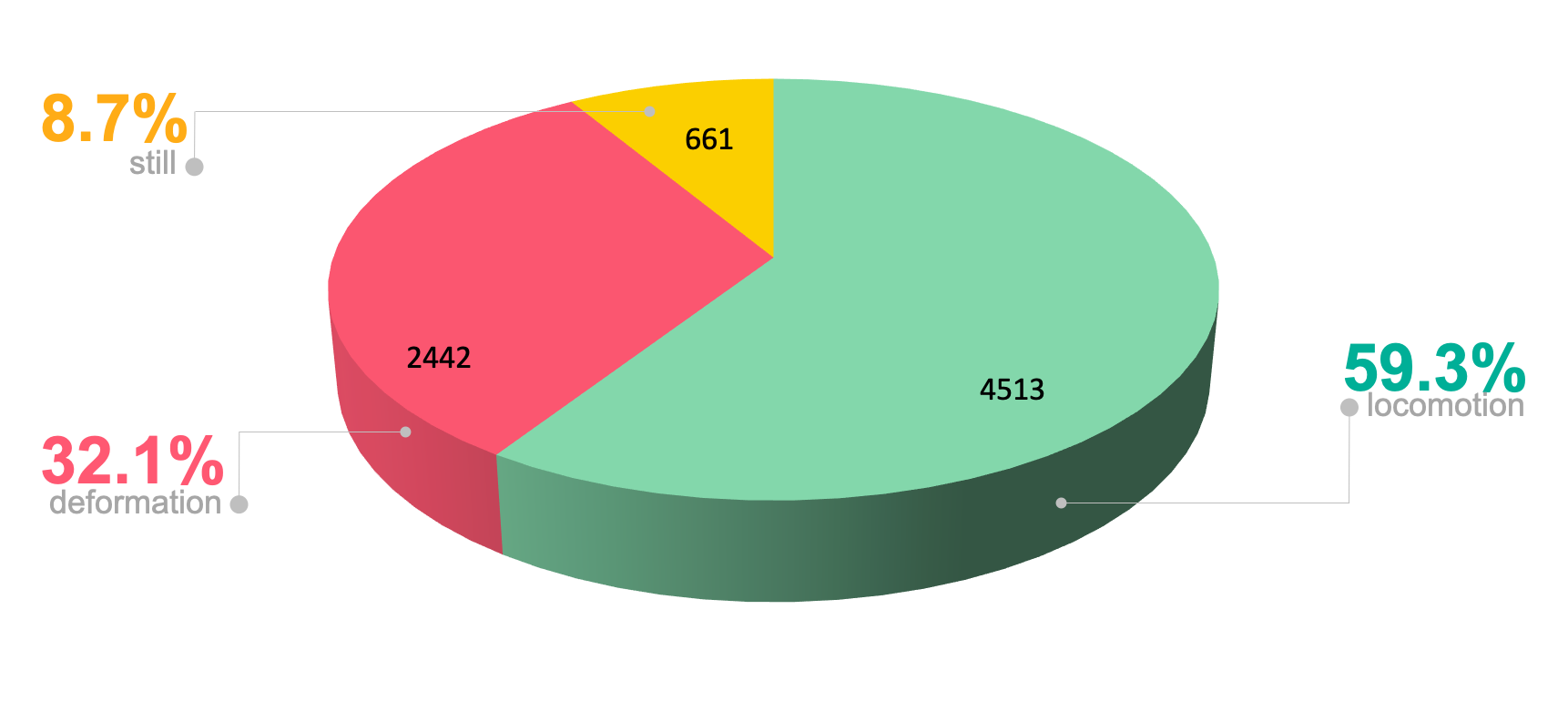}
  \caption{Distribution of motion labels.}
  \label{fig:stat2}
\end{subfigure}
\caption{Statistics for the Moving Camouflaged Animals~(MoCA) dataset. In (a), the  x-axis denotes the video duration, and the y-axis denotes the number of video sequences. (b) the distribution of frames according to their motion types (still, deformation and locomotion), see text for the detailed definitions.}
\end{figure}
\begin{figure}[!h]
    \centering
    \includegraphics[width=\textwidth]{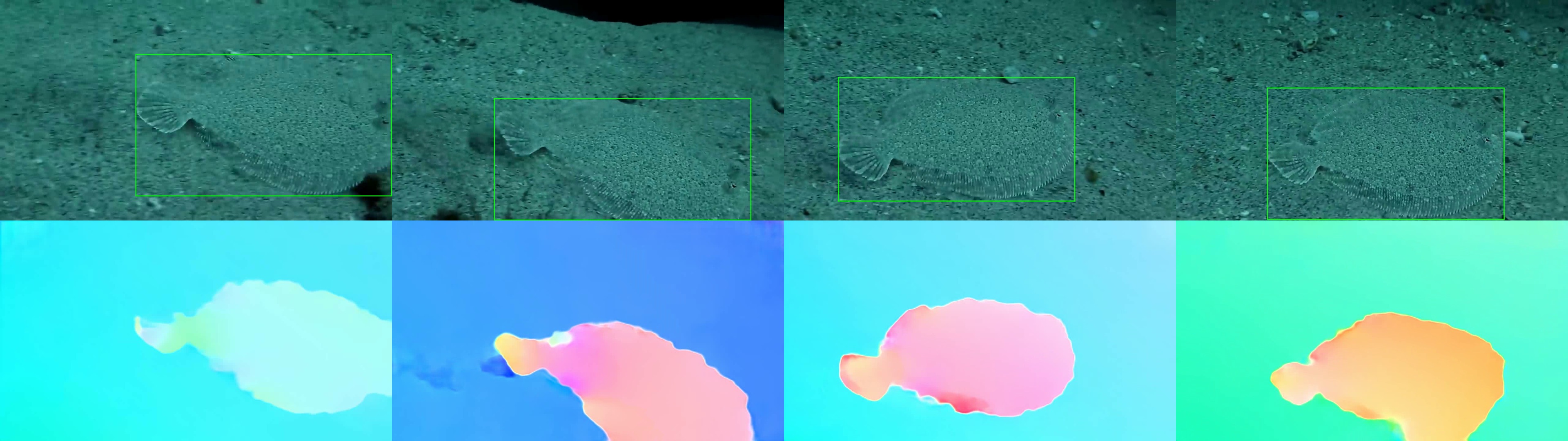}
    \includegraphics[width=\textwidth]{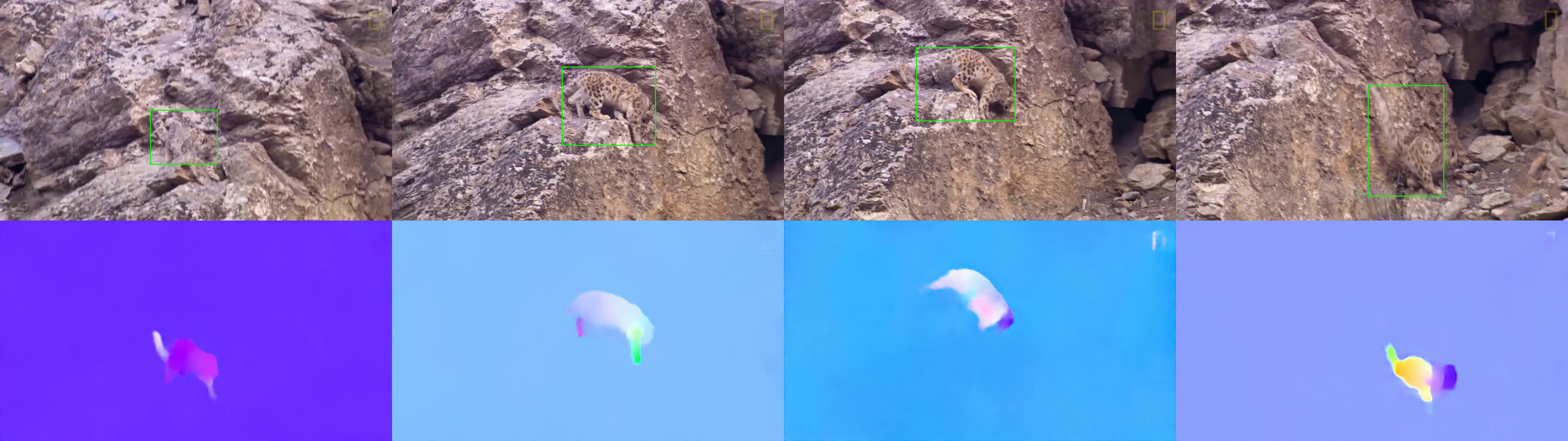}
    \caption{Example sequences from the Moving Camouflaged Animals~(MoCA) dataset 
             with their corresponding optical flows.}
    \label{fig:demo}
\end{figure}
\noindent
The distribution of the video lengths is shown in Figure \ref{fig:stat1}.
The dataset is labelled with a bounding box in each frame, 
as well as a motion label for the type of motion.
\noindent
While annotating the data, we distinguish three types of motion:
\begin{itemize}
\setlength\itemsep{.5em}
\item \textbf{Locomotion:} when the animal engages in a movement that leads to a significant change of its location within the scene {\it e.g.} walking, running, climbing, flying, crawling, slithering, swimming, 
{\em etc.}
\item \textbf{Deformation:} when the animal engages in a more delicate movement that only leads to a change in its pose while remaining in the same location {\it e.g.} moving a part of its body.
\item \textbf{Static:} when the animal remains still. 
\end{itemize}
\noindent
As shown in Figure~\ref{fig:stat2}, motion wise, 
the dataset contains 59.3\% locomotion, 32.1\% deformations, and 8.7\% still frames.
Examples from the dataset are shown in Figure~\ref{fig:demo}.
\section{Experiments}
In this section, 
we detail the experimental setting used in this paper, including
the datasets, evaluation metrics, baseline approaches and training details.

\subsection{Datasets}
\label{sec:datasets}
{\noindent}{\bf DAVIS2016 } 
refers to the Densely Annotated VIdeo Segmentation dataset~\cite{Perazzi16}. 
It consists of 50 sequences, 
30 for training and 20 for testing, 
captured at $24fps$ and provided at two resolutions. 
We use the $480p$ version in all experiments. 
This dataset has the advantage that accurate pixelwise ground truth segmentations are provided, 
$3,455$ annotations in total,
as well as spanning a variety of challenges, such as occlusions, fast-motion, 
non-linear deformation and motion-blur. 
We train the model on the DAVIS 2016 training set and report results on its validation splits.\\

{\noindent}{\bf Synthetic Moving Chairs. }
\label{data:geometry_moving_chairs}
In order to train the differentiable registration module properly, 
we use video sequences that are synthetically generated.
Specifically, 
we use the 3D-rendered objects from the Flying Chairs dataset as foreground,
and take random images from YouTube-VOS as background.
We then apply rigid motions, {\em e.g.}~homographies, 
to the background, and simulate an independent motion for the foreground object. 
Note that, with such a synthetic dataset,
we have complete information about the background homography transformation, optical flow, inlier maps, and object masks,
which enables us to better initialise the registration module before training on real video sequences.
These synthetic video sequences were only used to pre-train the registration and motion segmentation modules. 
\rev{We include example images in~\ref{ap:synthetic_chairs}}.\\

{\noindent}{\bf Evaluation Metrics. }
\label{sec:metrics}
Depending on the benchmark dataset used, we consider two different evaluation metrics.
For {\bf DAVIS2016}, 
we follow the standard protocol for unsupervised video object segmentation proposed in~\cite{Perazzi16}, 
namely, the mean region similarity $\mathcal{J}$,
which is the intersection-over-union~(IOU) of the prediction and ground truth; 
and mean contour accuracy $\mathcal{F}$, 
which is the F-measure defined on contour points from the prediction and the ground truth. 
For {\bf MoCA}, 
as we only have the bounding box annotations, 
we define the metric as the IOU between the ground truth box and the minimum box that includes the predicted segmentation mask.
Note that, we follow the same protocol used in Bideau~{\em et al.}~\cite{Bideau16a}, 
meaning, we only evaluate the segmentation of the animals under locomotion or deformation~({\em not} in static frames). 

\subsection{Baselines}
 \label{sec:baselines}
We compare with five previous state-of-the-art approaches~\cite{Tokmakov17,Tokmakov19,Dave19,Yang19,Lu_2019_CVPR}.
In~\cite{Tokmakov19}, 
Tokmakov~{\em et al.}, the LVO method uses a two-stream network for motion segmentation,
where the motion stream accepts optical flow as input via a MPNet~\cite{Tokmakov17} architecture, 
and the appearance stream uses an RGB image as input.
Similar to ours, a memory module is also applied to recurrently process the frames.
A more recent approach~\cite{Dave19} adapts the Mask-RCNN architecture to motion segmentation, 
by leveraging motion cues from optical flow as a bottom-up signal for separating objects from each other, 
and combines this with appearance evidence for capturing the full objects.
For fair comparison across methods, we use the same optical flow computed from PWCNet for all the flow-based methods. 
To this end, we re-implement the original MPNet~\cite{Tokmakov17},
and train on the synthetic FT3D dataset~\cite{Mayer16}. 
For LVO~\cite{Tokmakov19}, 
we also re-train the pipeline on DAVIS2016.
For Seg-det~\cite{Dave19} and Seg-track~\cite{Dave19}, 
we directly replace the flows from Flownet2~\cite{flownet17} with the ones from PWCNet in the model provided by the authors.
In all cases, our re-implemented models outperform or match the performance reported in the original papers.
In addition, we also compare our method to AnchorDiff~\cite{Yang19} and COSNet~\cite{Lu_2019_CVPR},
both approaches have been trained for unsupervised video object segmentation with only RGB video clips, 
and show very strong performance on DAVIS.

\subsection{Training and Architecture Details }
\label{sec:train_details}
\rev{Here we describe the main modules of our pipeline. More details can be found in~\ref{ap:detailed_architecture}}.\\
\newline 
{\noindent}{\bf Registration Module. }
We adopt the architecture of~\cite{Moo18},
which is MLPs with $12$ layer residual blocks.
We first train the registration module on the Synthetic Moving Chairs sequences described in~\ref{data:geometry_moving_chairs}, 
for $10K$ iterations using an Adam optimizer with a weight decay of $0.005$ and a batch size of $4$. 
For a more stable training, we use a lower learning rate, 
{\em i.e.}~$5 \times 10^{-5}$, avoiding the ill-conditioned matrix in the SVD.\\

{\noindent}{\bf Motion Segmentation Module. }
We adopt a randomly initialized ResNet-18.
Frame-wise segmentation is trained from scratch on the synthetic dataset, 
together with the pre-trained registration module. 
We further include the bidirectional ConvGRU and finetune the whole pipeline on DAVIS 2016, 
with each sequence of length $11$, batch size of 2, for a total of $25K$ iterations. 
For all \rev{training experiments}, we use frames with a resolution of $\mathcal{R}^{256\times256\times3}$.

\section{Results}
In this section, we first describe the performance of our model and previous state-of-the-art approaches on the new MoCA dataset, and then compare segmentation performance on the DAVIS2016 benchmark.
\begin{table}[!htb]
\centering
\caption{Mean Intersection Over Union  on MoCA for the different motion type subsets.
``All\_Motion'' refers to the overall performance}
\setlength{\tabcolsep}{1.5pt}
\footnotesize
\begin{tabular}{|c|c|cccc|ccc|}
\hline
 Input & Model  & RGB & Flow & Register & Memory & Locomotion & Deform & All\_Motion\\ \hline
\multirow{5}{*}{{Flow}}
&MPNet~\cite{Tokmakov17} & $\times$ & $\checkmark$ & $\times$  & $\checkmark$ & 21.3 & \bf 23.5  &22.2 \\
&ours-A& $\times$ & $\checkmark$ & $\times$  & $\checkmark$ & 31.3 & 17.8  & 27.6\\
&ours-B & $\times$ & $\checkmark$ & MLP  & $\times$ &  29.9 & 15.6  &   25.5 \\
&ours-C & $\times$ & $\checkmark$ & MLP  & $\checkmark$ & \bf 47.8 &  20.7  &  \bf 39.4 \\
&ours-D & $\times$ & $\checkmark$ & RANSAC  & $\checkmark$ & 42.9 & 19.2  & 35.8 \\\hline
\multirow{2}{*}{{RGB}}
& AnchorDiff~\cite{Yang19} & $\checkmark$ & $\times$  & $\times$  & $\times$ & 30.9 & 29.4  & 30.4 \\
& COSNet~\cite{Lu_2019_CVPR} & $\checkmark$ & $\times$  & $\times$  & $\times$ & 35.9 &  35.1  & 36.2\\\hline
\multirow{4}{*}{{Both}}
& LVO~\cite{Tokmakov19}  & $\checkmark$ & $\checkmark$ & $\times$  & $\checkmark$ & 30.6 & 34.9  & 30.6\\
& Seg-det~\cite{Dave19} & $\checkmark$ & $\checkmark$ & $\times$  & $\times$ & 16.9 & 18.7  & 17.9\\
& Seg-track~\cite{Dave19}& $\checkmark$ & $\checkmark$ & $\times$  & $\times$ & 29.9 & 32.2  & 30.2\\
& ours-E & $\checkmark$ & $\checkmark$ & MLP  & $\checkmark$ & \bf 45.0 & \bf 38.0  & \bf 42.4\\  \hline
\end{tabular}
\label{tab:perfMoCA}
\end{table}

\subsection{Results on the MoCA Benchmark}
\label{sec:moca_results}
\begin{figure}[!htb]
  \centering
  \begin{subfigure}[]{0.30\textwidth}
  \centering
  \includegraphics[width=1\textwidth]{image/experiments/00010.jpg}
  \includegraphics[width=1\textwidth]{image/experiments/00011.jpg}
   \includegraphics[width=1\textwidth]{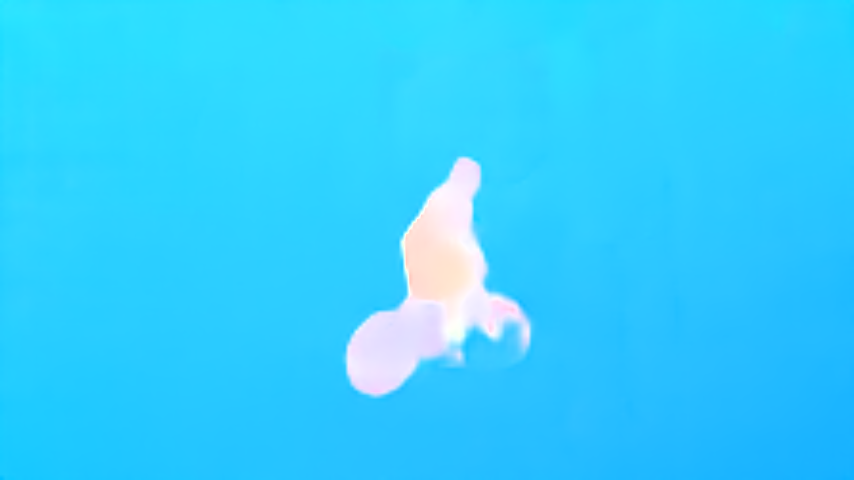}
   \includegraphics[width=1\textwidth]{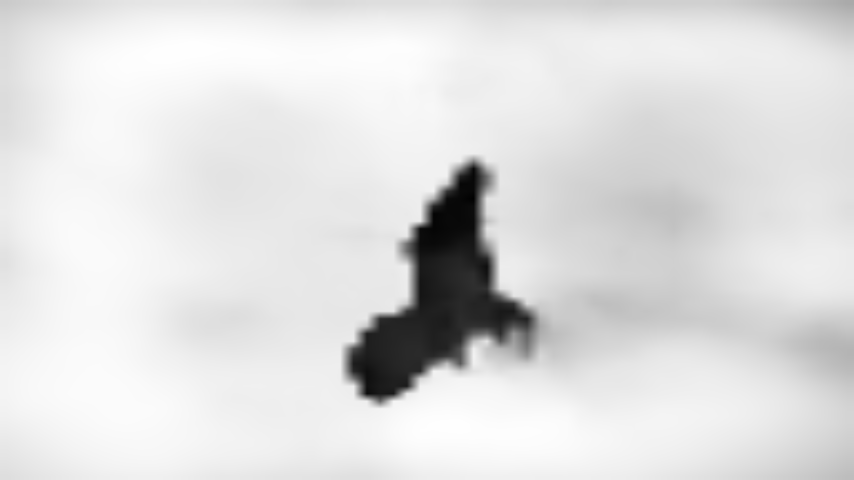}
  \includegraphics[width=1\textwidth]{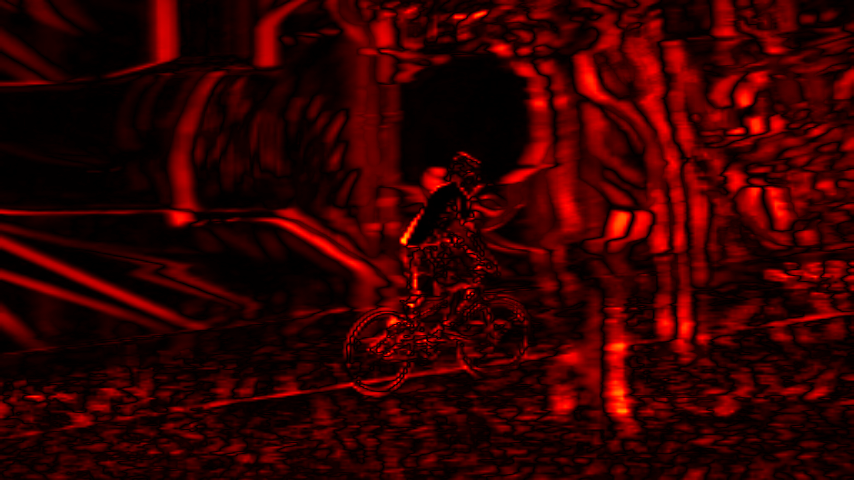}
  \includegraphics[width=1\textwidth]{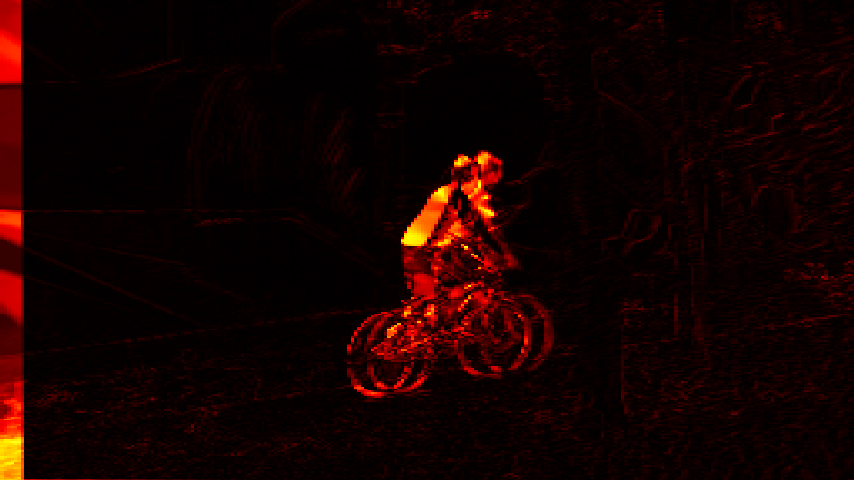} 
  \caption{}
  \end{subfigure}
    \centering
    \begin{subfigure}[]{0.30\textwidth}
  \centering
    \includegraphics[width=1\textwidth]{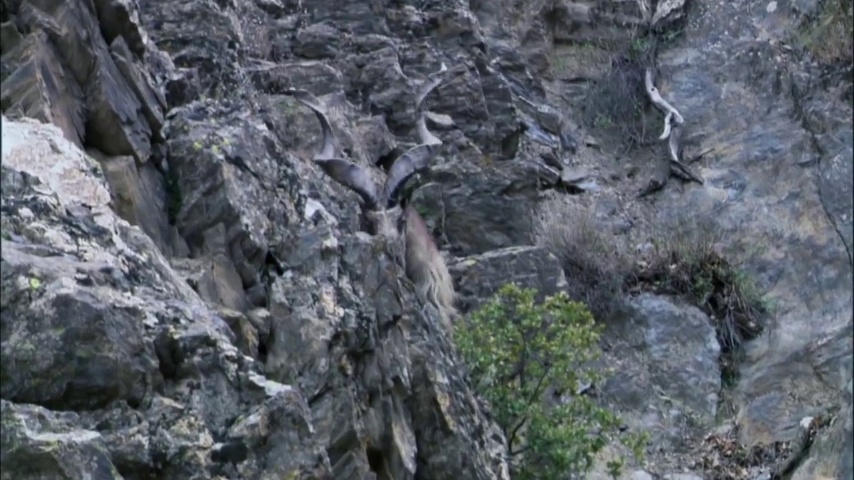}
  \includegraphics[width=1\textwidth]{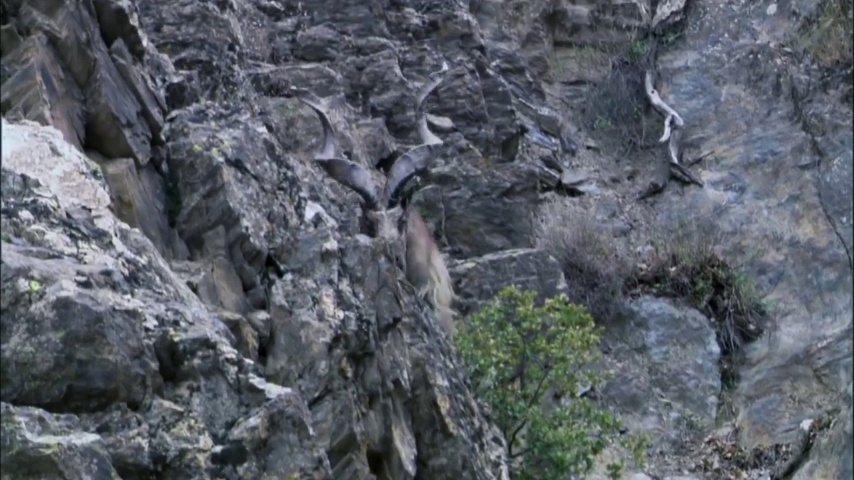}
  \includegraphics[width=1\textwidth]{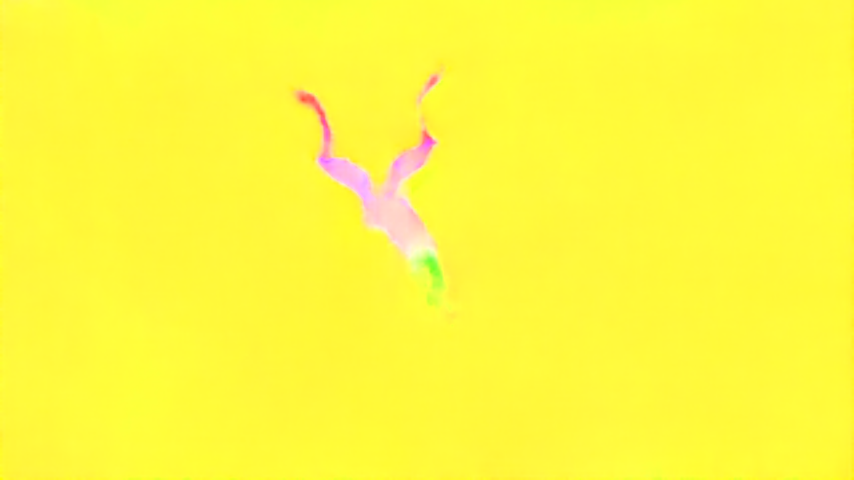}
   \includegraphics[width=1\textwidth]{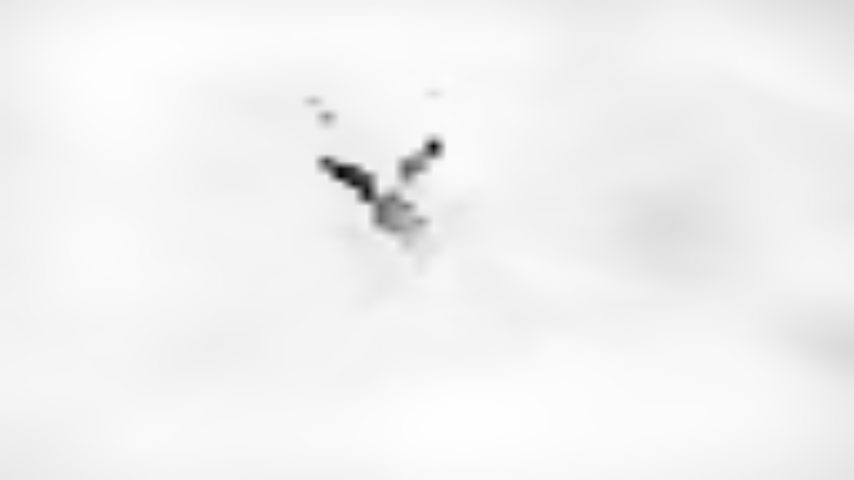}
   \includegraphics[width=1\textwidth]{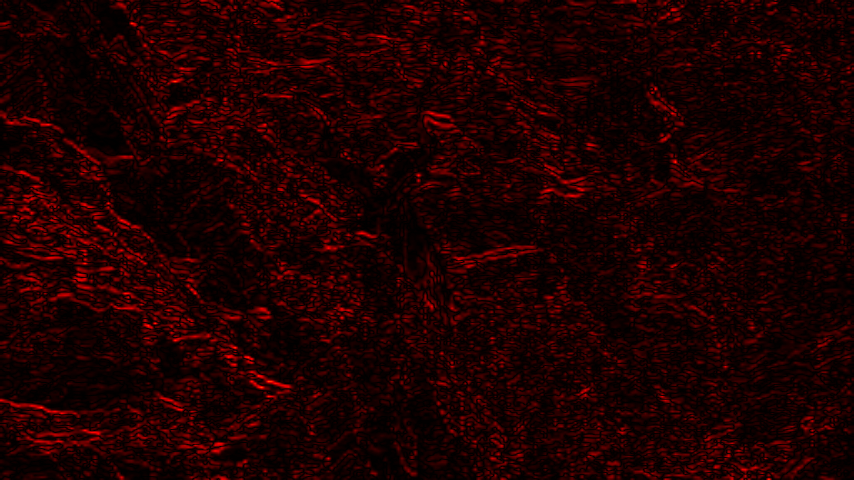}
   \includegraphics[width=1\textwidth]{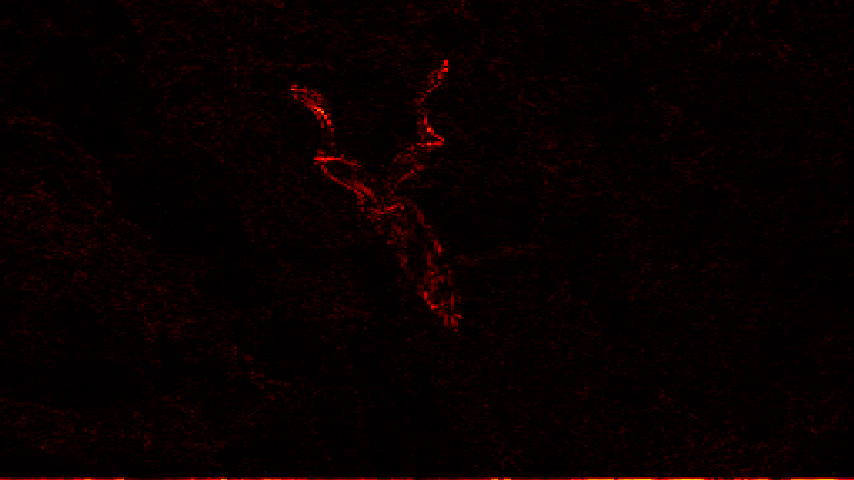} 
     \caption{}
   \end{subfigure}
     \centering
    \begin{subfigure}{0.3\textwidth}
      \centering
    \includegraphics[width=1\textwidth]{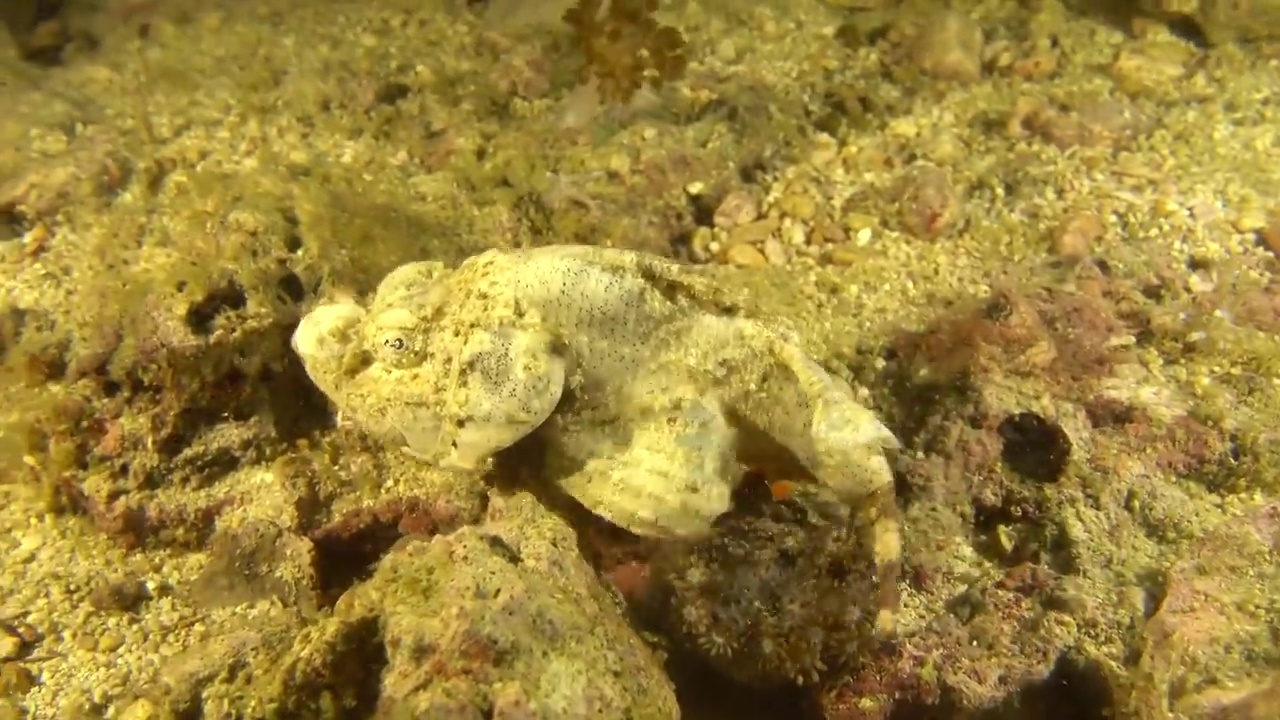}
    \includegraphics[width=1\textwidth]{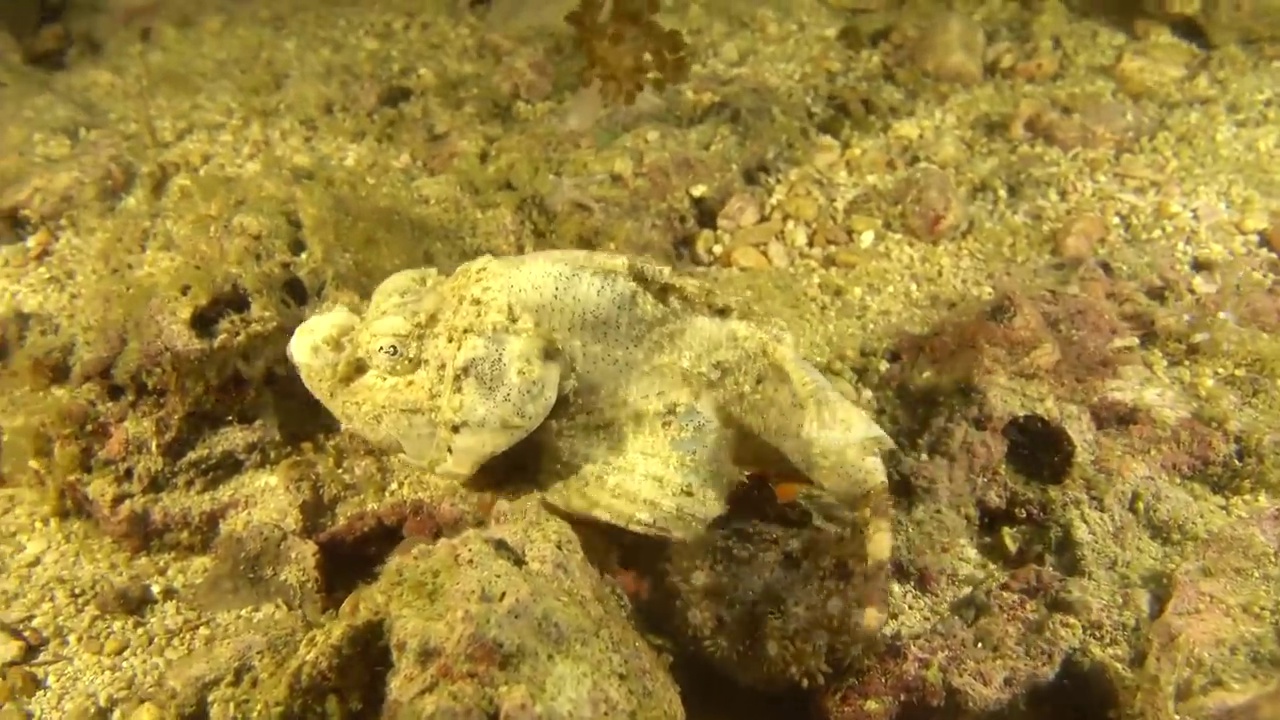}
    \includegraphics[width=1\textwidth]{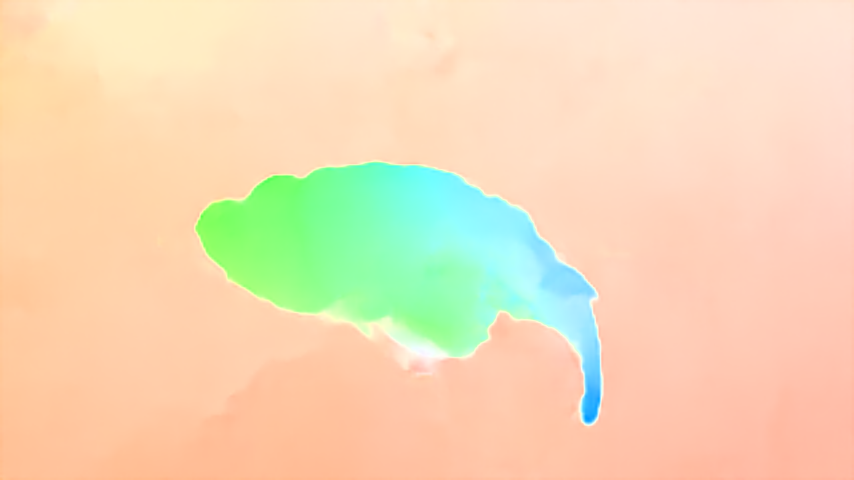}
    \includegraphics[width=1\textwidth]{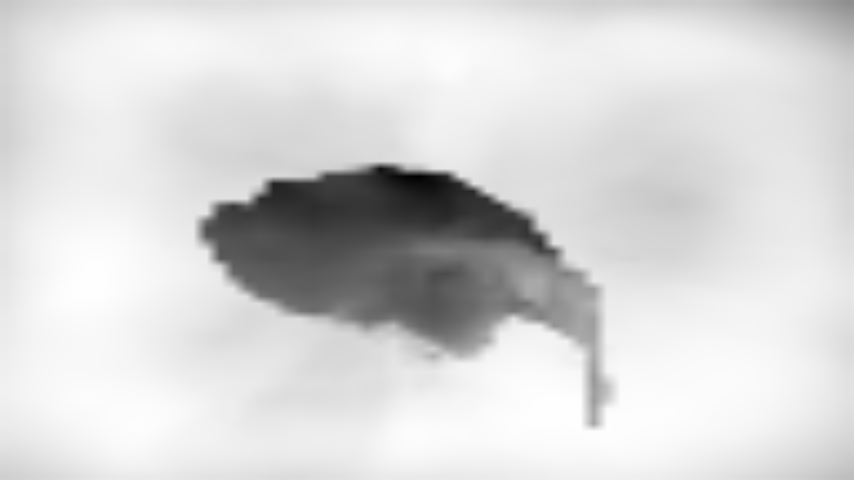}
    \includegraphics[width=1\textwidth]{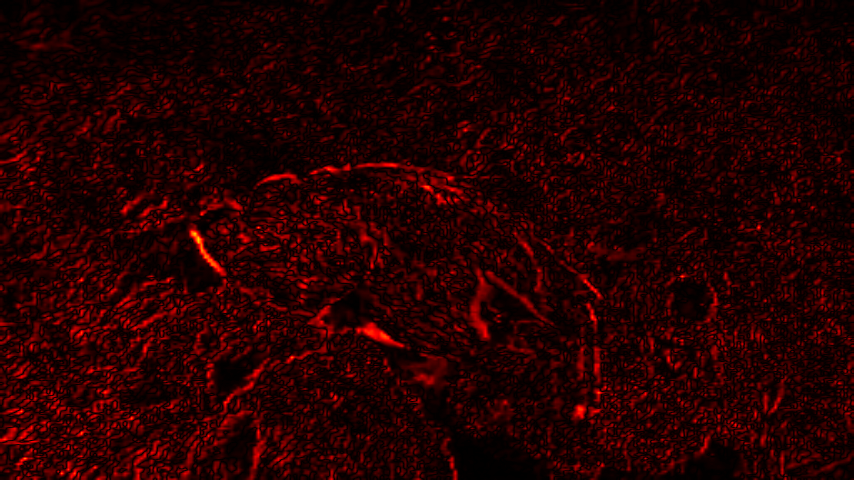}
    \includegraphics[width=1\textwidth]{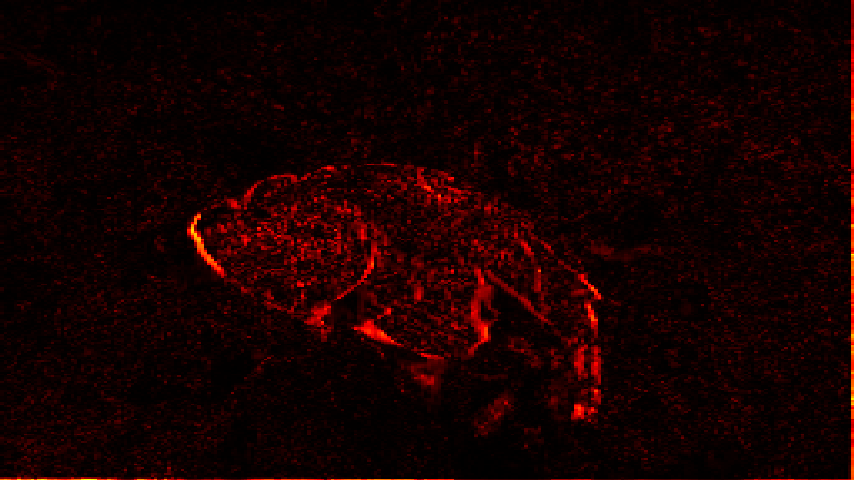}
      \caption{}
  \end{subfigure}
\caption{Registration results from (a) the validation set of DAVIS 2016; and (b)(c) MoCA. 
From top to bottom: Frame $t$, Frame $t+1$, 
Forward PWCNet optical flow, inlier weights~(background pixels are shown as gray or white), 
image difference without alignment, 
aligned image difference.}
\label{fig:RegNet}
\end{figure}

We summarise all the quantitative results in Table~\ref{tab:perfMoCA} and discuss them in the following sections. 
In Figure~\ref{fig:RegNet}, we illustrate the effect of the differentiable registration
and Figure~\ref{fig:resCamo} shows examples of the overall segmentation method. 
\rev{More examples are presented in~\ref{ap:more_qualitative_res}.}\\

{\noindent}{\bf Effectiveness of registration/alignment. }
To demonstrate the usefulness of the differentiable registration module, 
we carry out an ablation study. By comparing ours-A~(without registration), ours-D~(registration using RANSAC) and  Ours-C~(registration using our trainable MLPs), 
it is clear that the model with trainable MLPs for alignment helps to improve both the animal discovery on video sequences with locomotion and deformation, outperforming ours-A~(without registration) and ours-D~(RANSAC).

In Figure~\ref{fig:RegNet} we visualise the results from the registration module,
{\em e.g.}~the inlier map, difference image before alignment~(second last row) and after alignment~(last row).
It is clear that the difference images computed after alignment are able to ignore the background, 
and successfully highlight the boundary of the moving objects, 
{\em e.g.~}the wheels of the bicycle from the first column.\\

{\noindent}{\bf Effectiveness of memory. } 
Comparing model-B~(without memory) and model-C~(with memory), 
the only difference lies on whether the frames are processed individually or recurrently.
As shown by the results, 
a significant boost is obtained with the help of the memory module~(25.5 vs.\ 39.4 on All\_Motion), showing its effectiveness.\\

{\noindent}{\bf Comparison to baselines and previous approaches. }  
From Table~\ref{tab:perfMoCA}, we make the following observations:
{\em First}, 
when comparing with MPNet~\cite{Tokmakov17}, which also processes optical flow,
our model demonstrates a superior performance on All\_Motion~($39.4$ vs $22.2$),
and an even larger gap on Locomotion, 
showing the usefulness of image registration and memory modules;
{\em Second}, 
as expected, 
the state-of-the-art unsupervised video segmentation approaches relying on appearance~(RGB image as input), 
{\em e.g.} AnchorDiff~\cite{Yang19} and COSNet~\cite{Lu_2019_CVPR}, 
tend to struggle on this camouflage breaking task, 
as the animals often blend with the background, 
and appearance then does not provide informative cues for segmentation, emphasising the importance of motion information (by design) in this dataset;
{\em Third}, 
when we adopt a two-stream model, 
{\em i.e.}~extend the architecture with a Deeplabv3-based appearance model,
and naively average the prediction from appearance and flow models,
the performance can be further boosted from $39.4$ to $42.4$,
significantly outperforming all the other two-stream competitors, 
{\em e.g.}~LVO~\cite{Tokmakov19}, Seg-det and Seg-seg~\cite{Dave19}.

\begin{figure}[!htb]
  \centering
    \begin{subfigure}{1\textwidth}
  \centering
    \includegraphics[width=0.24\textwidth]{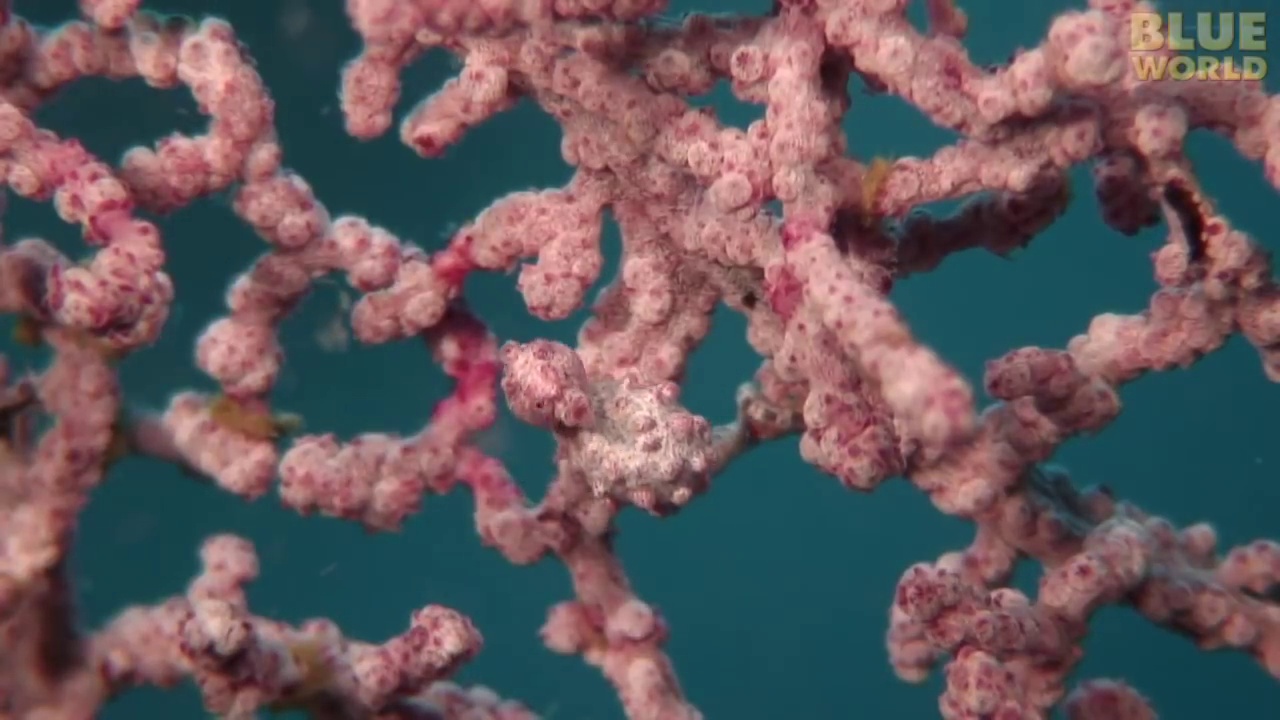}
    \includegraphics[width=0.24\textwidth]{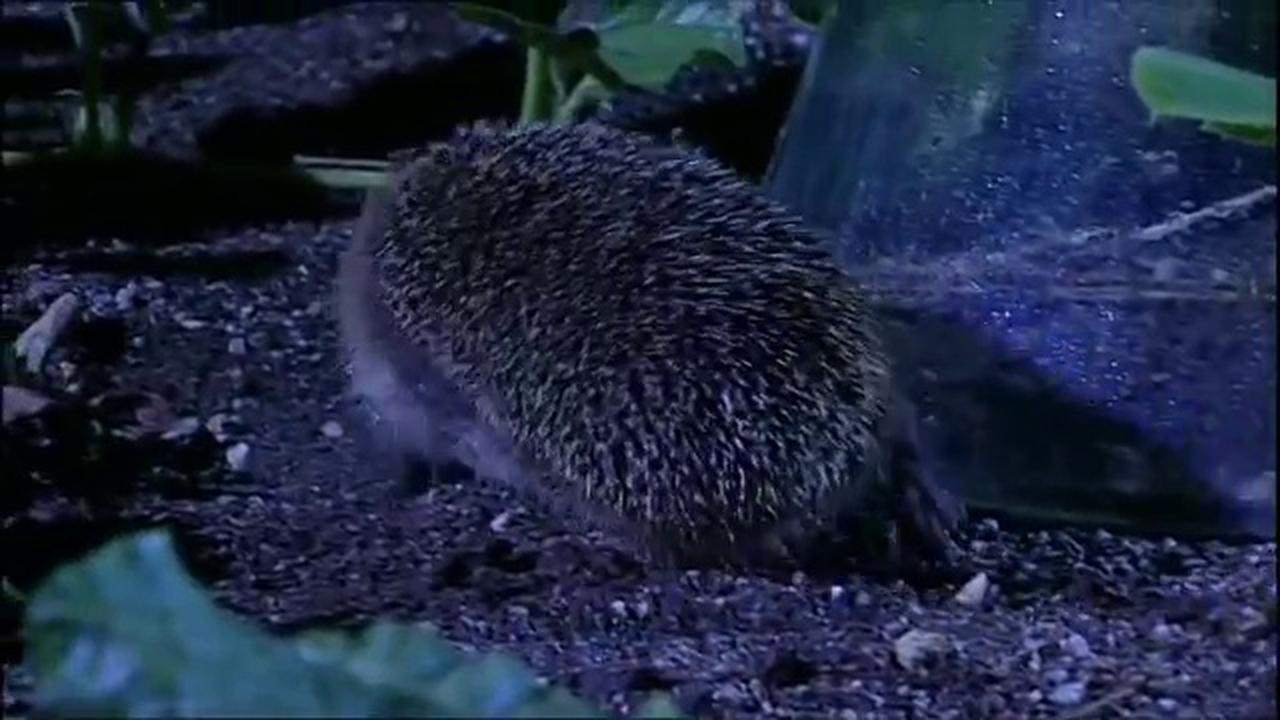}
    \includegraphics[width=0.24\textwidth]{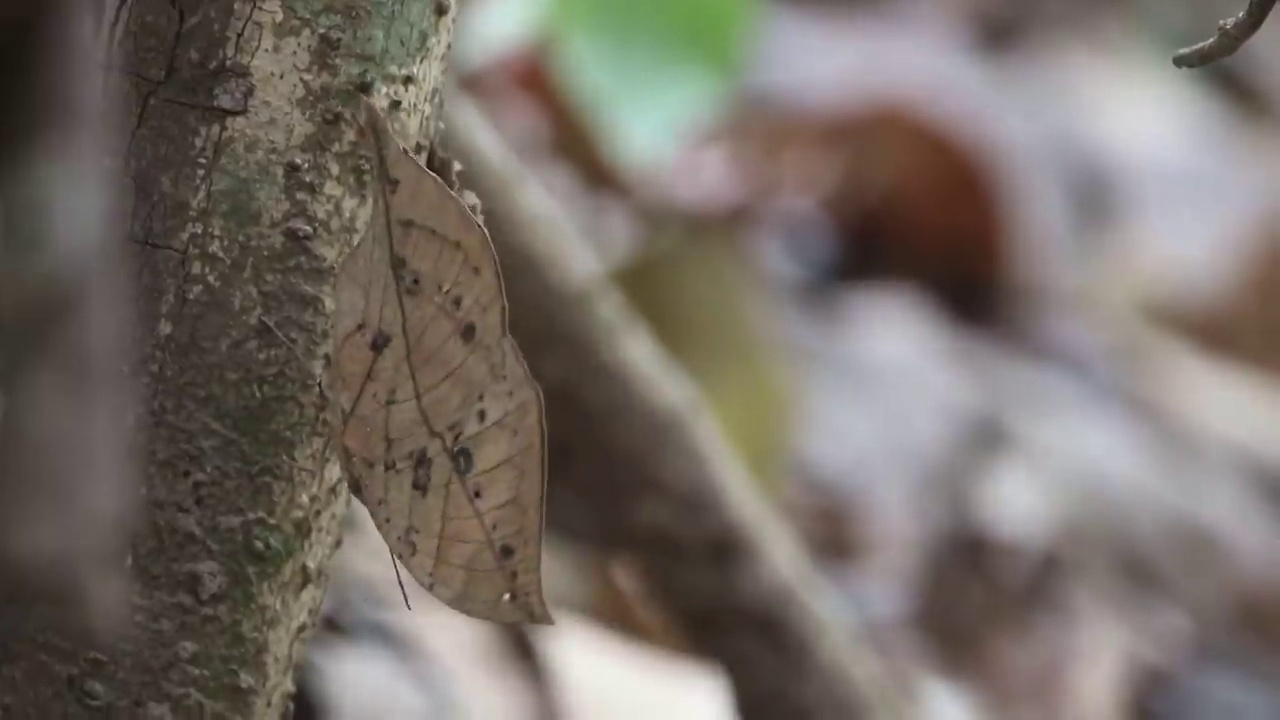}
    \includegraphics[width=0.24\textwidth]{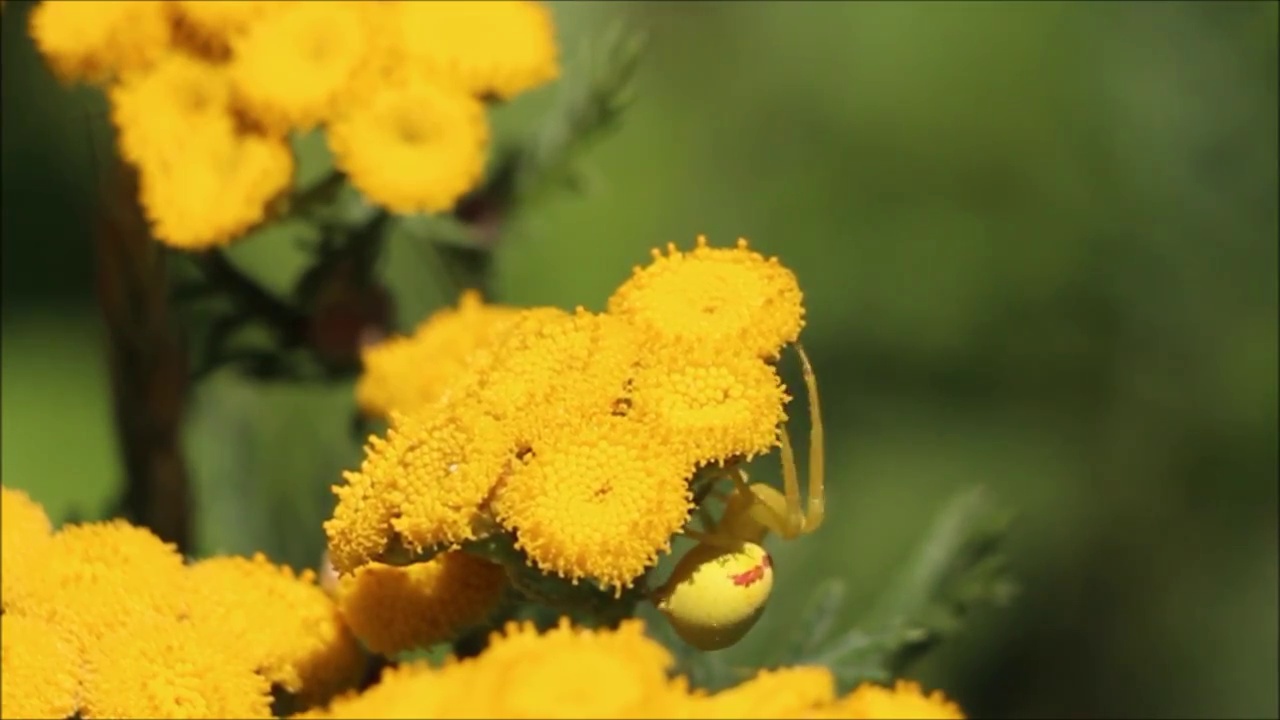}
  \end{subfigure}
  \begin{subfigure}{1\textwidth}
  \centering
    \includegraphics[width=0.24\textwidth]{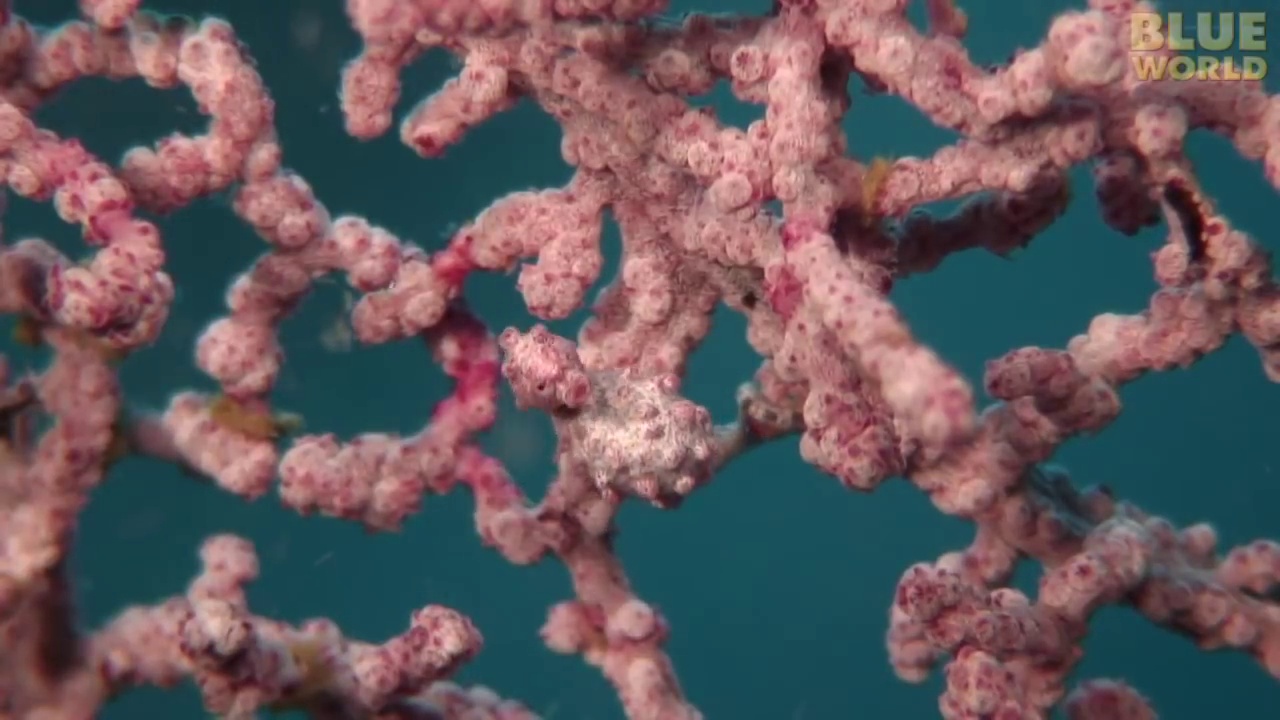}
    \includegraphics[width=0.24\textwidth]{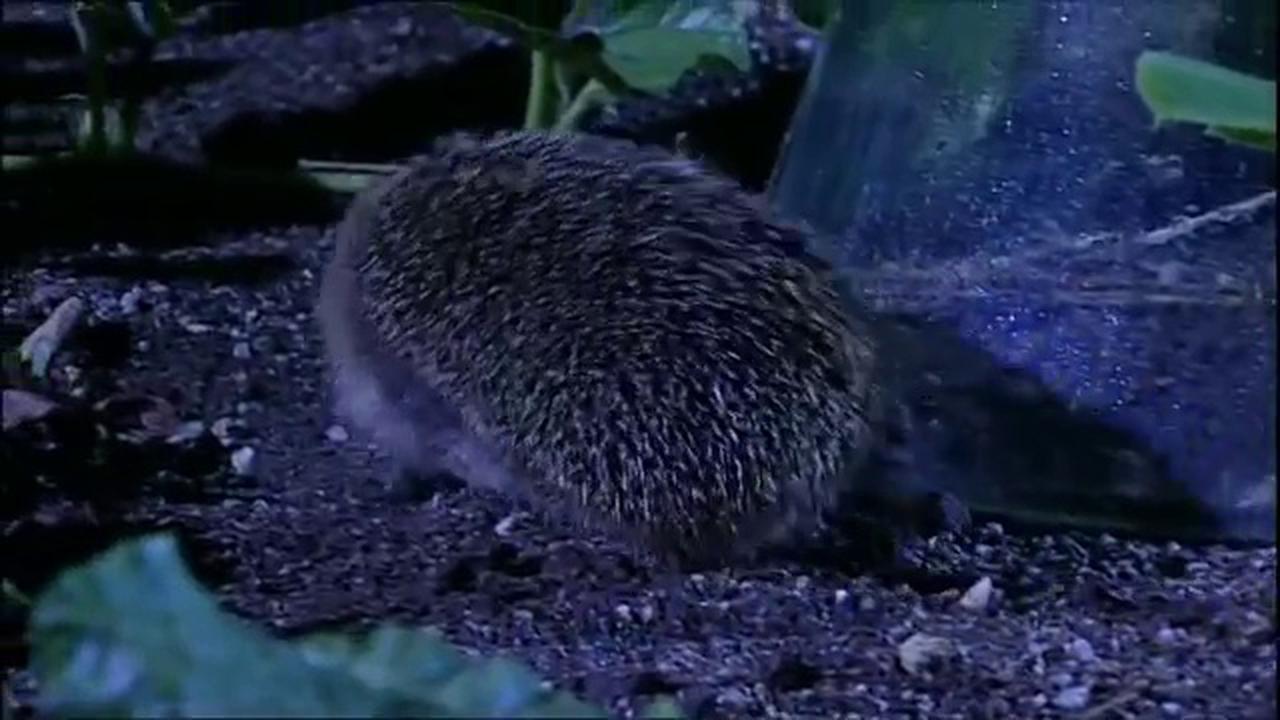}
    \includegraphics[width=0.24\textwidth]{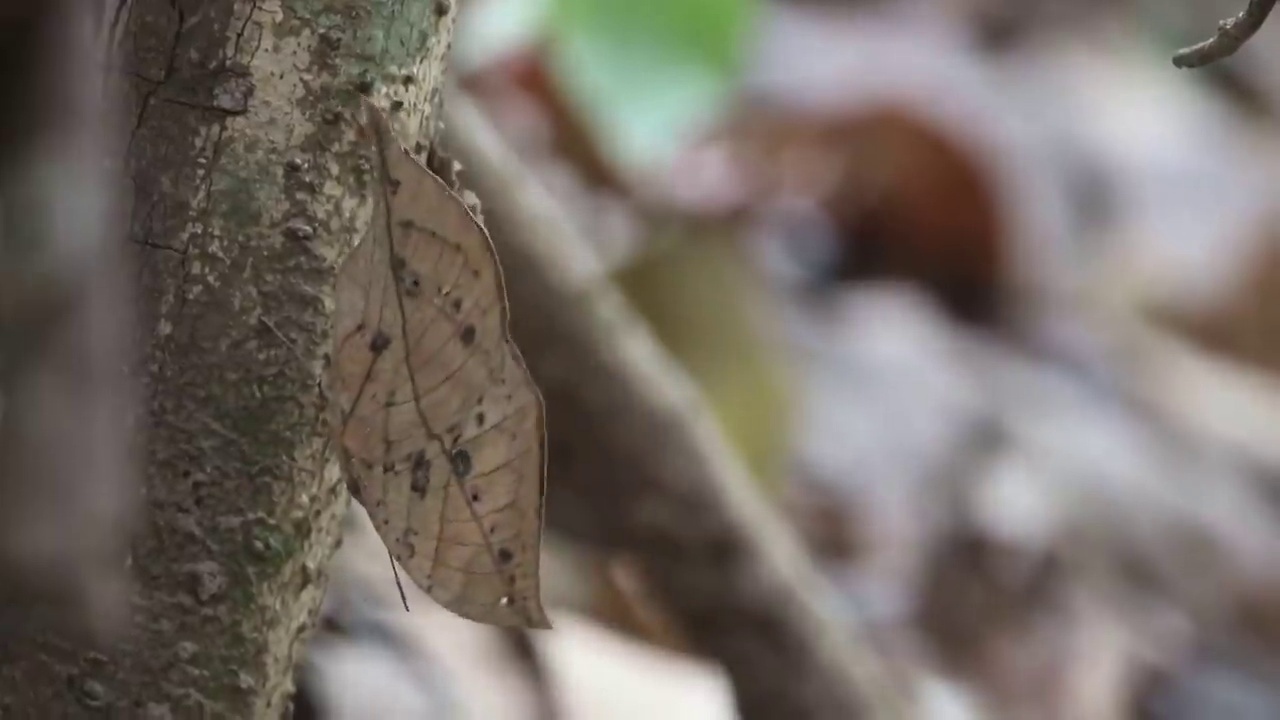}
    \includegraphics[width=0.24\textwidth]{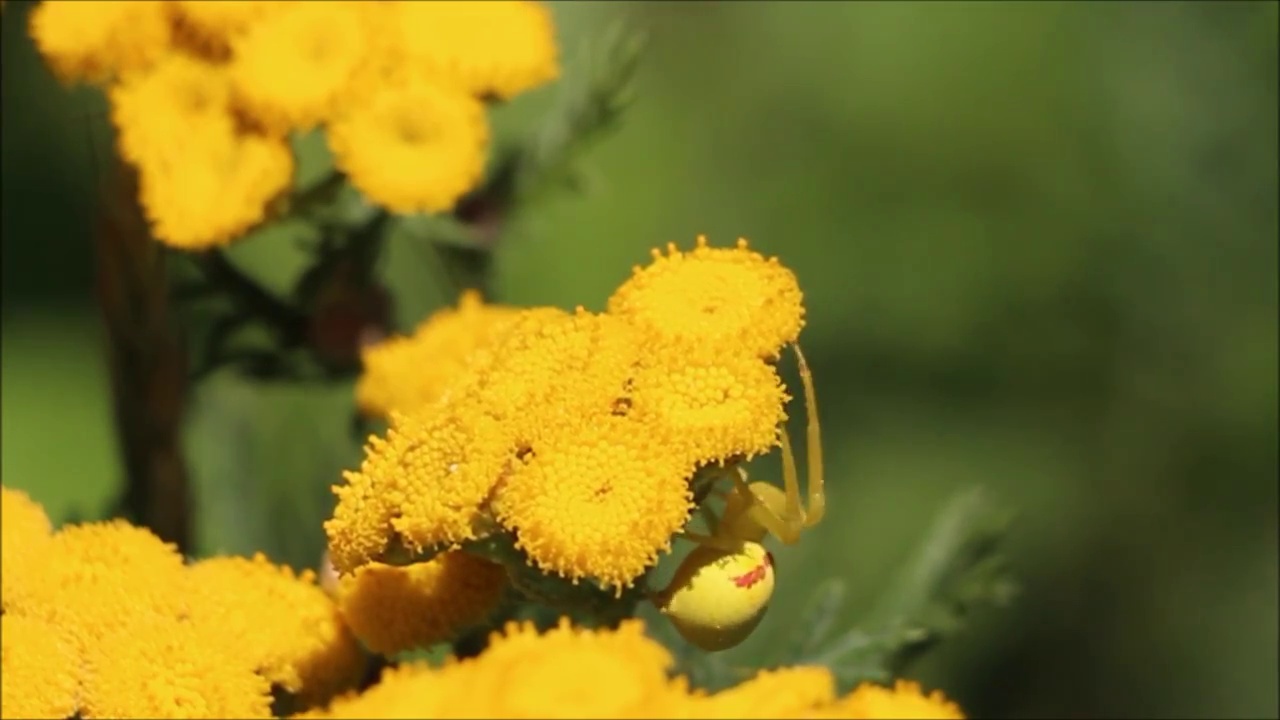}
  \end{subfigure}
   \begin{subfigure}{\textwidth}
    \centering
    \includegraphics[width=0.24\textwidth]{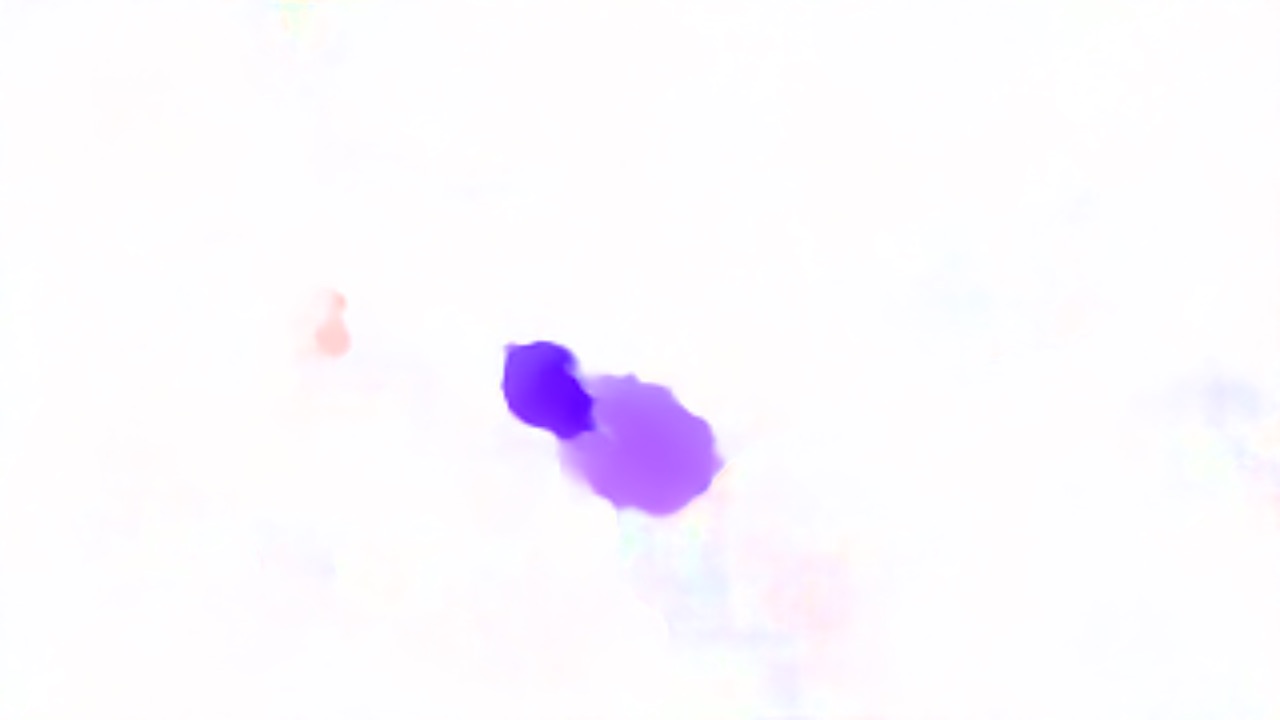}
    \includegraphics[width=0.24\textwidth]{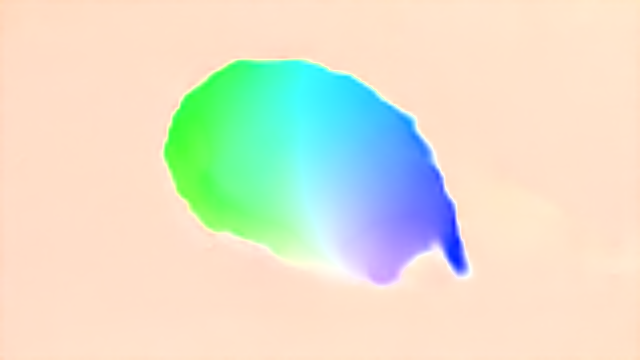}
    \includegraphics[width=0.24\textwidth]{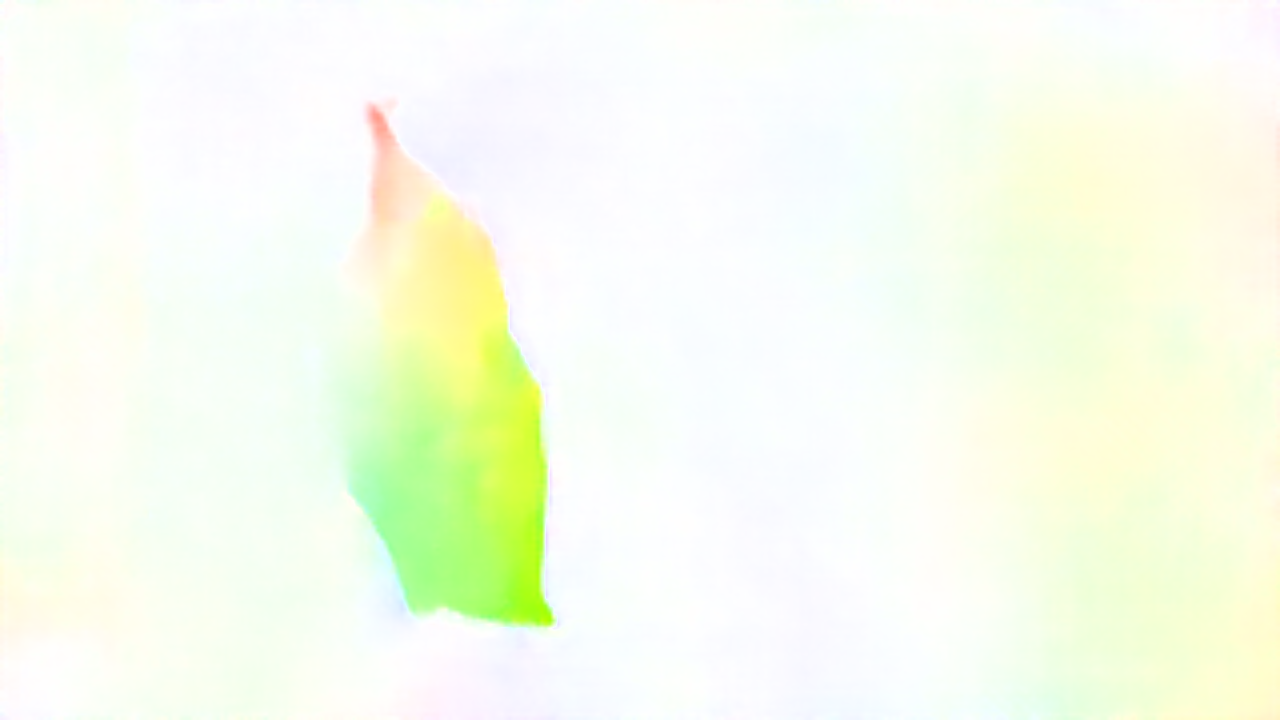}
    \includegraphics[width=0.24\textwidth]{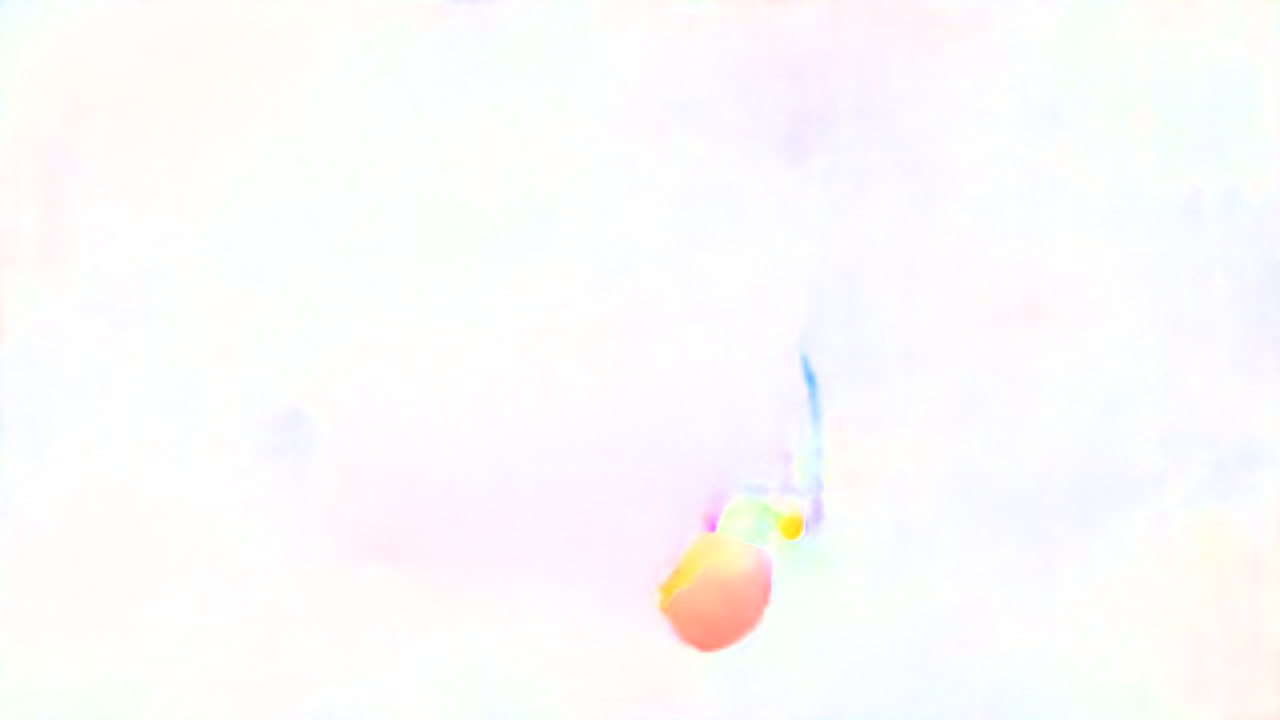}
  \end{subfigure}
     \begin{subfigure}{\textwidth}
    \centering
    \includegraphics[width=0.24\textwidth]{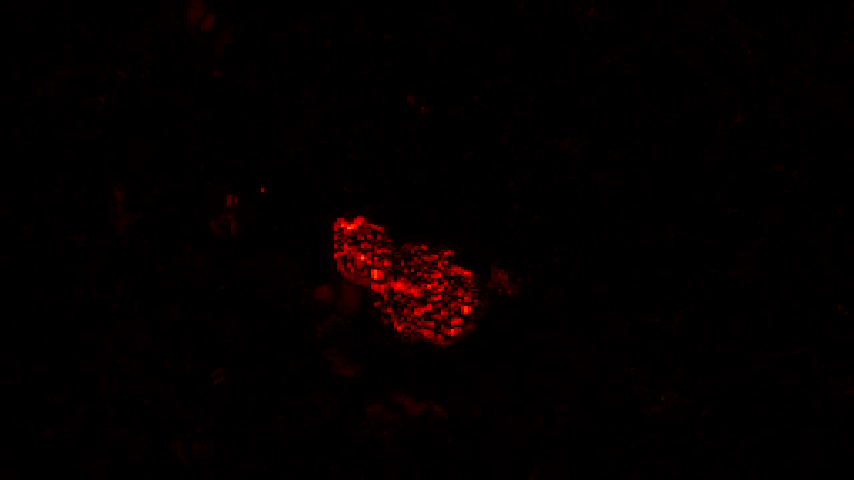}
    \includegraphics[width=0.24\textwidth]{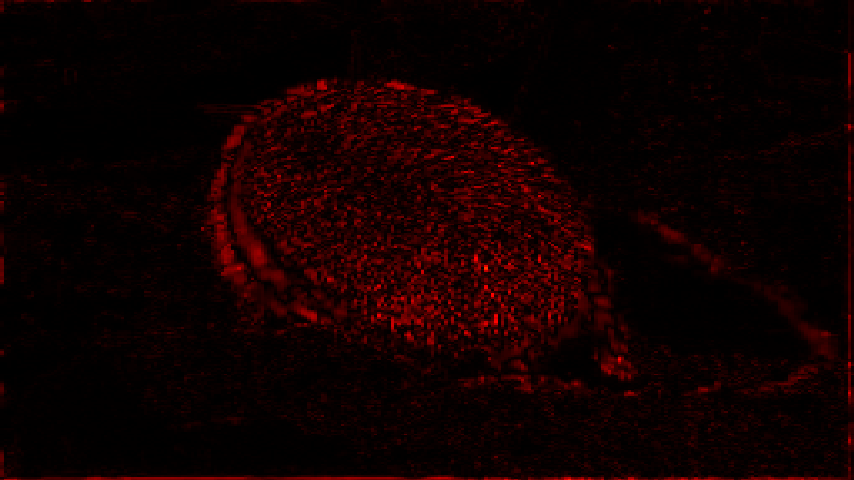}
    \includegraphics[width=0.24\textwidth]{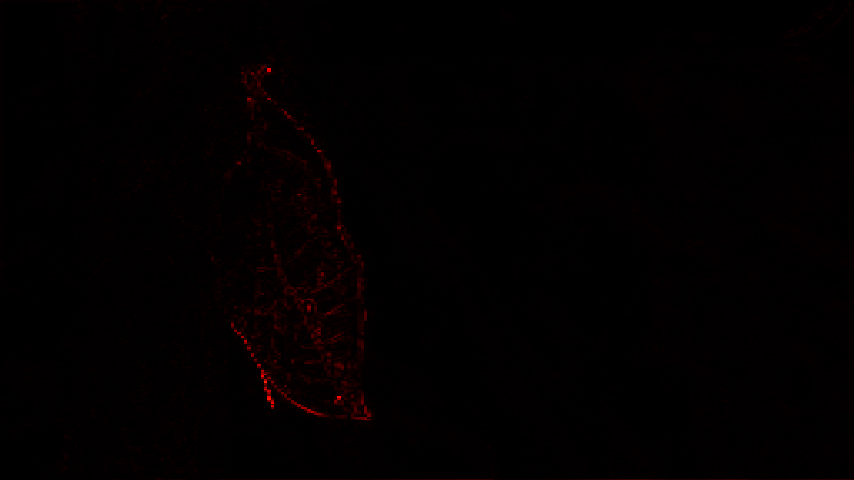}
    \includegraphics[width=0.24\textwidth]{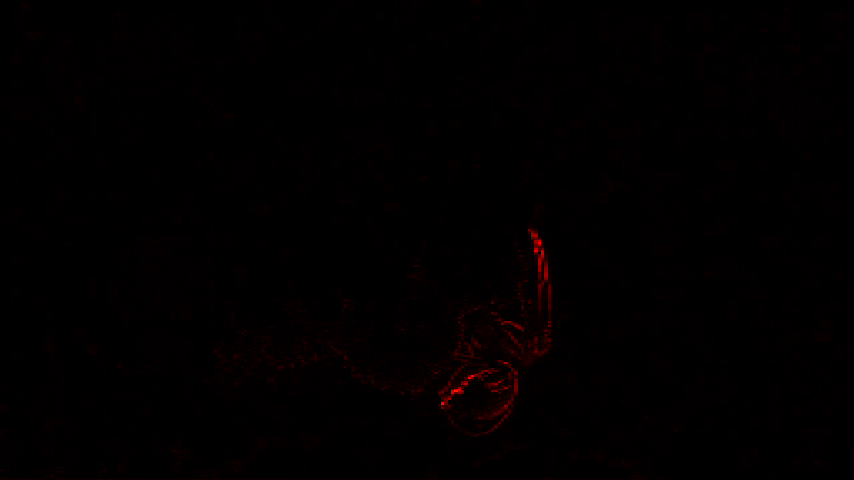}
  \end{subfigure}
   \begin{subfigure}{\textwidth}
   \centering
    \includegraphics[width=0.24\textwidth]{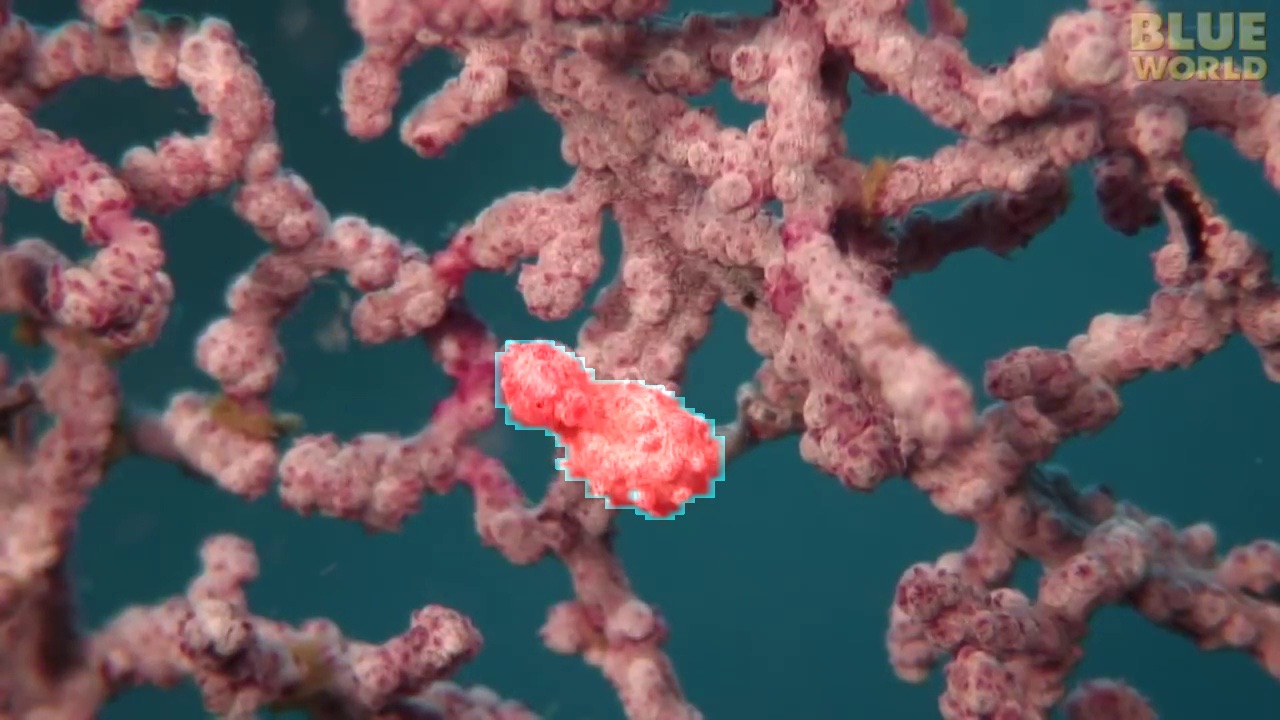}
    \includegraphics[width=0.24\textwidth]{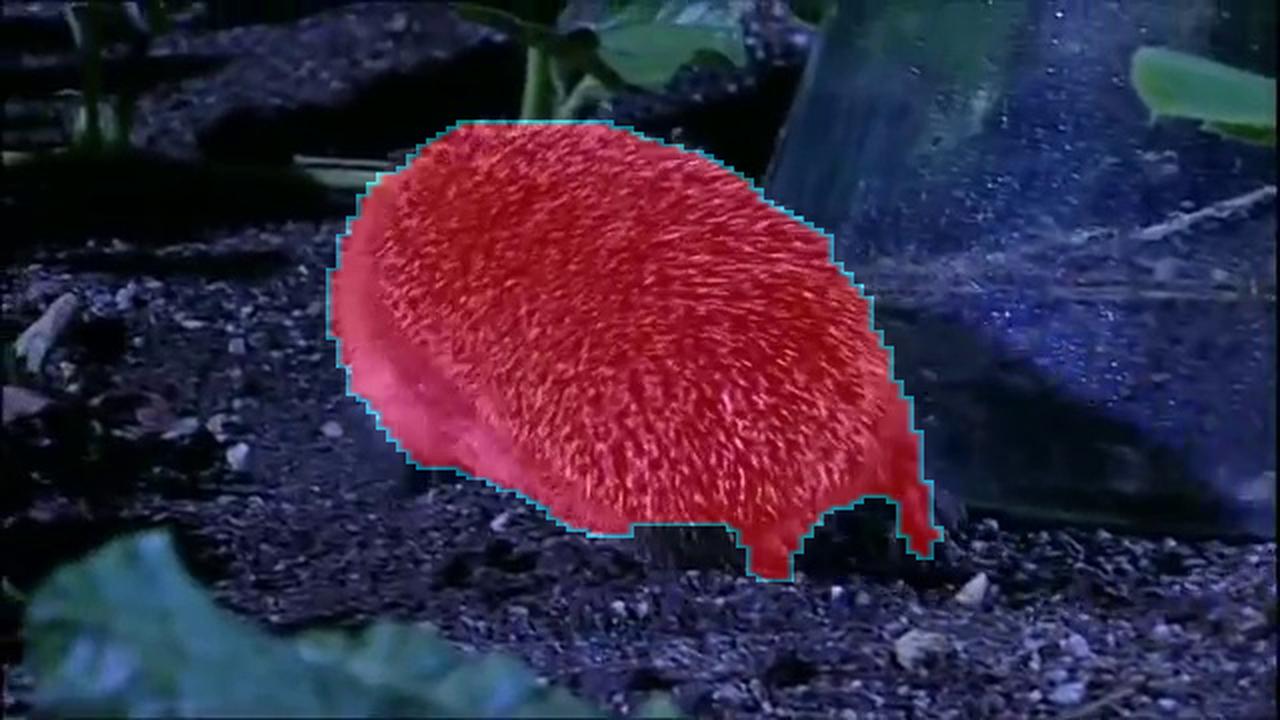}
    \includegraphics[width=0.24\textwidth]{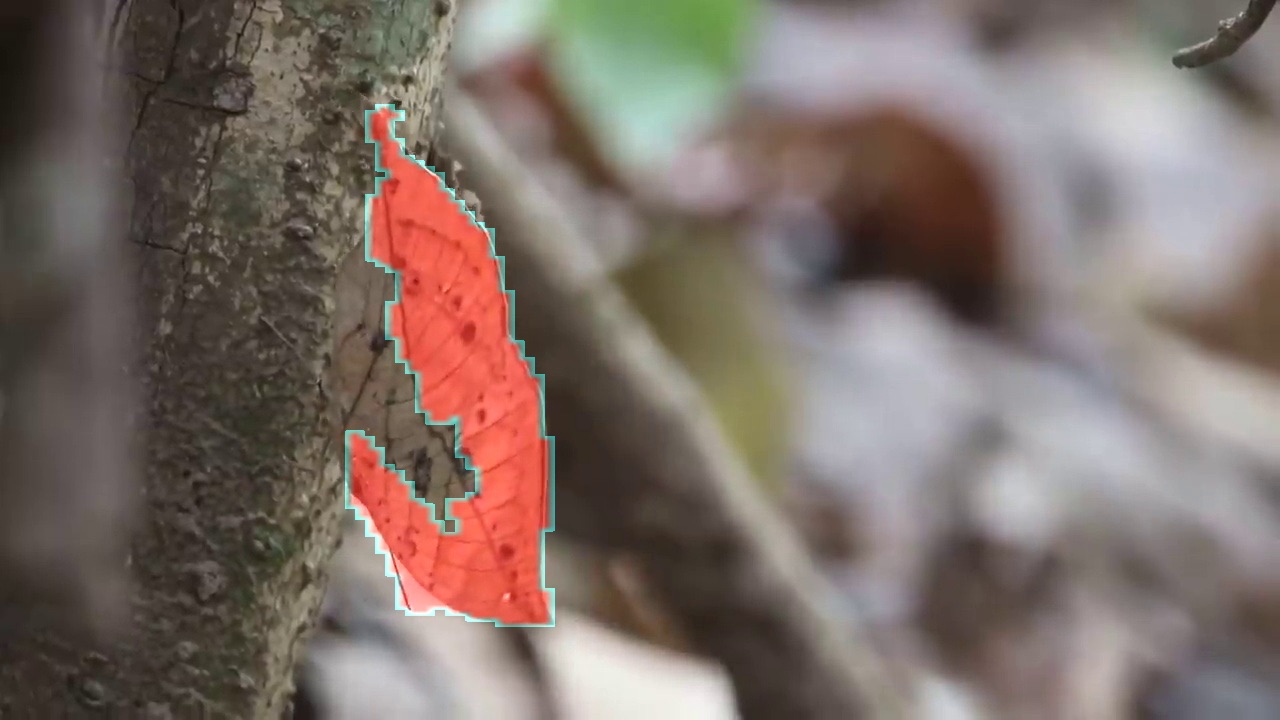}
    \includegraphics[width=0.24\textwidth]{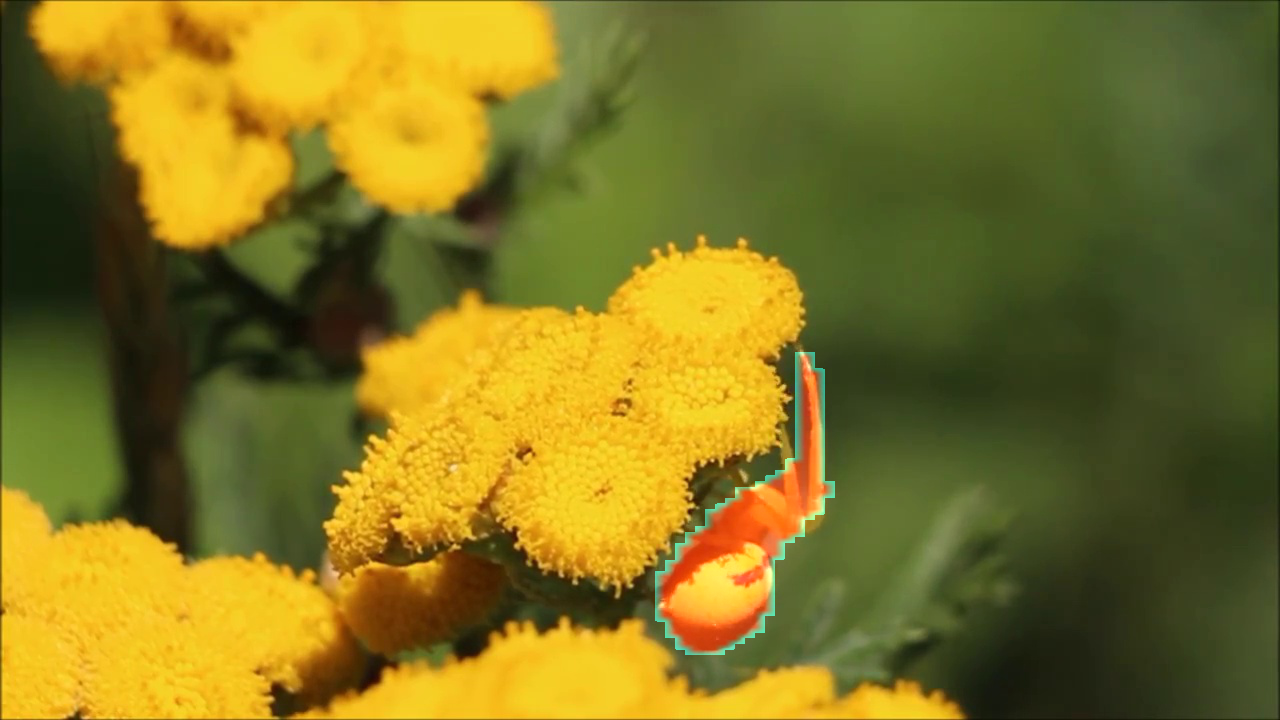}
  \end{subfigure}
\caption{Motion segmentation results on MoCA. 
From top to bottom: frame $t$, frame $t+1$, PWCNet optical flow, aligned image difference, moving object segmentation.}
\label{fig:resCamo}
\end{figure}
\subsection{Results on DAVIS2016 Benchmark}
\label{sec:davis_results}
We compare to previous approaches on
the Unsupervised
Video Object Segmentation protocol.
In all experiments, we {\em do not} use any post-processing {\em e.g.}~CRF. 
Both our re-trained MPNet and LVO model outperform the original reported results in the papers, 
which guarantees a fair comparison with their model. 
For Seg-det, 
replacing the PWC-Flow leads to a small drop~(around $2.3\%$) in the mean $\mathcal{J}$,
but this will not affect our conclusion here. 

As shown in Table~\ref{tab:perfDavis}, 
when compared with MPNet which also take flow-only input, 
our model~(ours-C) outperforms it on all metrics by a large margin.
Note that, the DAVIS benchmark is fundamentally different from MoCA, 
as the approaches only relying on appearance are very effective~\cite{Yang19},
indicating that the objects in the DAVIS sequences can indeed be well-identified by the appearance, 
and motion is not playing the dominant role as it is in MoCA.
This can also be observed from the results for Seg-det, Seg-track, AnchorDiff and COSNet,  
which show significantly stronger performance on DAVIS than on MoCA.
Given this difference, our flow-based architecture still shows very competitive performance.
Moreover, 
when we extend our model with an RGB appearance stream, we do observe a performance boost, 
but since our appearance model has only been finetuned on the DAVIS training set~(30 training sequences),
the two stream~(ours-E) is not comparable with other models trained with more segmentation data.

\begin{table}[!ht]
\centering
\caption{Results on the validation set of DAVIS 2016.}
\setlength{\tabcolsep}{1pt}
\footnotesize
\begin{tabular}{cc|cc|cc|cccc|}
\hline
                                   &         & \multicolumn{2}{c|}{Flow-based}        & \multicolumn{2}{c|}{RGB-based} &        & \multicolumn{2}{c}{Two-stream}             &                                \\ \hline
\multicolumn{1}{c|}{}              & Measure & MPNet\cite{Tokmakov17} & \textbf{ours-C} & AD\cite{Yang19}   & COSNet\cite{Lu_2019_CVPR} &  LVO\cite{Tokmakov19}  & \cite{Dave19}-det         & \cite{Dave19}-track    & \textbf{ours-E} \\ \cline{2-10} 
\multicolumn{1}{c|}{$\mathcal{J}$} & Mean    & 60.3  & \textbf{65.3} &  \bf 80.3        &    77.6    &   69.8 & \bf 76.8 & 75.8      & 69.9                           \\
\multicolumn{1}{c|}{}              & Recall  & 69.9  & \textbf{77.3} &       90.0       &    \bf 91.4    &     83.8 & 84.8                           & 81.8      &\bf 85.3 \\ \hline
\multicolumn{1}{c|}{$\mathcal{F}$} & Mean    & 58.7  & \textbf{65.1} &      \bf 79.3                &   77.5  &   70.1 & \bf 77.8 & 76.1      & 70.3                           \\
\multicolumn{1}{c|}{}              & Recall  & 64.3  & \textbf{74.7} &             84.7     &    \bf 87.4     &   84.3 & \bf 91.3     & 88.7      & 82.9                           \\ \hline
\end{tabular}
\label{tab:perfDavis}
\end{table}
\section{Conclusions}
To summarise, in this paper we consider the problem of breaking animal camouflage in videos.
Specifically,
we propose a novel and effective architecture with two components:  
a differentiable registration module to highlight object boundaries; 
and a motion segmentation module with memory that discovers moving regions.
As future work, we propose to improve the architecture from two aspects:
{\em First},
building more effective memory module for handling longer video sequences, 
for example, a Transformer~\cite{Vaswani17}.
{\em Second}, 
for objects that are only partially moving, 
RGB appearance is required to get the sense of objectness, 
therefore a future direction is to explore RGB inputs, 
for both visual-matching-based registration and appearance features.
\subsubsection*{Acknowledgements}
This research was supported by the UK EPSRC CDT in AIMS, Schlumberger Studentship, and the UK EPSRC Programme Grant Seebibyte EP/M013774/1.
\bibliographystyle{splncs}
\bibliography{bib/shortstrings,bib/vgg_local,bib/vgg_other}

\begin{thebibliography}{10}

\bibitem{Goodale92}
Goodale, M.A., Milner, A.D.:
\newblock Separate visual pathways for perception and action.
\newblock Trends in Neurosciences \textbf{15} (1992)  20--25

\bibitem{Tokmakov19}
Tokmakov, P., Schmid, C., Alahari, K.:
\newblock Learning to segment moving objects.
\newblock IJCV (2019)

\bibitem{Bideau16a}
Bideau, P., Learned-Miller, E.:
\newblock A detailed rubric for motion segmentation.
\newblock arXiv preprint arXiv:1610.10033 (2016)

\bibitem{Ponttuset17}
Pont-Tuset, J., Perazzi, F., Caelles, S., Arbeláez, P., Sorkine-Hornung, A.,
  Gool, L.V.:
\newblock The 2017 davis challenge on video object segmentation.
\newblock arXiv preprint arXiv:1704.00675 (2017)

\bibitem{xu18}
Xu, N., Yang, L., Fan, Y., Yue, D., Liang, Y., Yang, J., Huang, T.:
\newblock Youtube-vos: A large-scale video object segmentation benchmark.
\newblock In: Proc. ECCV. (2018)

\bibitem{Brox10}
Brox, T., Malik, J.:
\newblock Object segmentation by long term analysis of point trajectories.
\newblock In: Proc. ECCV. (2010)

\bibitem{Ochs11}
Ochs, P., Brox, T.:
\newblock Object segmentation in video: a hierarchical variational approach for
  turning point trajectories into dense regions.
\newblock In: Proc. ICCV. (2011)

\bibitem{Papazoglou13}
Papazoglou, A., Ferrari, V.:
\newblock Fast object segmentation in unconstrained video.
\newblock In: Proc. ICCV. (2013)

\bibitem{Jain17}
Jain, S.D., Xiong, B., Grauman, K.:
\newblock Fusionseg: Learning to combine motion and appearance for fully
  automatic segmentation of generic objects in videos.
\newblock In: Proc. CVPR. (2017)

\bibitem{Dave19}
Dave, A., Tokmakov, P., Ramanan, D.:
\newblock Towards segmenting anything that moves.
\newblock In: ICCV Workshop on Holistic Video Understanding. (2019)

\bibitem{iccv19_stm}
Oh, S.W., Lee, J.Y., Xu, N., Kim, S.J.:
\newblock Video object segmentation using space-time memory networks.
\newblock In: Proc. ICCV. (2019)

\bibitem{cvpr19_feelvos}
Voigtlaender, P., Chai, Y., Schroff, F., Adam, H., Leibe, B., Chen, L.C.:
\newblock Feelvos: Fast end-to-end embedding learning for video object
  segmentation.
\newblock In: Proc. CVPR. (2019)

\bibitem{Vondrick18}
Vondrick, C., Shrivastava, A., Fathi, A., Guadarrama, S., Murphy, K.:
\newblock Tracking emerges by colorizing videos.
\newblock In: ECCV. (2018)

\bibitem{Wang19}
Wang, W., Lu, X., Shen, J., Crandall, D.J., Shao, L.:
\newblock Zero-shot video object segmentation via attentive graph neural
  networks.
\newblock In: Proc. ICCV. (2019)

\bibitem{Lai19}
Lai, Z., Xie, W.:
\newblock Self-supervised learning for video correspondence flow.
\newblock In: Proc. BMVC. (2019)

\bibitem{Lai20}
Lai, Z., Lu, E., Xie, W.:
\newblock {MAST}: A memory-augmented self-supervised tracker.
\newblock In: Proc. CVPR. (2020)

\bibitem{tpami18_osvos-s}
Maninis, K.K., Caelles, S., Chen, Y., Pont-Tuset, J., Leal-Taix{\'e}, L.,
  Cremers, D., Van~Gool, L.:
\newblock Video object segmentation without temporal information.
\newblock (2018)

\bibitem{bmvc17_OnAVOS}
Voigtlaender, P., Leibe, B.:
\newblock Online adaptation of convolutional neural networks for video object
  segmentation.
\newblock arXiv (2017)

\bibitem{cvpr17_OSVOS}
Caelles, S., Maninis, K.K., Pont-Tuset, J., Leal-Taix{\'e}, L., Cremers, D.,
  Van~Gool, L.:
\newblock One-shot video object segmentation.
\newblock In: Proc. CVPR. (2017)

\bibitem{Fragkiadaki12}
Fragkiadaki, K., Zhang, G., Shi, J.:
\newblock Video segmentation by tracing discontinuities in a trajectory
  embedding.
\newblock In: Proc. CVPR. (2012)

\bibitem{Keuper15}
Keuper, M., Andres, B., Brox, T.:
\newblock Motion trajectory segmentation via minimum cost multicuts.
\newblock In: Proc. ICCV. (2015)

\bibitem{Yang19}
Yang, Z., Wang, Q., Bertinetto, L., Bai, S., Hu, W., Torr, P.H.:
\newblock Anchor diffusion for unsupervised video object segmentation.
\newblock In: Proc. ICCV. (2019)

\bibitem{Lu_2019_CVPR}
Xiankai, L., Wenguan, W., Chao, M., Jianbing, S., Ling, S., Fatih, P.:
\newblock See more, know more: Unsupervised video object segmentation with
  co-attention siamese networks.
\newblock In: Proc. CVPR. (2019)

\bibitem{Jun17}
Jun~Koh, Y., Kim, C.S.:
\newblock Primary object segmentation in videos based on region augmentation
  and reduction.
\newblock In: Proc. CVPR. (2017)

\bibitem{Fan19}
Fan, D.P., Wang, W., Cheng, M.M., Shen, J.:
\newblock Shifting more attention to video salient object detection.
\newblock In: Proc. CVPR. (2019)

\bibitem{Le19}
Le, T.N., Nguyen, T.V., Nie, Z., Tran, M.T., Sugimoto, A.:
\newblock Anabranch network for camouflaged object segmentation.
\newblock CVIU (2016)

\bibitem{Hartley04c}
Hartley, R.I., Zisserman, A.:
\newblock Multiple View Geometry in Computer Vision. Second edn.
\newblock Cambridge University Press, ISBN: 0521540518 (2004)

\bibitem{szeliski2004image}
Szeliski, R.:
\newblock Image alignment and stitching: A tutorial.
\newblock Technical Report MSR-TR-2004-92 (2004)

\bibitem{Lowe99}
Lowe, D.:
\newblock Object recognition from local scale-invariant features.
\newblock In: Proc. ICCV. (1999)  1150--1157

\bibitem{Fischler81}
Fischler, M.A., Bolles, R.C.:
\newblock Random sample consensus: {A} paradigm for model fitting with
  applications to image analysis and automated cartography.
\newblock Comm. ACM \textbf{24} (1981)  381--395

\bibitem{Brachmann17}
Brachmann, E., Krull, A., Nowozin, S., Shotton, J., Michel, F., Gumhold, S.,
  Rother, C.:
\newblock Dsac-differentiable ransac for camera localization.
\newblock In: Proc. CVPR. (2017)

\bibitem{Brachmann18}
Brachmann, E., Rother, C.:
\newblock Learning less is more-6d camera localization via 3d surface
  regression.
\newblock In: Proc. CVPR. (2018)

\bibitem{Ranftl18}
Ranftl, R., Koltun, V.:
\newblock Deep fundamental matrix estimation.
\newblock In: Proc. ECCV. (2018)

\bibitem{Rocco18}
Rocco, I., Arandjelovic, R., Sivic, J.:
\newblock End-to-end weakly-supervised semantic alignment.
\newblock In: Proc. CVPR. (2018)

\bibitem{Brachmann19}
Brachmann, E., Rother, C.:
\newblock {N}eural- {G}uided {RANSAC}: {L}earning where to sample model
  hypotheses.
\newblock In: Proc. ICCV. (2019)

\bibitem{Sun18}
Sun, D., Yang, X., Liu, M.Y., Kautz, J.:
\newblock {PWC-Net}: {CNNs} for optical flow using pyramid, warping, and cost
  volume.
\newblock In: Proc. CVPR. (2018)

\bibitem{Hartley04a}
Hartley, R.I., Zisserman, A.:
\newblock Multiple View Geometry in Computer Vision. Second edn.
\newblock Cambridge University Press, ISBN: 0521540518 (2004)

\bibitem{Moo18}
Moo~Yi, K., Trulls, E., Ono, Y., Lepetit, V., Salzmann, M., Fua, P.:
\newblock Learning to find good correspondences.
\newblock In: Proc. CVPR. (2018)

\bibitem{Ronneberger15}
Ronneberger, O., Fischer, P., Brox, T.:
\newblock U-net: Convolutional networks for biomedical image segmentation.
\newblock In: Proc. MICCAI. (2015)

\bibitem{Ballas16}
Ballas, N., Yao, L., Pal, C., Courville, A.:
\newblock Delving deeper into convolutional networks for learning video
  representations.
\newblock In: Proc. ICLR. (2016)

\bibitem{Bideau16}
Bideau, P., Learned-Miller, E.:
\newblock It’s moving! a probabilistic model for causal motion segmentation
  in moving camera videos.
\newblock In: Proc. ECCV. (2016)

\bibitem{Perazzi16}
Perazzi, F., Pont-Tuset, J., McWilliams, B., Van~Gool, L., Gross, M.,
  Sorkine-Hornung, A.:
\newblock A benchmark dataset and evaluation methodology for video object
  segmentation.
\newblock In: Proc. CVPR. (2016)

\bibitem{Tokmakov17}
Tokmakov, P., Alahari, K., Schmid, C.:
\newblock Learning motion patterns in videos.
\newblock In: Proc. CVPR. (2017)

\bibitem{Mayer16}
Mayer, N., Ilg, E., Hausser, P., Fischer, P., Cremers, D., Dosovitskiy, A.,
  Brox, T.:
\newblock A large dataset to train convolutional networks for disparity,
  optical flow, and scene flow estimation.
\newblock In: Proc. CVPR. (2016)

\bibitem{flownet17}
Ilg, E., Mayer, N., Saikia, T., Keuper, M., Dosovitskiy, A., Brox, T.:
\newblock Flownet 2.0: Evolution of optical flow estimation with deep network.
\newblock In: Proc. CVPR. (2017)

\bibitem{Vaswani17}
Vaswani, A., Shazeer, N., Parmar, N., Uszkoreit, J., Jones, L., Gomez, A.N.,
  Kaiser, L., Polosukhin, I.:
\newblock Attention is all you need.
\newblock In: NIPS. (2017)

\bibitem{detone2016deep}
DeTone, D., Malisiewicz, T., Rabinovich, A.:
\newblock Deep image homography estimation.
\newblock arXiv preprint arXiv:1606.03798 (2016)

\bibitem{cvpr18_rgmp}
Wug~Oh, S., Lee, J.Y., Sunkavalli, K., Joo~Kim, S.:
\newblock Fast video object segmentation by reference-guided mask propagation.
\newblock In: Proc. CVPR. (2018)

\end{thebibliography}
\appendix
\newpage
\appendix
\section{Appendix}
\subsection{Homography Transformation Estimation}
\label{ap:homography_estim}
This section provides further details on the estimation of the homography transformation described in~\ref{sec:diff_regist}. 
Consider the matrix A referenced in equation~\ref{eqn:homogeneous1} as a standard DLT~\cite{Hartley04a} mapping of a grid of source points to their corresponding target points, 
\textit{i.\@e.\@} $\Omega = \{({x_{i}}^{s},{y_{i}}^{s},{x_{i}}^{t},{y_{i}}^{t}), i \in[\![1,N]\!] \} $. Specifically, A is expressed as the following:
\begin{center} 
$\setlength\arraycolsep{2pt} A = \begin{pmatrix}
{x_{1}}^{s} &{y_{1}}^{s}& 1 & 0& 0&0&-{x_{1}}^{s} {x_{1}}^{t} & -{y_{1}}^{s}{x_{1}}^{t} & -{x_{1}}^{t}\\
0& 0& 0  &{x_{1}}^{s}& {y_{1}}^{s}& 1&- {x_{1}}^{s}{y_{1}}^{t} & -{y_{1}}^{s}{y_{1}}^{t} &- {y_{1}}^{t}\\
& & &  &  & \vdots & & & & \\
{x_{i}}^{s} &{y_{i}}^{s}& 1 & 0& 0&0&-{x_{i}}^{s} {x_{i}}^{t} & -{y_{i}}^{s}{x_{i}}^{t} & -{x_{i}}^{t}\\
0& 0& 0  &{x_{i}}^{s}& {y_{i}}^{s}& 1&- {x_{i}}^{s}{y_{i}}^{t} & -{y_{i}}^{s}{y_{i}}^{t} &- {y_{i}}^{t}\\
  & & &  &  & \vdots & & & &  \\
  {x_{N}}^{s}& {y_{N}}^{s}& 1 & 0& 0& 0&-{x_{N}}^{s}{x_{N}}^{t} &- {y_{N}}^{s}{x_{N}}^{t} & -{x_{N}}^{t}\\
0&0& 0&{x_{N}}^{s}& {y_{N}}^{s}& 1& -{x_{N}}^{s}{y_{N}}^{t} &- {y_{N}}^{s}{y_{N}}^{t} &-{y_{N}}^{t}
\end{pmatrix} $
\end{center} 
Note that each pair of corresponding points constitutes two rows in the matrix A, 
and $H$ has $8$ degrees of freedom. Hence, traditional methods find a minimal set of 4 pairs of linearly independent corresponding points. We use a normalized equidistant grid $\Omega$ of $N=64\times64$ correspondences derived from the forward optical flow mapping: 
$\Omega = \{{(x_{i}}^{s},{y_{i}}^{s}, {x_{i}}^{s}+{F^x}_{s\rightarrow t}({x_{i}}^{s},{y_{i}}^{s}),{y_{i}}^{s}+{F^y}_{s\rightarrow t}({x_{i}}^{s},{y_{i}}^{s})), i \in[\![1,N]\!] \}$.

Finally, as in~\cite{Ranftl18}, the corresponding homography matrix is
estimated via a sigular value decomposition:
 $U \Sigma V^T = A^T diag(w) A$, 
with $h = \frac{v_0}{\left\lVert v_0\right\rVert}$ where $v_0$ is the right singular vector corresponding to the smallest eigen value.

\subsection{Synthetic Moving Chairs}
\label{ap:synthetic_chairs}
\begin{figure}[!htb]
\centering
\begin{subfigure}[hb]{0.2\textwidth}
    \makebox[0pt][r]{\makebox[25pt]{\raisebox{15pt}{\rotatebox[origin=c]{90}{ $Q_{t}$}}}}%
    \includegraphics[width=\textwidth]
    {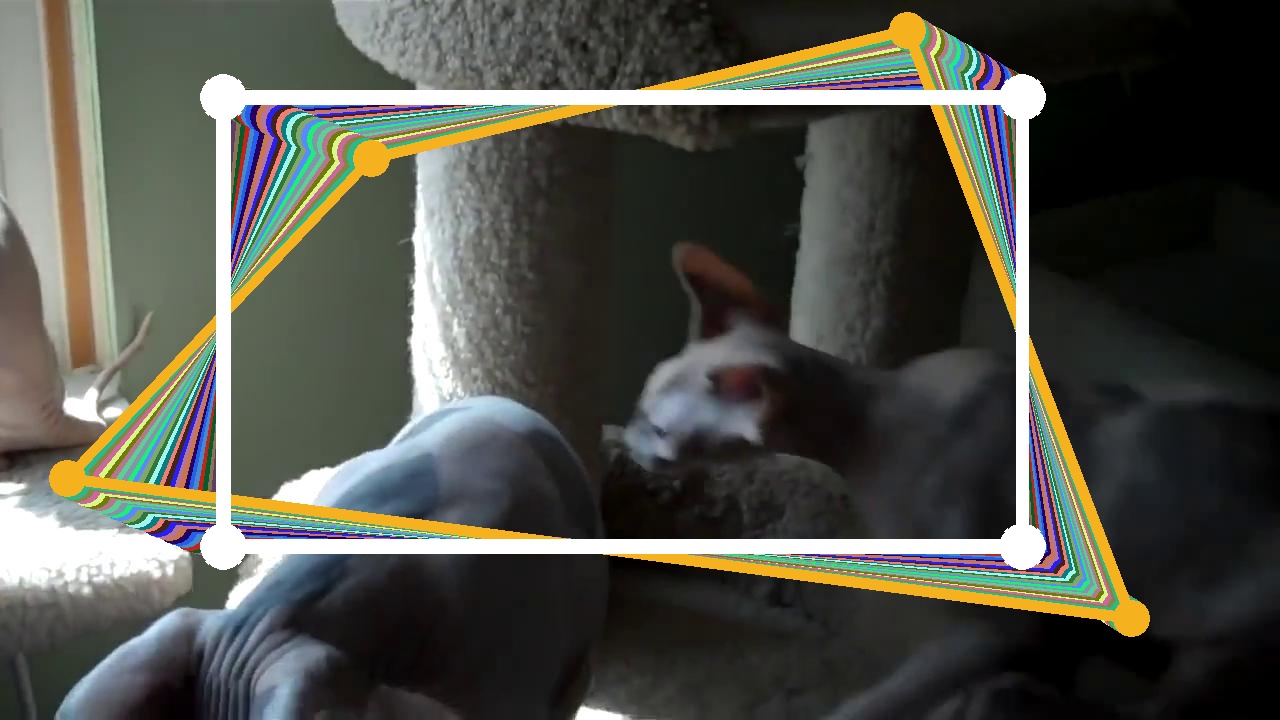} \par\bigskip
    \makebox[0pt][r]{\makebox[25pt]{\raisebox{15pt}{\rotatebox[origin=c]{90}{$Frame_t$}}}}%
    \includegraphics[width=\textwidth]
    {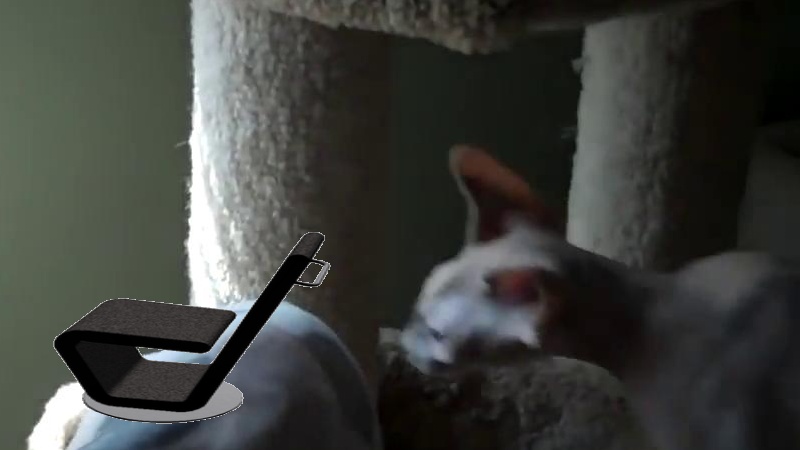} \par\bigskip
    \makebox[0pt][r]{\makebox[25pt]{\raisebox{15pt}{\rotatebox[origin=c]{90}{${F}_{t\rightarrow t+1}$}}}}%
    \includegraphics[width=\textwidth]
    {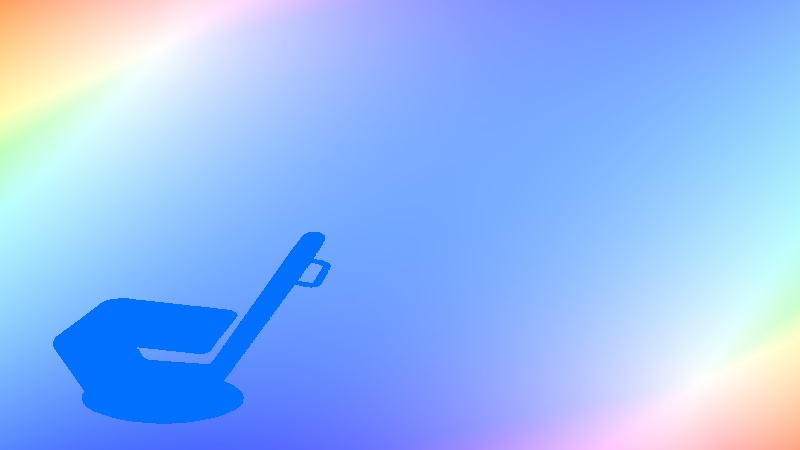}
    \caption{$t=0$}
\end{subfigure}
\begin{subfigure}[hb]{0.2\textwidth}
    \includegraphics[width=\textwidth]
    {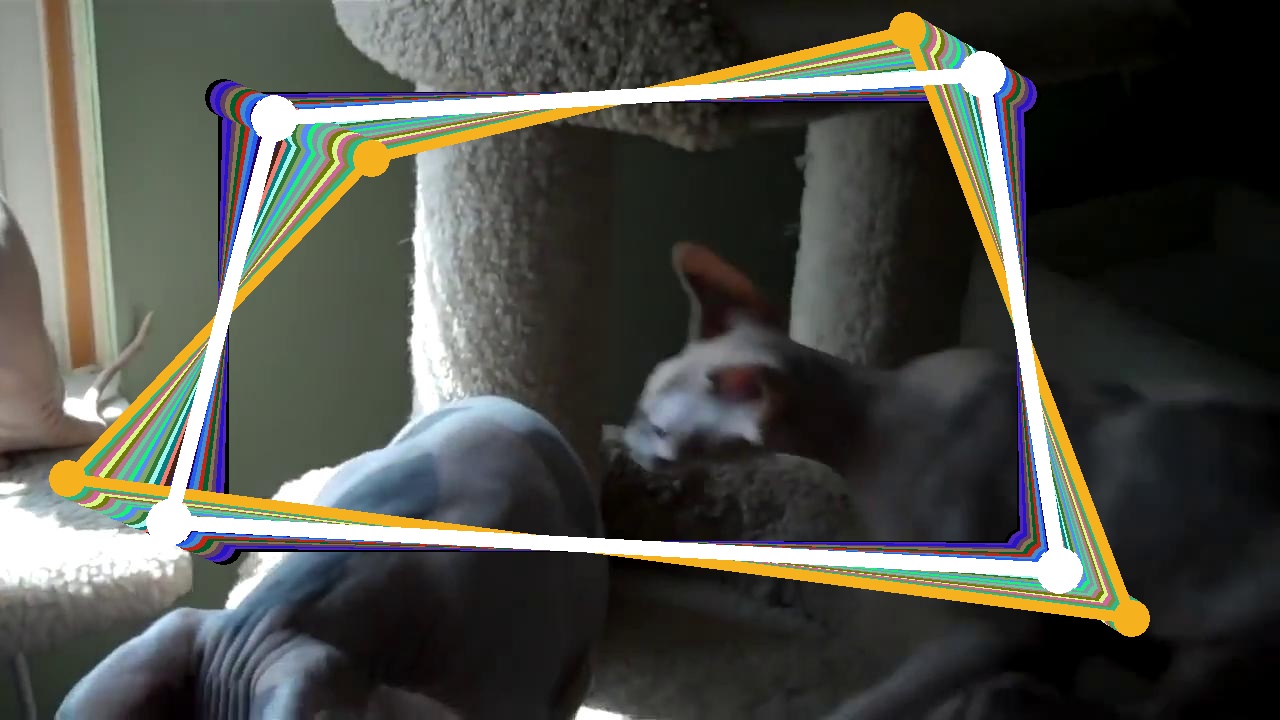}
    \par\bigskip
    \includegraphics[width=\textwidth] 
    {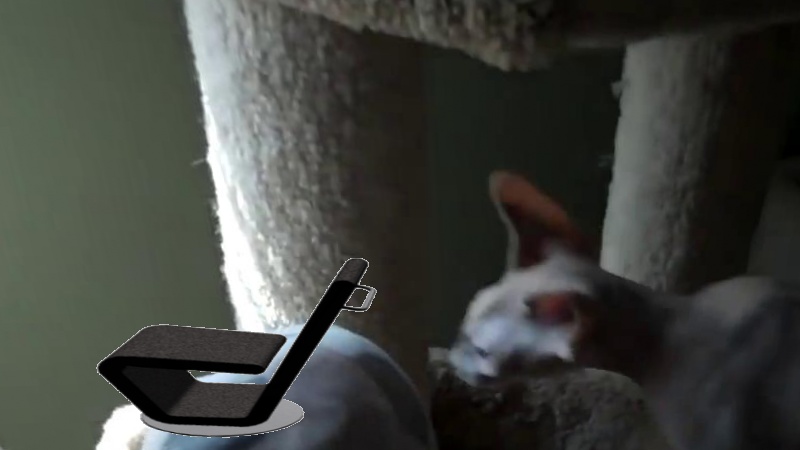}\par\bigskip
    \includegraphics[width=\textwidth]
    {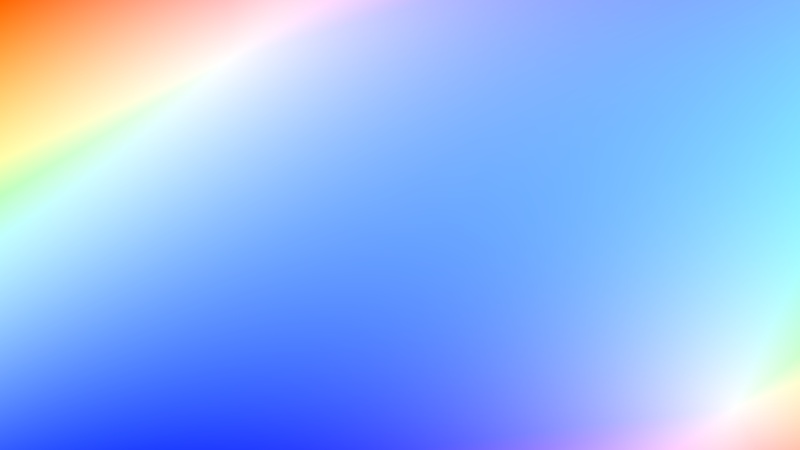}
    \caption{$t=10$}
\end{subfigure}
\begin{subfigure}[hb]{0.2\textwidth}
    \includegraphics[width=\textwidth]
    {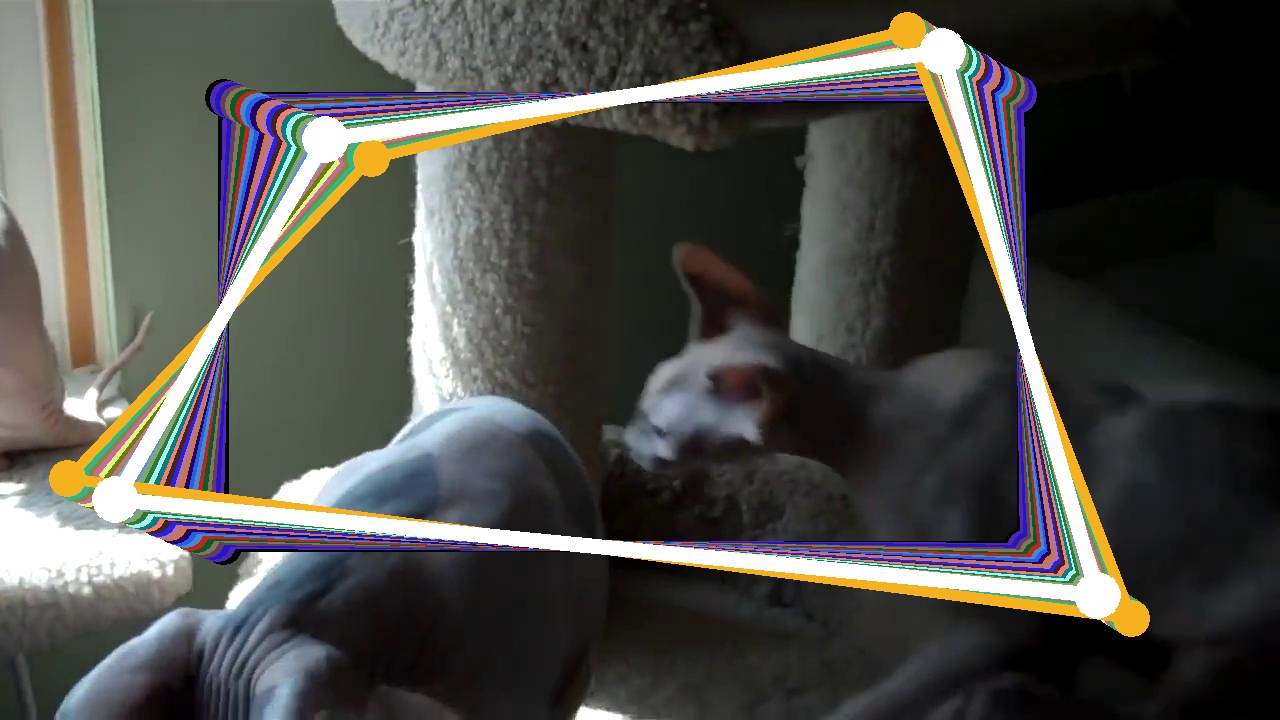}\par\bigskip
    \includegraphics[width=\textwidth]  
    {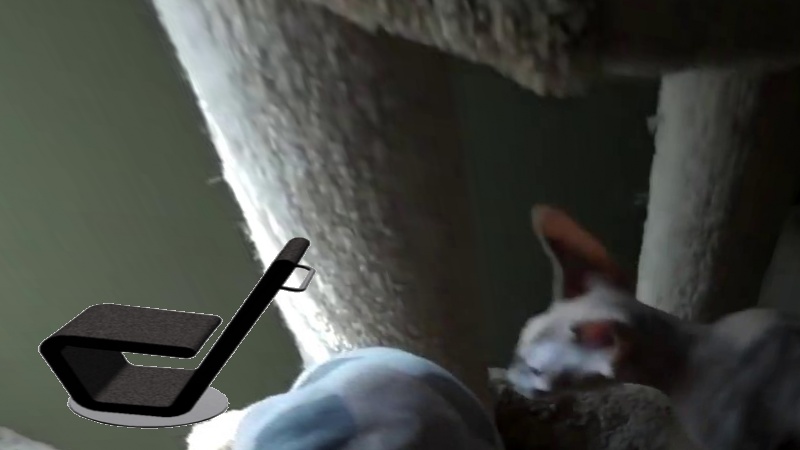}\par\bigskip
    \includegraphics[width=\textwidth]
    {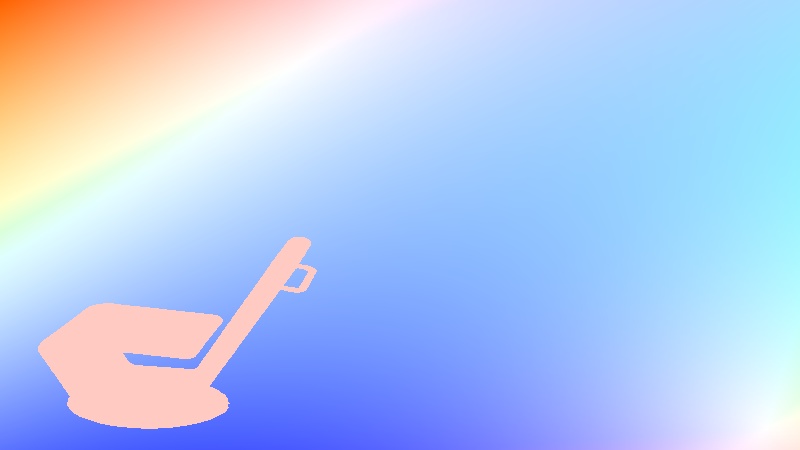}
    \caption{$t=20$}
\end{subfigure}
\begin{subfigure}[hb]{0.2\textwidth}
    \includegraphics[width=\textwidth]
    {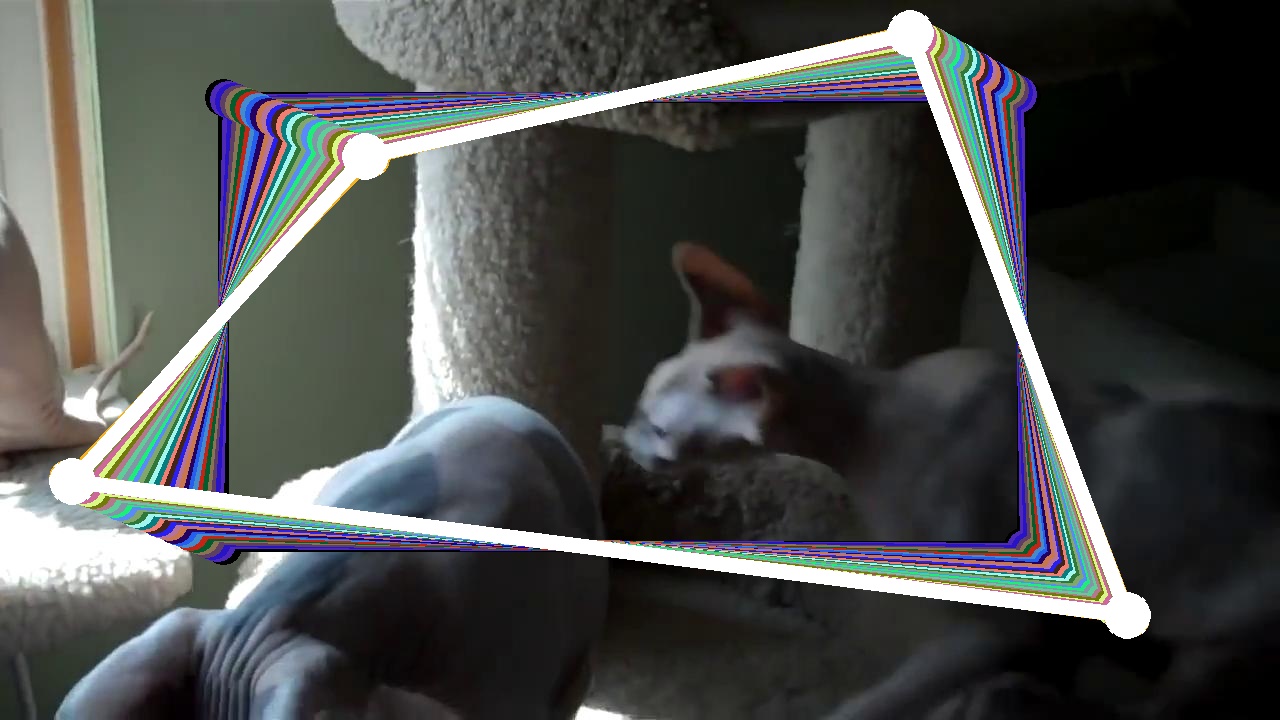}
    \par\bigskip
    \includegraphics[width=\textwidth]  
    {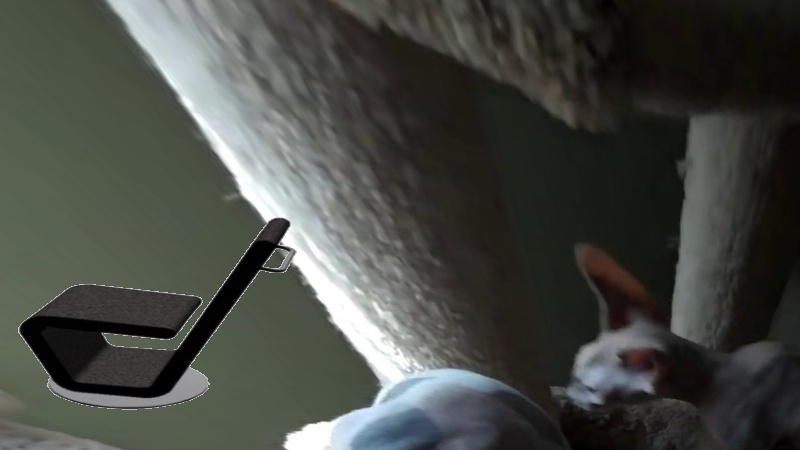}
    \par\bigskip
    \includegraphics[width=\textwidth]
    {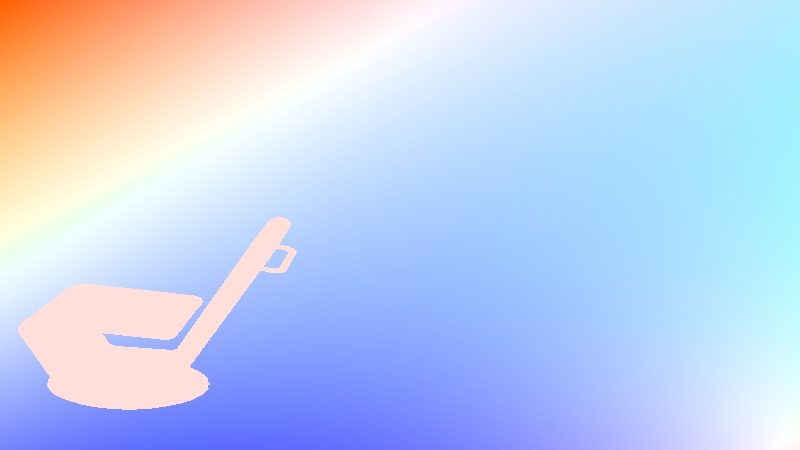}
    \caption{$t=28$}
\end{subfigure}
\caption{Examples from a continuous sequence with a momentarily static object towards $t=10$. The corresponding quadrilateral is represented in white}
\label{fig:continuous}
\end{figure}

\begin{figure}[!htb]
\centering
\begin{subfigure}[hb]{0.2\textwidth}
    \makebox[0pt][r]{\makebox[25pt]{\raisebox{15pt}{\rotatebox[origin=c]{90}{ $Q_{t}$}}}}%
    \includegraphics[width=\textwidth]
    {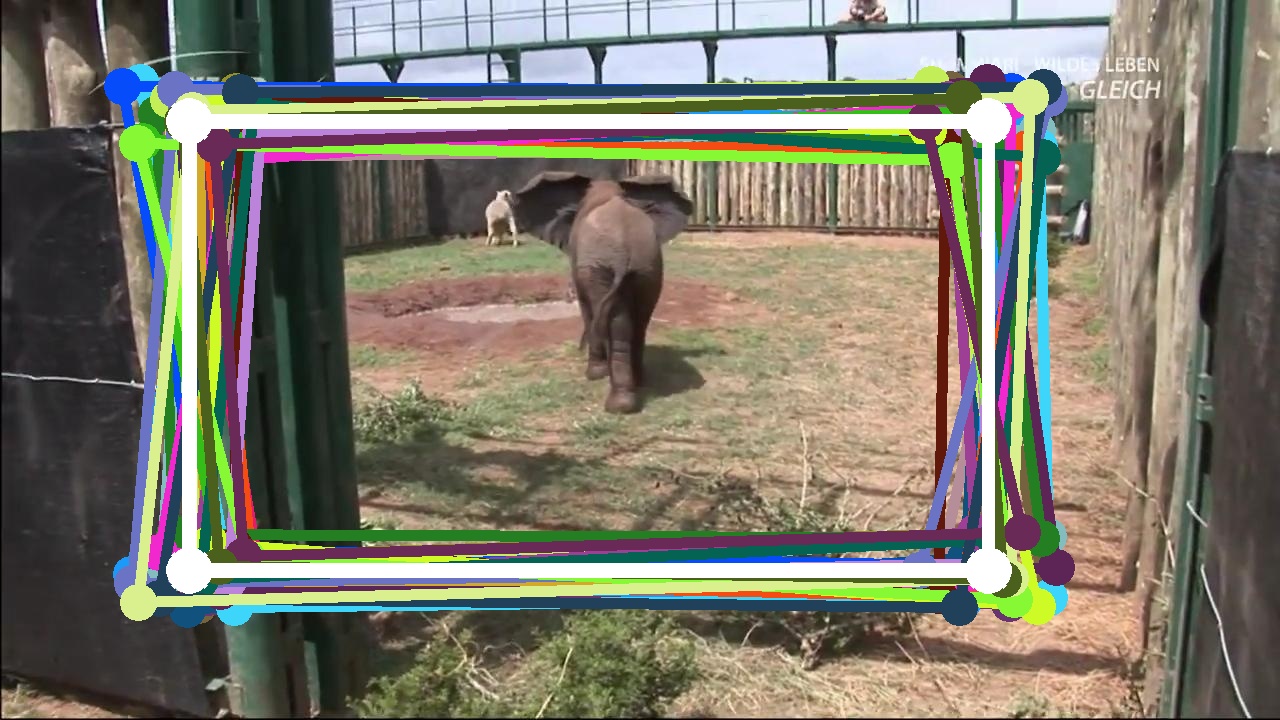} \par\bigskip
    \makebox[0pt][r]{\makebox[25pt]{\raisebox{15pt}{\rotatebox[origin=c]{90}{$Frame_t$}}}}%
    \includegraphics[width=\textwidth]
    {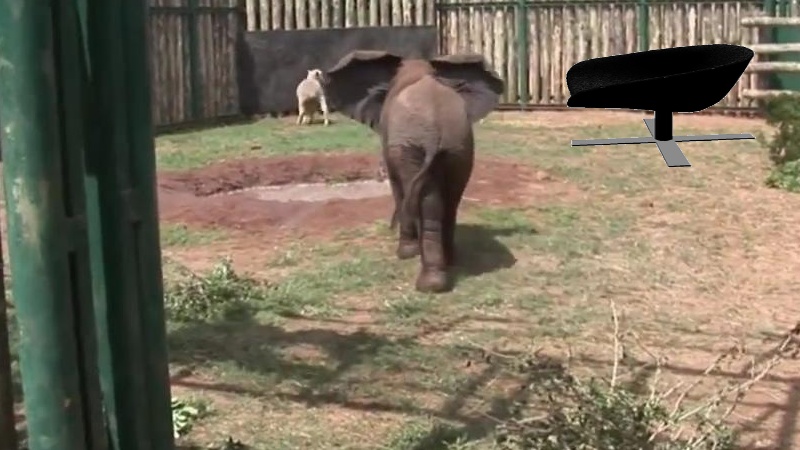} \par\bigskip
    \makebox[0pt][r]{\makebox[25pt]{\raisebox{15pt}{\rotatebox[origin=c]{90}{${F}_{t\rightarrow t+1}$}}}}%
    \includegraphics[width=\textwidth]
    {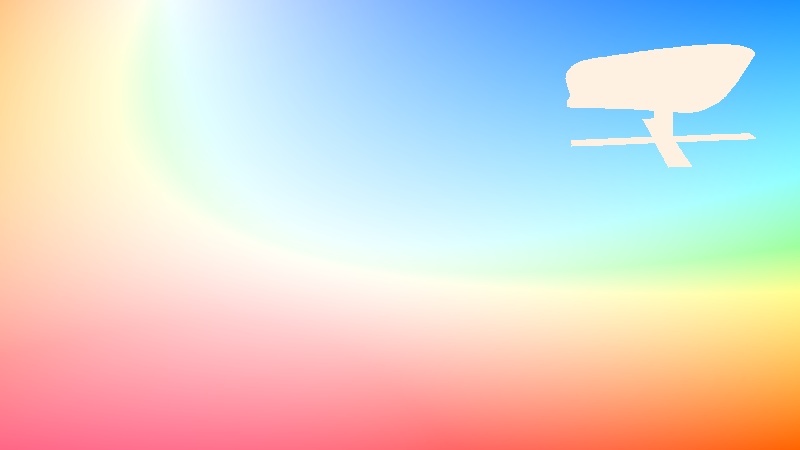}
    \caption{$t=0$}
\end{subfigure}
\begin{subfigure}[htb]{0.2\textwidth}
    \includegraphics[width=\textwidth]
    {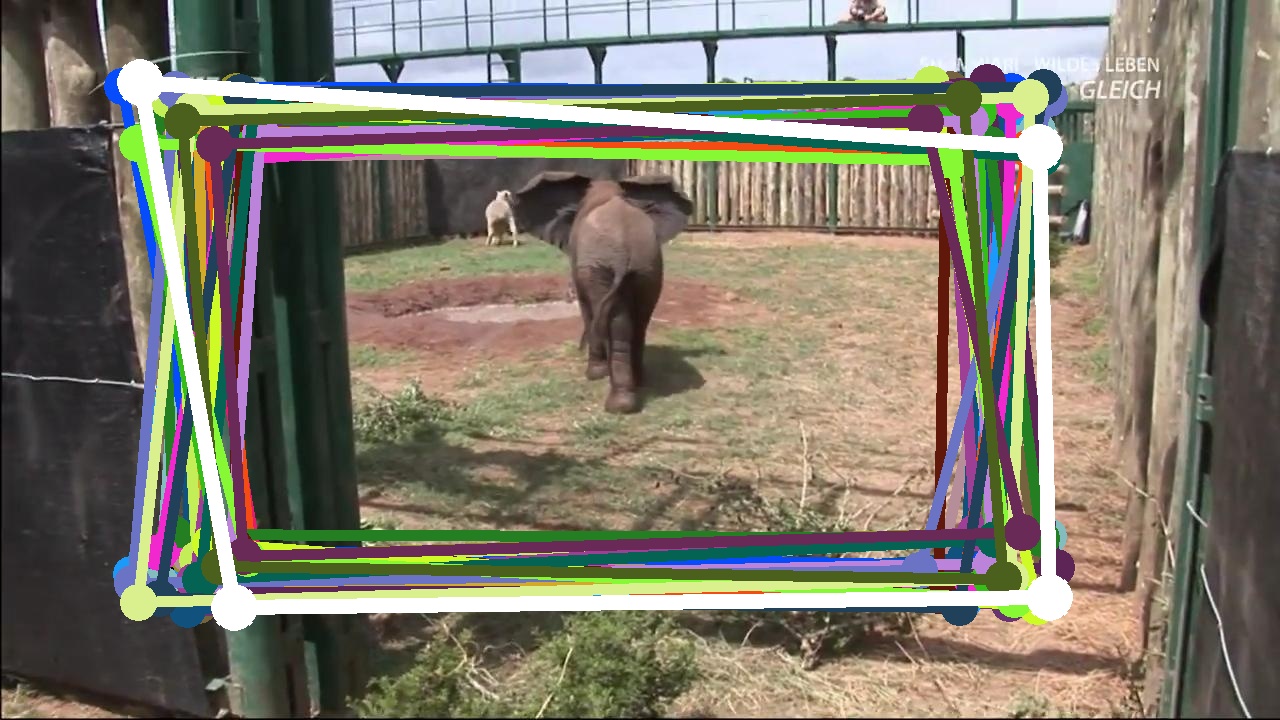}
    \par\bigskip
    \includegraphics[width=\textwidth] 
    {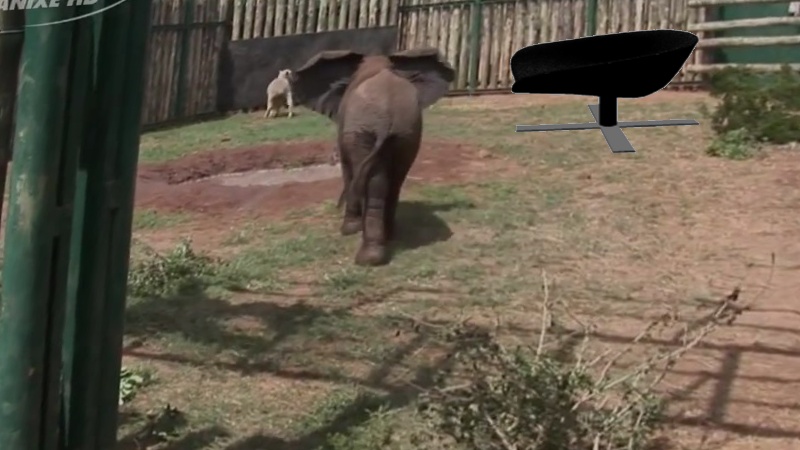}\par\bigskip
    \includegraphics[width=\textwidth]
    {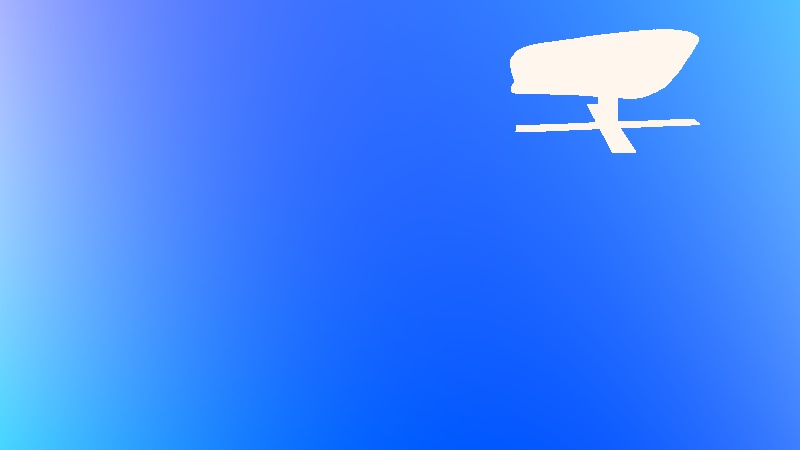}
    \caption{$t=10$}
\end{subfigure}
\begin{subfigure}[htb]{0.2\textwidth}
    \includegraphics[width=\textwidth]
    {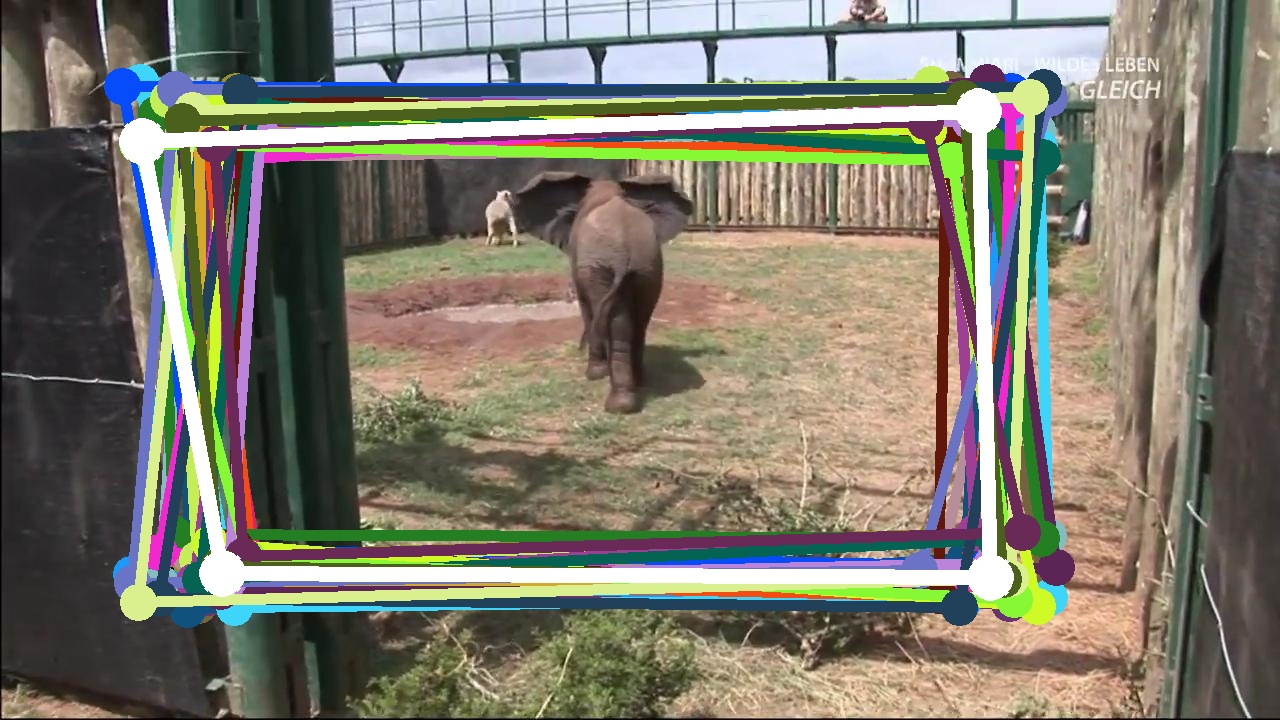}\par\bigskip
    \includegraphics[width=\textwidth]  
    {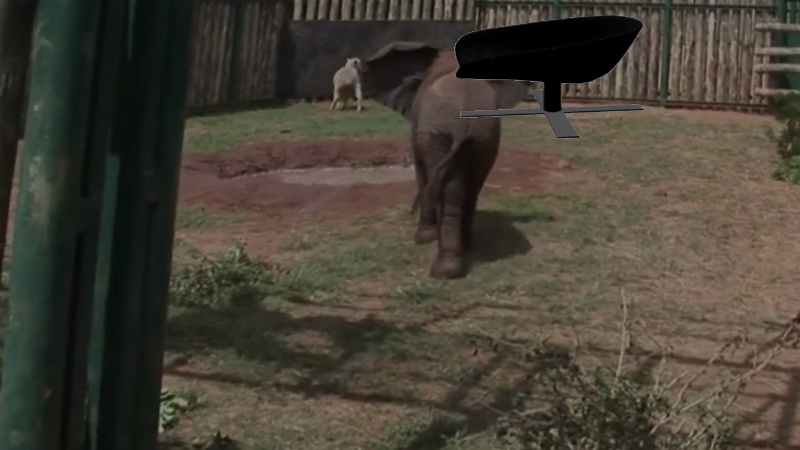}\par\bigskip
    \includegraphics[width=\textwidth]
    {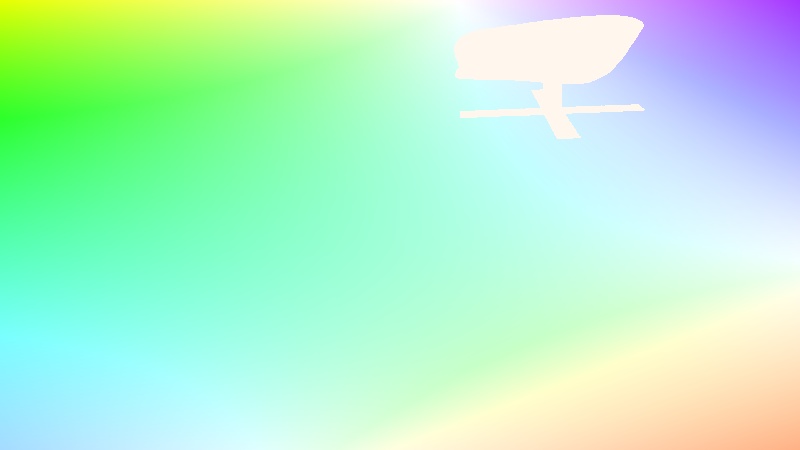}
    \caption{$t=20$}
\end{subfigure}
\begin{subfigure}[htb]{0.2\textwidth}
    \includegraphics[width=\textwidth]
    {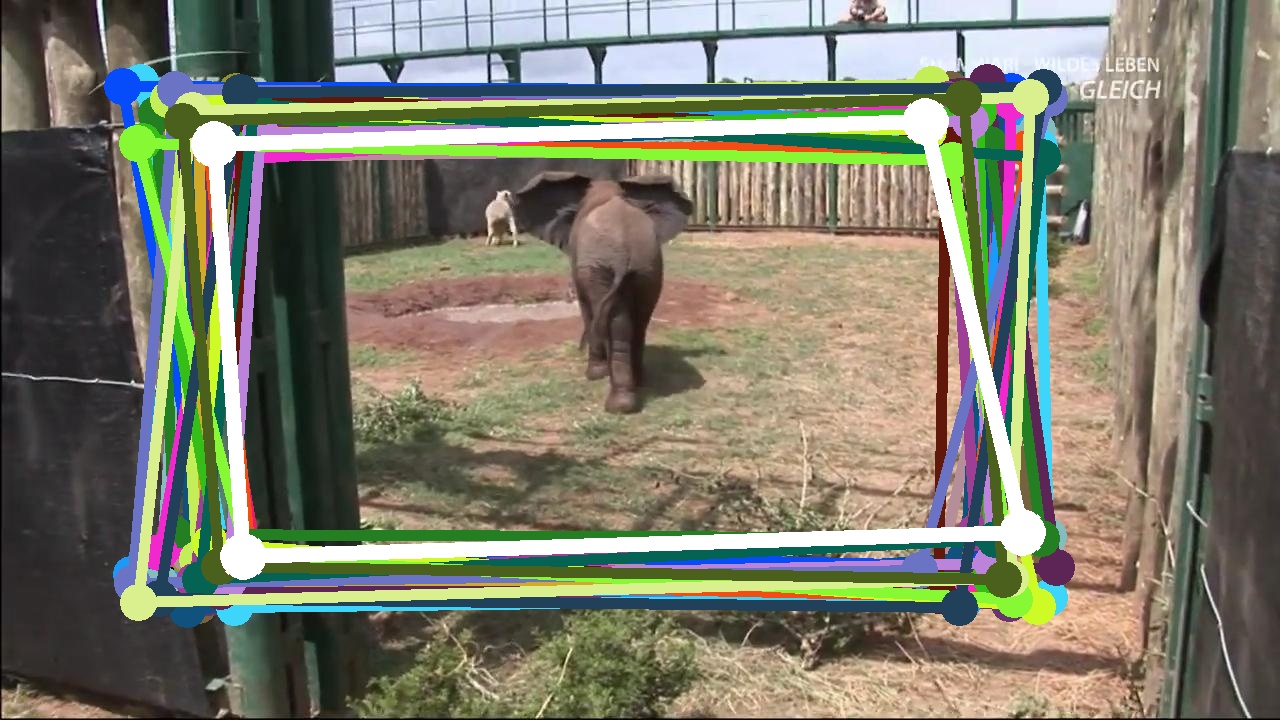}
    \par\bigskip
    \includegraphics[width=\textwidth]  
    {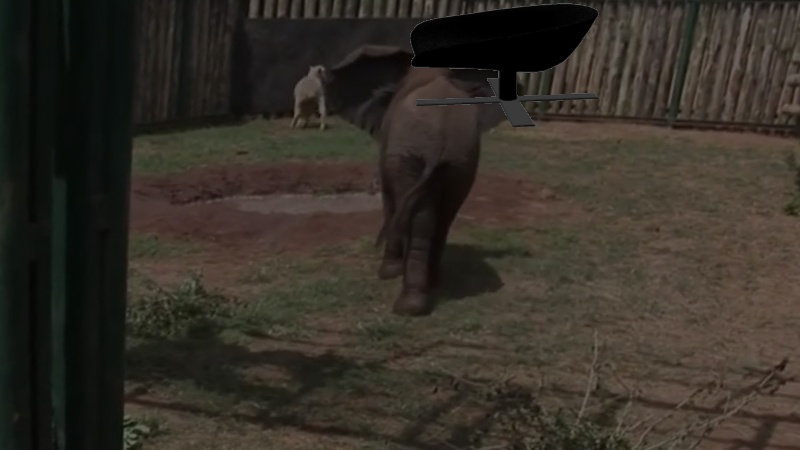}
    \par\bigskip
    \includegraphics[width=\textwidth]
    {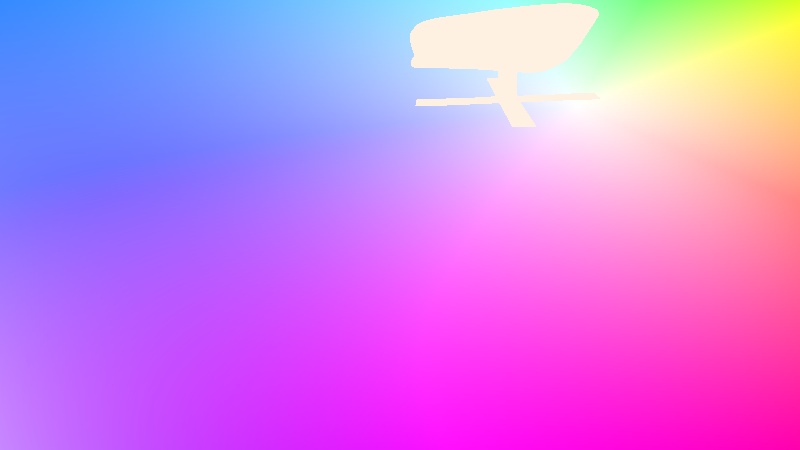}
    \caption{$t=28$}
\end{subfigure}
\caption{Examples from a random sequence with brightness change. The corresponding quadrilateral is represented in white}
\label{fig:random}
\end{figure}

Our generation protocol is closely inspired by the method described in~\cite{detone2016deep}, 
which we extend from a pair of images to a sequence. 
Namely, we generate a sequence from a single background image by applying a sequence of homographies.  We first select the initial $F_{t=0}$ frame by cropping a random rectangle of which we jitter the vertices in order to generate the quadrilaterals defining the sequence images. 
For each quadrilateral $Q_{t=t_{i}} \neq Q_{t=0}$ we compute the underlying homography, \textit{i.\@e.\@} mapping $Q_{t=t_{i}}$ to $Q_{t=t_{0}}$, 
and apply it to the image $F_{t=0}$ to obtain $F_{t=t_{i}}$. 
We consider two types of sequences: 
continuous sequences~(Figure~\ref{fig:continuous}), 
computed via linear interpolation, to mimic a continuous camera motion; 
and random sequences~(Figure~\ref{fig:random}), 
to simulate a shaking camera scenario. 
We further incorporate momentarily static objects and brightness change to imitate real cases.

\vspace{-5pt}

\clearpage
\subsection{End-to-end Motion segmentation pipeline}
\label{ap:detailed_architecture}
In this section, we provide a detailed description of the main modules of our end-to-end motion segmentation pipeline~(Figure~\ref{fig:architecture}).

\subsubsection{Differentiable registration.}
We adopt the model introduced in~\cite{Moo18} totalling $12$ residual blocks where each block is composed of two linear layers, 
separated by a global context normalization (GCN), batch normalization (BN1d) and ReLU~(Table~\ref{tab:mlp}).

\begin{table}[hb]
\centering
\begin{tabular}{c|c|c}
\hline
Architecture& \begin{tabular}[c]{@{}l@{}}Input $\in \mathcal{R}^{N\times4\times64\times64}$\\ \end{tabular}                                                                                                                                & Output                  \\ \hline
\multicolumn{1}{l|}{}       & Reshape   & $N\times4\times4096$ \\ \cline{2-3} 
\multicolumn{1}{l|}{}       & Conv1d, $(1, 4, 128)$    & $N\times128\times4096$ \\ \cline{2-3} 
\multicolumn{1}{l|}{Registration} &
\begin{tabular}[c]{@{}l@{}}Residual block:\\$\begin{bmatrix} \text{Conv1d}, (1, 128, 128)\\\text{GCN}\\\text{BN1d}+\text{ReLU}\\ \text{Conv1d}, (1, 128, 128)\\ \text{GCN}\\ \text{BN1d}+\text{ReLU}\end{bmatrix} \times 12$\\\\\end{tabular} & $N\times128\times4096$ \\ \cline{2-3} 
\multicolumn{1}{l|}{}                 & \begin{tabular}[c]{@{}l@{}} \\$\begin{bmatrix}\text{Conv1d}, (1, 128, 1)\\ \text{Sigmoid} \end{bmatrix}$\\\\\end{tabular}                                                                                   & $N\times1\times4096$   \\ \hline
\end{tabular}
\vspace{4pt}
\caption{Architecture of the differentiable registartion module.}
\label{tab:mlp}
\end{table}

\noindent
Note that the GCN, as defined in~\cite{Moo18}, is a completely non-parametric instance normalization-like layer where each perceptron is normalized across the correspondences of each pair of images separately:
$GCN({o^l}_i) = \frac{{o^l}_i-\overline{{o^l}_i}}{std({o^l}_i)}$
where ${o^l}_i \in \mathcal{R}^d$ denotes the output of the perceptron $l$ for the correspondence $i$ and $\overline{{o^l}_i}$ and $std({o^l}_i)$ respectively the mean and standard deviation of the distribution of the output for all correspondences ${\{{o^l}_i\}}_{i \in [\![1,N]\!]}$.

\subsubsection{Motion Segmentation Encoder.} 
Here, we adopt a variant of ResNet18 architecture described in Table~\ref{tab:encoder}.

\subsubsection{Memory Module. }
We adopt a similar architecture to~\cite{Tokmakov19}. 
For a sequence with 11 frames, 
a convGRU cell of $hidden\_size = 64$ and $kernel\_size=7$ is applied, forward and backward, to the $N\times 11\times 256 \times 64 \times 64$ output of the encoder, where the second dimension refers to the sequence size. This results in $feat\_fwd$ and $feat\_bwd$ both of dimension $(N\times 11)\times 64 \times 64 \times 64 $. These features are further concatenated along the channel dimension and fed to a conv2d of $kernel\_size=3, padding=1$ to produce a bidirectional memory feature of size:  $(N\times 11)\times 64 \times 64 \times 64$.

\subsubsection{Motion Segmentation Decoder. }
The decoder is composed of a residual block and the refinement block introduced in~\cite{cvpr18_rgmp}, taking the output of memory module and the output of Conv\_block\_1 from the encoder, via skip connections, and upscaling by a factor of $2$ resulting in the final motion prediction of size $(N\times11)\times 1 \times 128 \times 128$.

\begin{table}[htb]
\begin{tabular}{c|c|c|}
\hline
         & \begin{tabular}[c]{@{}l@{}}Input $\in \mathcal{R}^{N\times256\times256\times3}$\\ \end{tabular}           & Outputs                                       \\ \hline
\multicolumn{1}{l|}{}        & \multicolumn{1}{l|}{\begin{tabular}[c]{@{}l@{}}Conv\_block\_1:\\ $\begin{bmatrix} \text{Conv2d}, 7\times7\, 3, 64, \text{stride}=2, \text{pad}=3 \\\text{BN2d}+\text{ReLU}\\ \end{bmatrix}$\end{tabular}}                                                                                           & \multicolumn{1}{l|}{$N\times64\times 128 \times 128 $} \\ \cline{2-3} 
\multicolumn{1}{l|}{} & \multicolumn{1}{l|}{\begin{tabular}[c]{@{}l@{}}Residual block\_1:
\\ $\begin{bmatrix}\\ \text{Conv2d}, 3\times3, 64, 64, \text{stride}=2, \text{pad}=1 \\\text{BN2d}+\text{ReLU}\\\text{Conv2d}, 3\times3, 64, 64, \text{stride}=1, \text{pad}=1 \\\text{BN2d}+\text{ReLU}\\\text{Conv2d}, 1\times1, 64, 64, \text{stride}=2, \text{pad}=0 \\\text{BN2d}+\text{ReLU}\\\\ \end{bmatrix} \times 2$\end{tabular}} & \multicolumn{1}{l|}{$N\times128\times64\times64$} \\ \cline{2-3} 

\multicolumn{1}{l|}{Encoder}   & \multicolumn{1}{l|}{\begin{tabular}[c]{@{}l@{}}Residual block\_2:\\ $\begin{bmatrix}\\ \text{Conv2d}, 3\times3, 64, 128, \text{stride}=3, \text{pad}=1 \\\text{BN2d}+\text{ReLU}\\\text{Conv2d}, 3\times3, 64, 128, \text{stride}=1, \text{pad}=1 \\\text{BN2d}+\text{ReLU}\\\text{Conv2d}, 1\times1, 64, 128, \text{stride}=3, \text{pad}=0 \\\text{BN2d}+\text{ReLU}\\\\ \end{bmatrix} \times 2$\end{tabular}} & $N\times128\times64\times64$                        \\ \cline{2-3} 

\multicolumn{1}{l|}{}   & \multicolumn{1}{l|}{\begin{tabular}[c]{@{}l@{}}Residual block\_3:\\ $\begin{bmatrix}\\ \text{Conv2d}, 3\times3, 128, 256, \text{stride}=2, \text{pad}=1 \\\text{BN2d}+\text{ReLU}\\\text{Conv2d}, 3\times3, 128, 256, \text{stride}=1, \text{pad}=1 \\\text{BN2d}+\text{ReLU}\\\text{Conv2d}, 1\times1, 128, 256, \text{stride}=2, \text{pad}=0 \\\text{BN2d}+\text{ReLU}\\\\ \end{bmatrix} \times 4$\end{tabular}} & $N\times256\times64\times64$                        \\ \cline{2-3} 
\hline
\end{tabular}
\vspace{5pt}
\caption{Architecture of the motion encoder.}
\label{tab:encoder}
\end{table}

\clearpage
\subsection{Qualitative results on MoCA}
\label{ap:more_qualitative_res}
We provide further qualitative results of our model on the MoCA dataset in figure~\ref{fig:resCamo}.
\begin{figure}[!htb]
  \centering
    \begin{subfigure}{1\textwidth}
  \centering
    \includegraphics[width=0.3\textwidth]{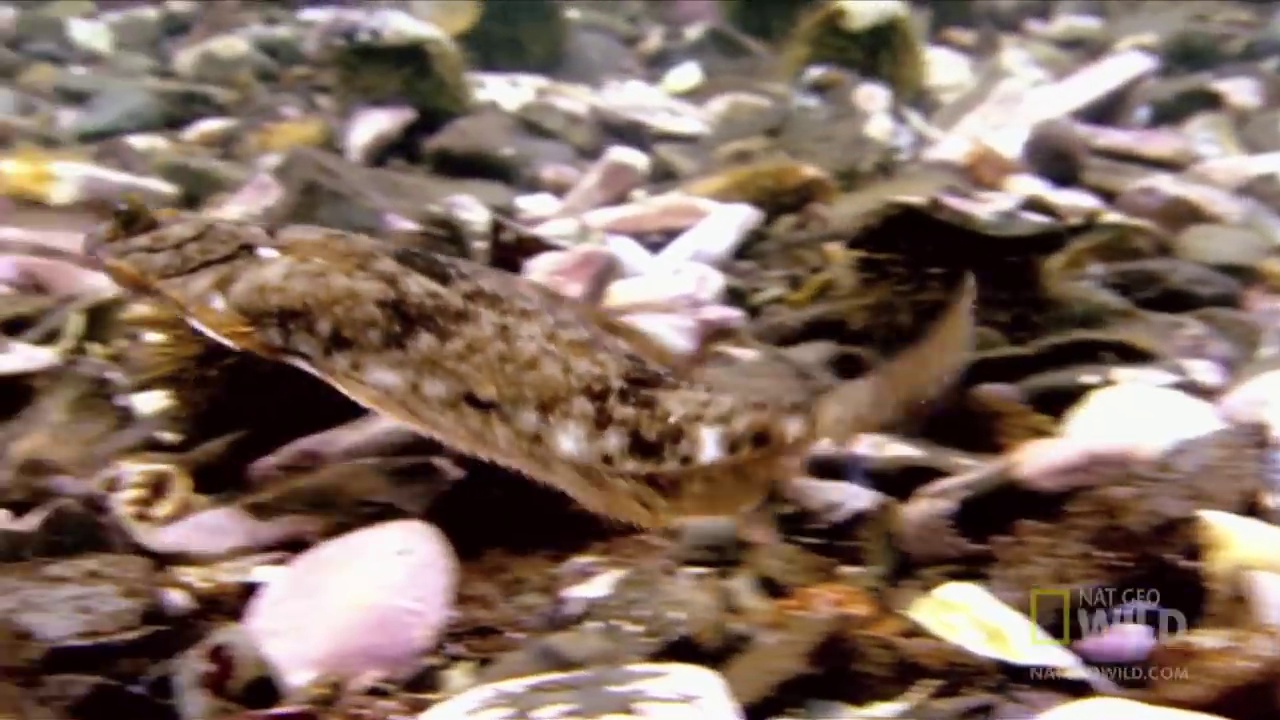}
    \includegraphics[width=0.3\textwidth]{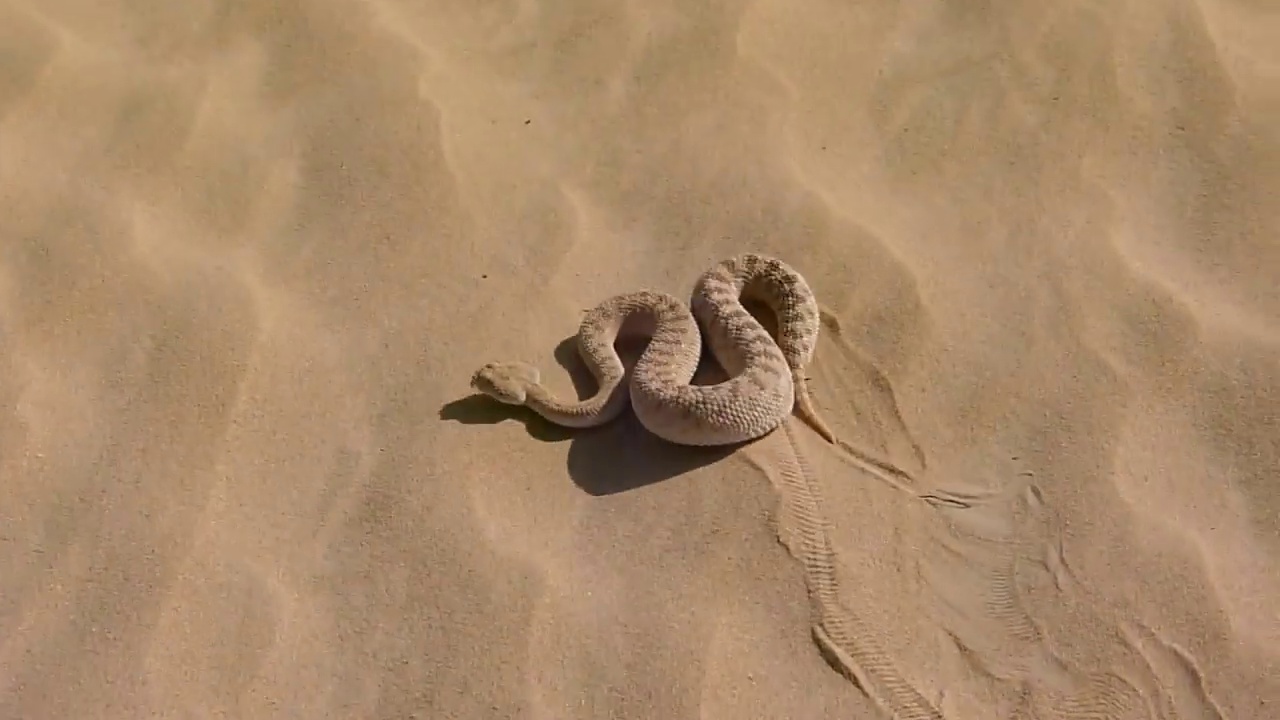}
    \includegraphics[width=0.3\textwidth]{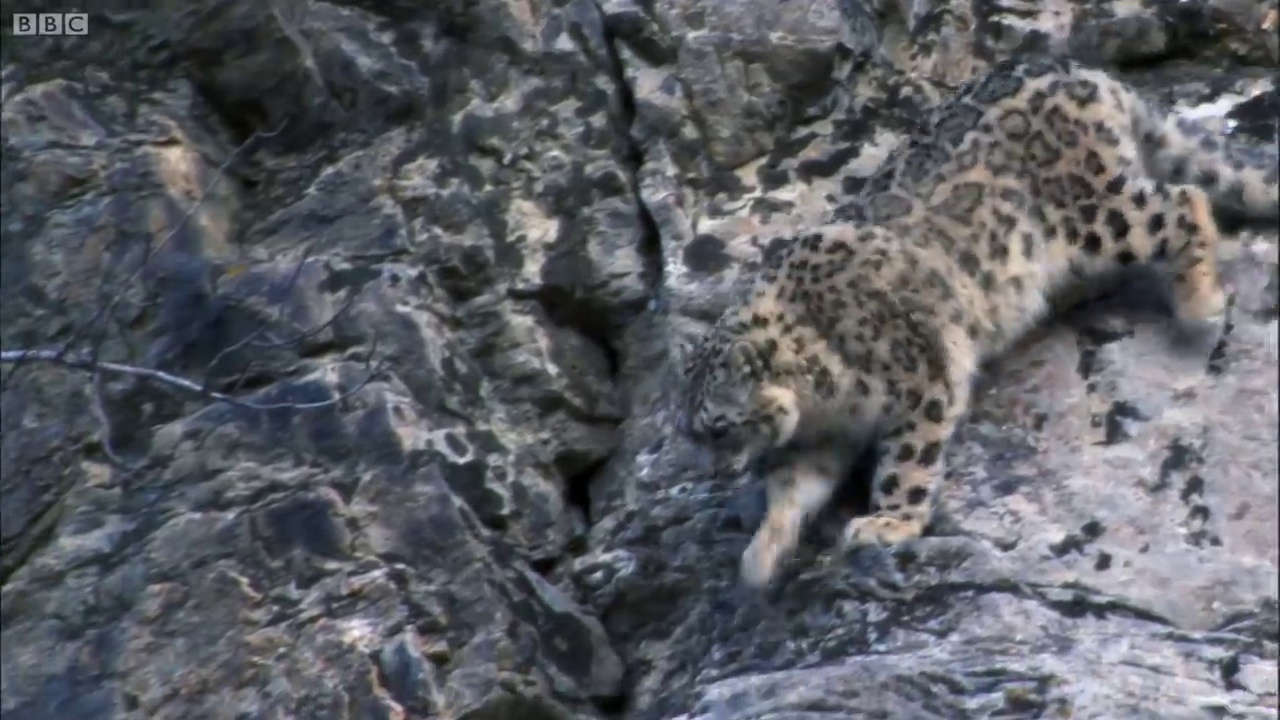}
  \end{subfigure}
  \begin{subfigure}{1\textwidth}
  \centering
    \includegraphics[width=0.3\textwidth]{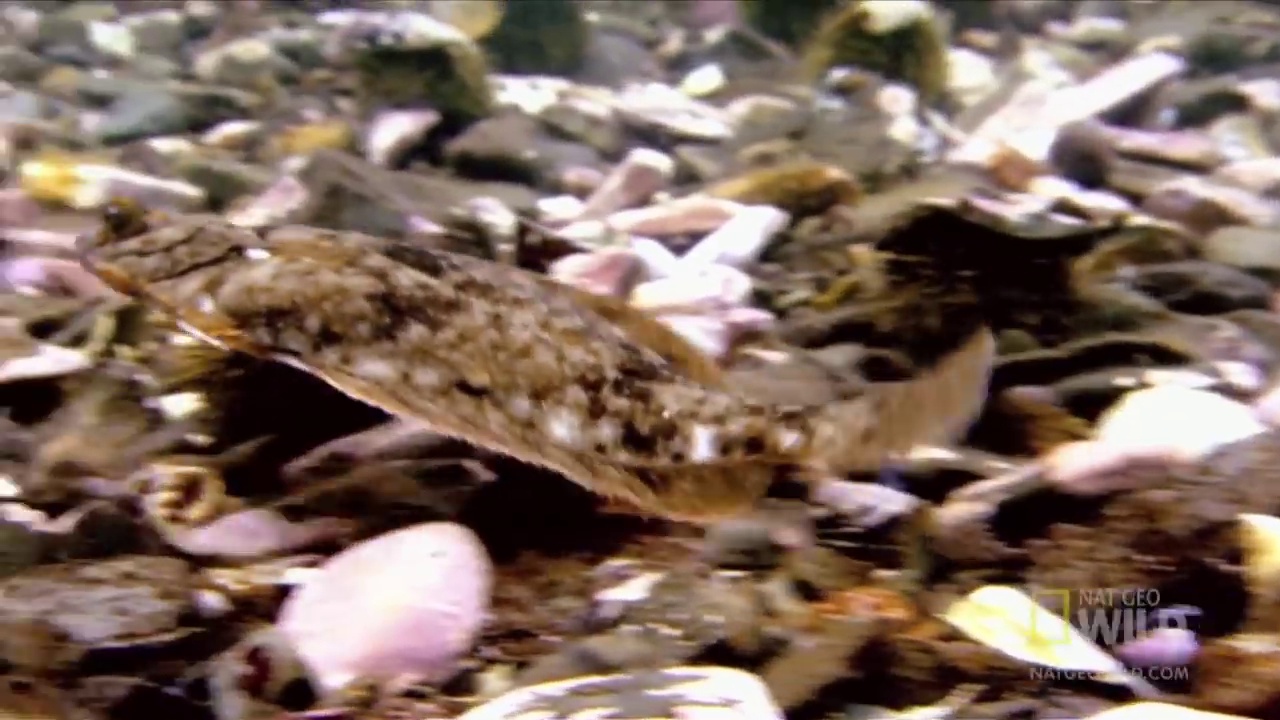}
    \includegraphics[width=0.3\textwidth]{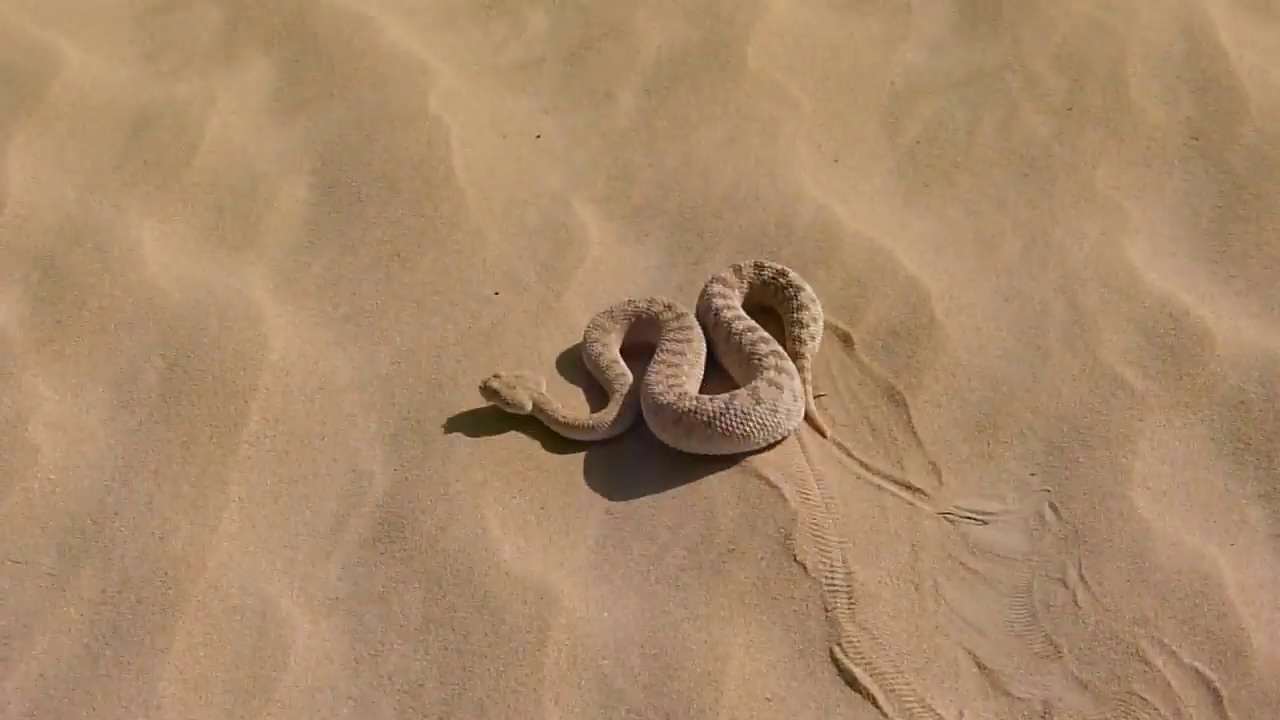}
    \includegraphics[width=0.3\textwidth]{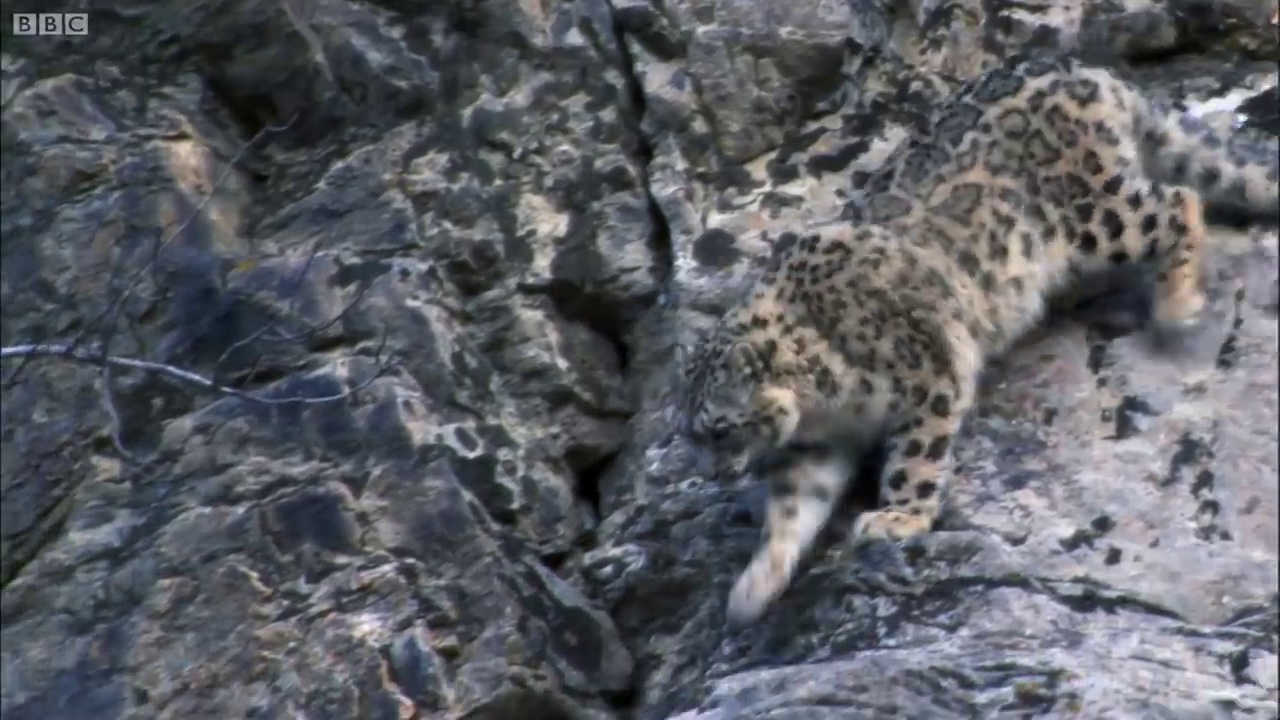}
  \end{subfigure}
   \begin{subfigure}{\textwidth}
    \centering
    \includegraphics[width=0.3\textwidth]{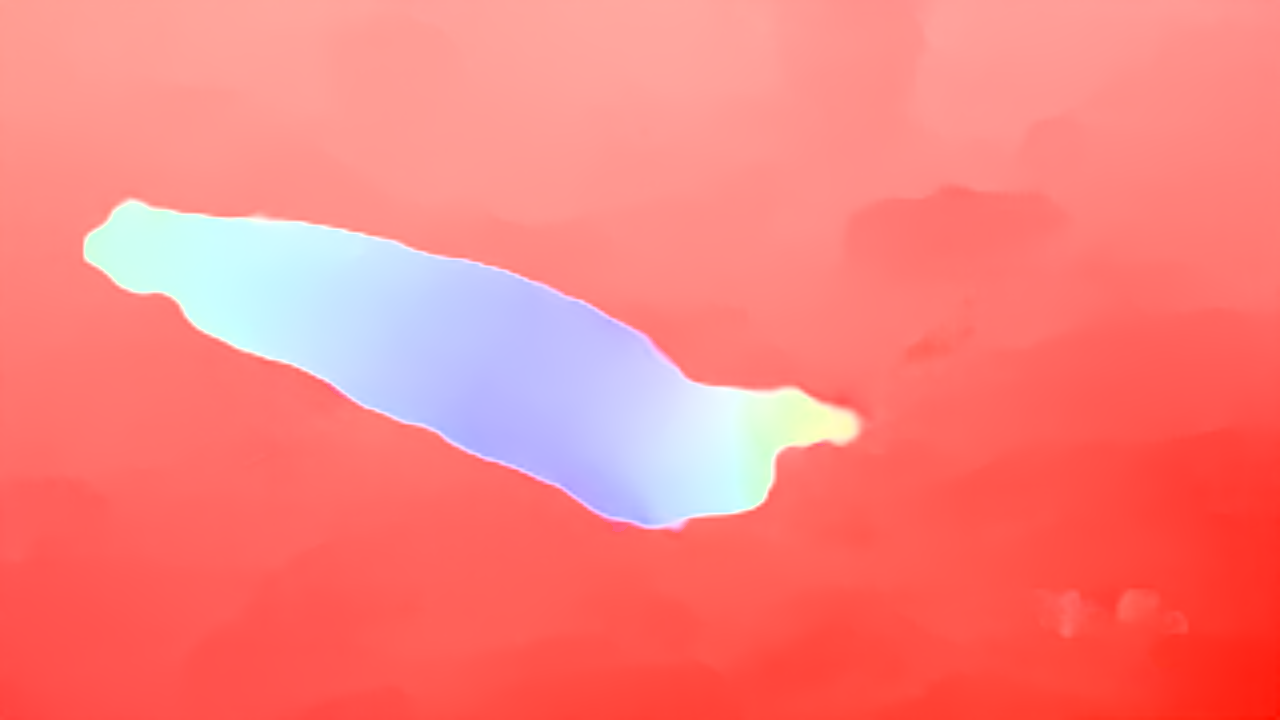}
    \includegraphics[width=0.3\textwidth]{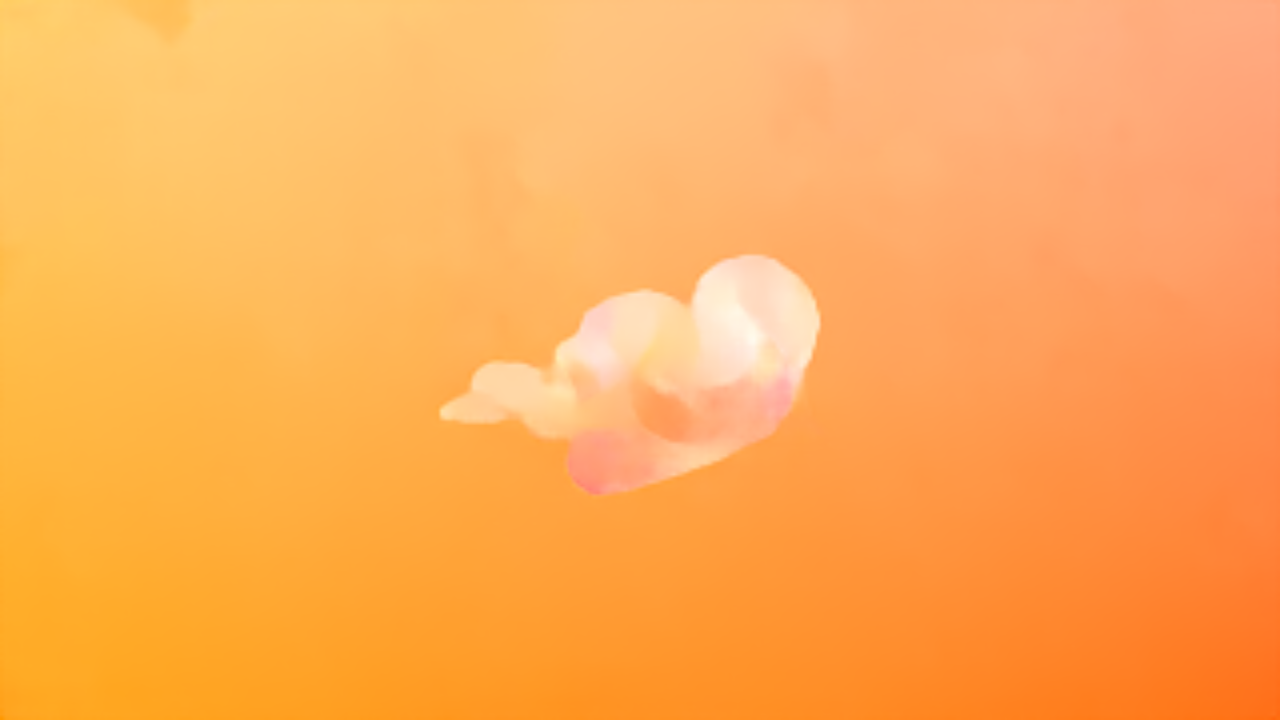}
    \includegraphics[width=0.3\textwidth]{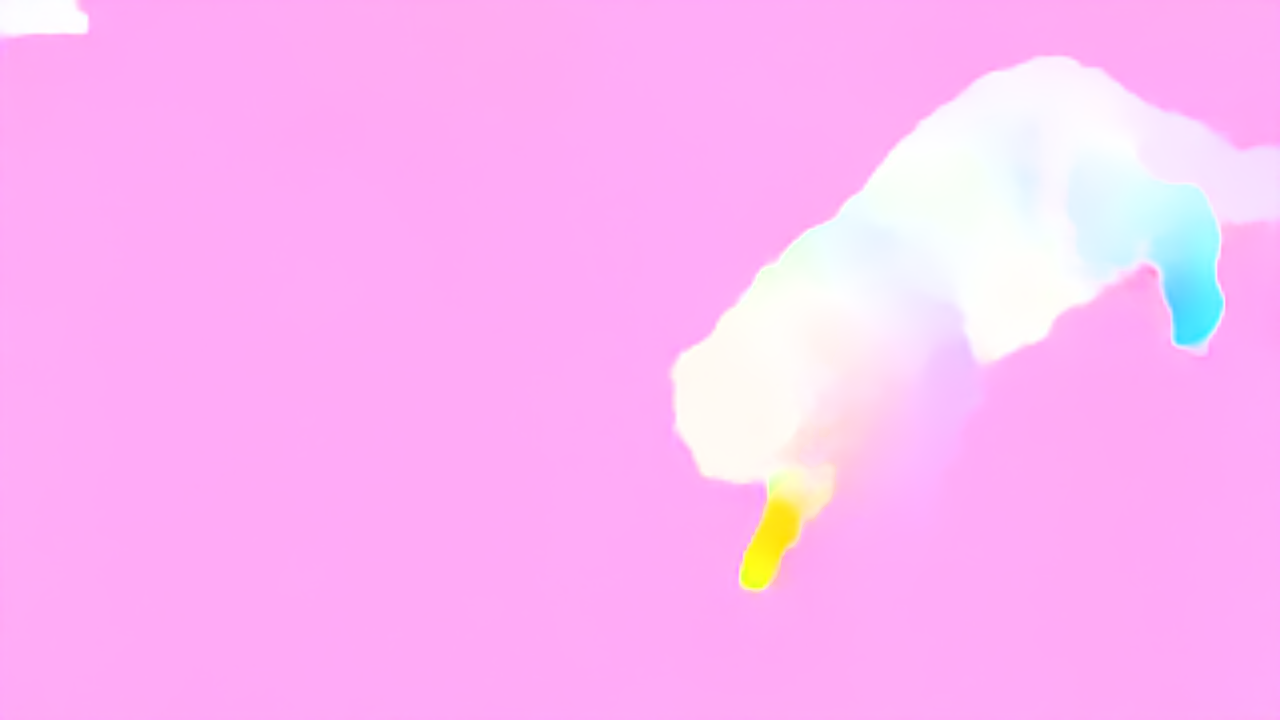}
  \end{subfigure}
     \begin{subfigure}{\textwidth}
    \centering
    \includegraphics[width=0.3\textwidth]{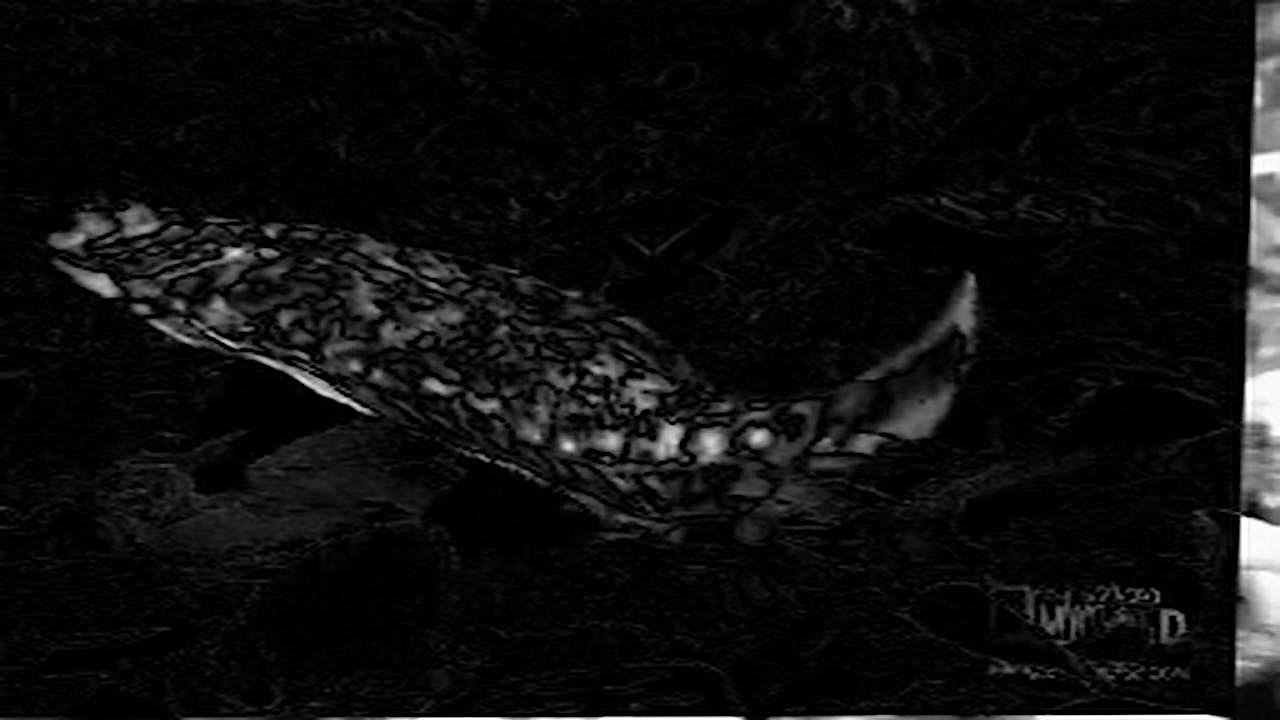}
    \includegraphics[width=0.3\textwidth]{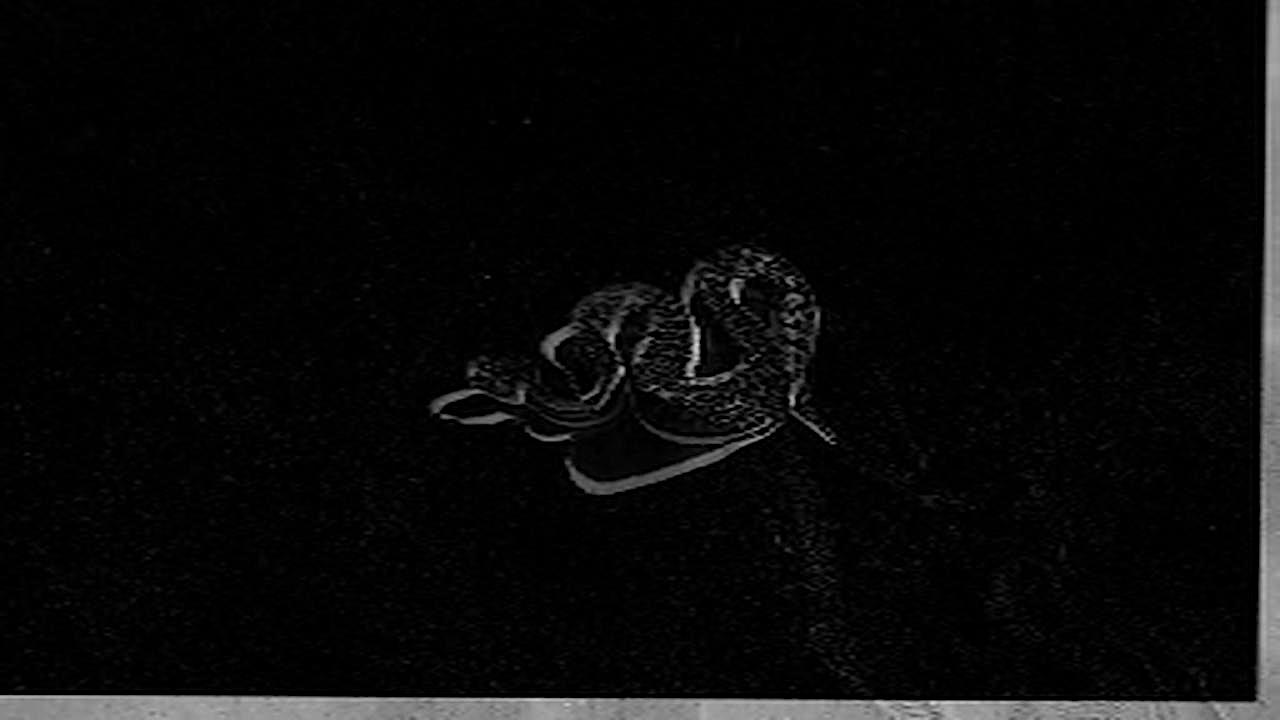}
    \includegraphics[width=0.3\textwidth]{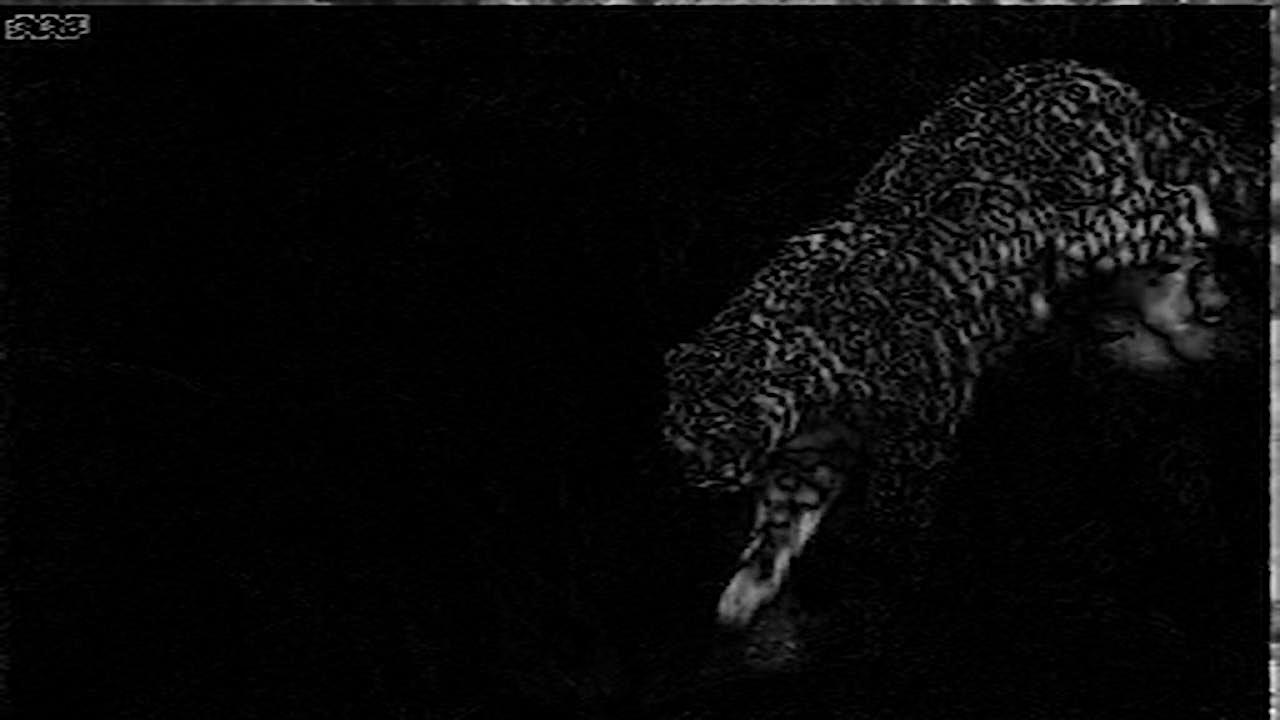}
  \end{subfigure}
   \begin{subfigure}{\textwidth}
   \centering
    \includegraphics[width=0.3\textwidth]{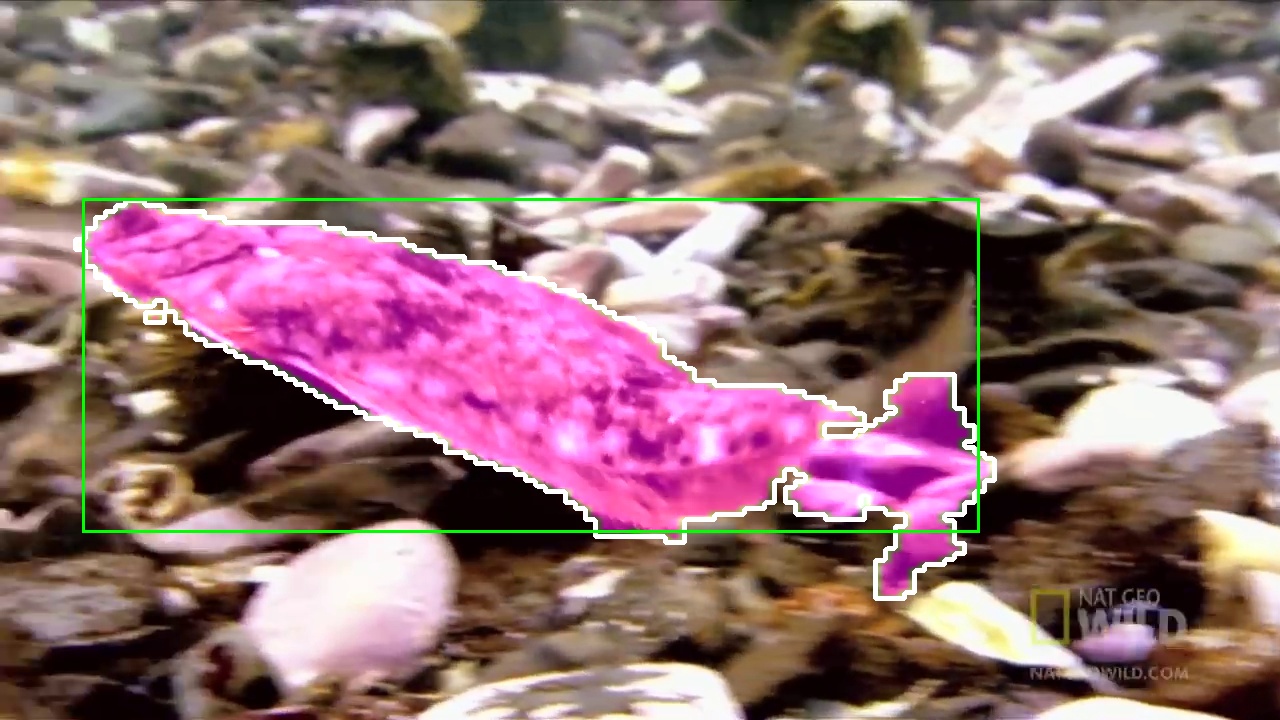}
    \includegraphics[width=0.3\textwidth]{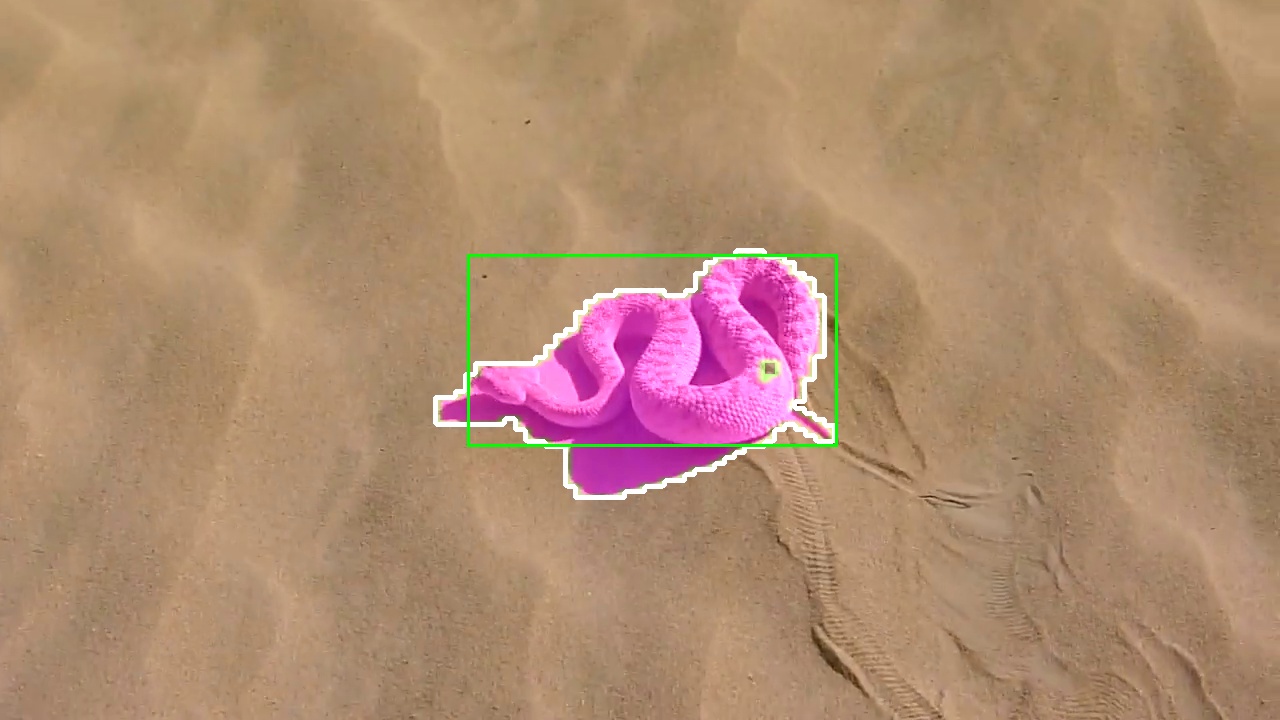}
    \includegraphics[width=0.3\textwidth]{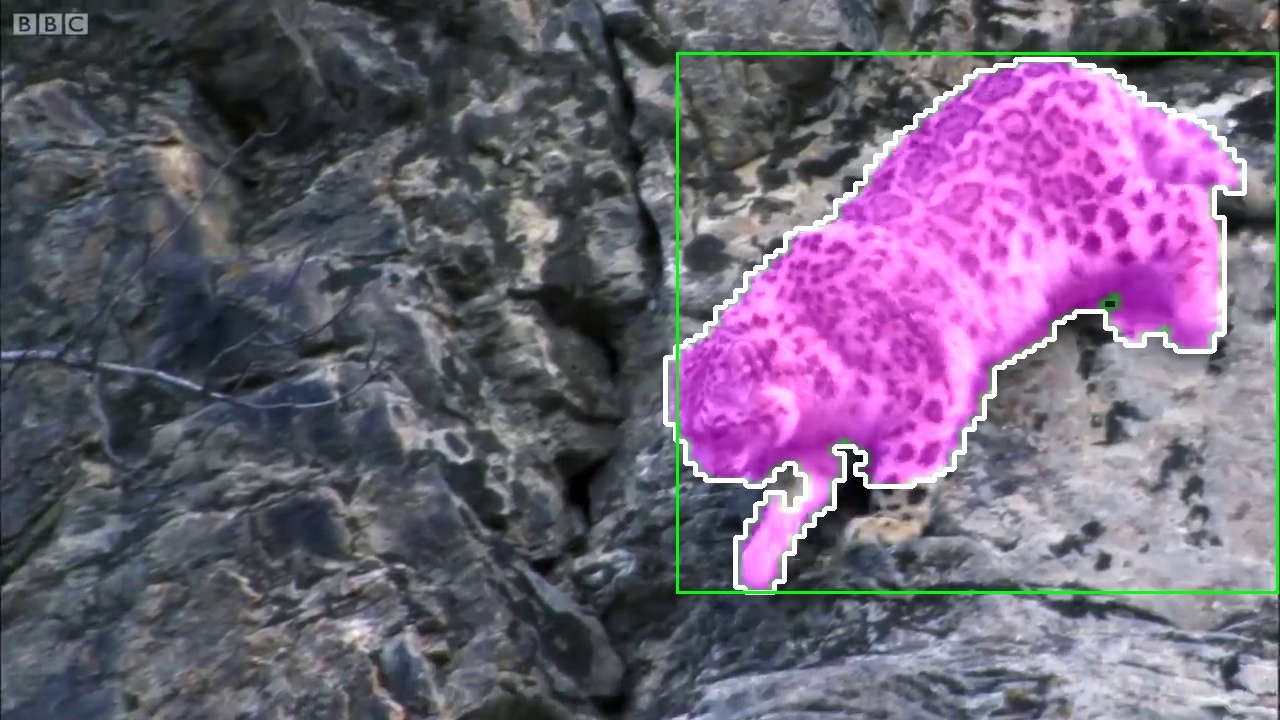}
  \end{subfigure}
\caption{More motion segmentation results on MoCA. 
From top to bottom: frame $t$, frame $t+1$, 
PWCNet optical flow, aligned image difference, and the predicted moving object segmentation (the output of our model). We also show the ground truth
annotation box in green}
\label{fig:resCamo}
\end{figure}
\end{document}